\documentclass[journal]{IEEEtran}
\IEEEoverridecommandlockouts
\usepackage{cite}
\usepackage{algorithm}
\usepackage{algorithmic}
\usepackage{graphicx}
\usepackage{textcomp}
\usepackage{xcolor}
\usepackage{hyperref}
\usepackage{orcidlink}
\hypersetup{hidelinks,
	colorlinks=true,
	allcolors=purple,
	pdfstartview=Fit,
	breaklinks=true}

\usepackage{color}
\usepackage{threeparttable}
\usepackage{amsmath,amssymb,amsfonts}
\usepackage{booktabs}
\usepackage{multirow}
\usepackage{multicol}
\usepackage{makecell}
\usepackage{rotating}
\usepackage{wrapfig}
\usepackage{subfigure}
\usepackage{colortbl}
\usepackage{marvosym}

\newcommand*\circled[1]{\tikz[baseline=(char.base)]{
            \node[shape=circle,draw,inner sep=1pt] (char) {\scriptsize #1};}}
 
\def\BibTeX{{\rm B\kern-.05em{\sc i\kern-.025em b}\kern-.08em
    T\kern-.1667em\lower.7ex\hbox{E}\kern-.125emX}}
\begin{document}

\title{Learning-based Sketches for Frequency Estimation in Data Streams without Ground Truth
\thanks{
    Xinyu~Yuan was with the School~of~Computer~Science~and~Information~Engineering, Hefei~University~of~Technology, Hefei, 230601, China. He is now with the College~of~Computer~Science~and~Technology, Zhejiang~University, Hangzhou, 310027, China (Email: yxy5315@gmail.com).
}% <-this % stops a space
\thanks{Yan~Qiao and Meng~Li are with the Key~Laboratory~of~Knowledge~Engineering~with~Big~Data, Hefei~University~of~Technology, Hefei, 230601, China. Zhenchun~Wei is with the School~of~Computer~Science~and~Information~Engineering, Hefei~University~of~Technology, Hefei, 230601, China. (Email: \{qiaoyan,mengli,weizc\}@hfut.edu.cn)}% <-this % stops a space
\thanks{Zonghui~Wang and Wenzhi~Chen are with the College~of~Computer~Science~and~Technology, Zhejiang~University, Hangzhou, 310027, China. (Email: \{zhwang,chenwz\}@zju.edu.cn)}% <-this % stops a space
\thanks{Cuiying~Feng is with the School~of~Information~and~Software~Engineering, University~of~Electronic~Science~and~Technology~of~China, Chengdu, 610054, China. (Email: cfeng19@uestc.edu.cn)}}

\author{Xinyu~Yuan\textsuperscript{\orcidlink{0009-0005-6287-5712}}, Yan~Qiao\textsuperscript{\Letter~\orcidlink{0000-0002-4407-1762}},~\IEEEmembership{Member,~IEEE},
Meng~Li\textsuperscript{\orcidlink{0000-0003-3553-0813}},~\IEEEmembership{Senior~Member,~IEEE}, Zhenchun~Wei\textsuperscript{\orcidlink{0000-0003-2751-6501}},~\IEEEmembership{Member,~IEEE}, Cuiying~Feng\textsuperscript{\orcidlink{0000-0002-8368-9944}}, Zonghui~Wang\textsuperscript{\Letter~\orcidlink{0000-0002-8212-4618}} and 
Wenzhi~Chen\textsuperscript{\orcidlink{0000-0003-1674-4701}},~\IEEEmembership{Member,~IEEE}
% \IEEEauthorblockA{\IEEEauthorrefmark{2}School~of~Computer~Science~and~Information~Engineering, Hefei~University~of~Technology\\
% \IEEEauthorrefmark{4}College~of~Computer~Science~and~Technology, Zhejiang~University}\\
% \IEEEauthorrefmark{3}School~of~Information~and~Software~Engineering, University~of~Electronic~Science~and~Technology~of~China
}

% The paper headers
\markboth{A Preprint - \textsc{IEEE Transactions on Knowledge and Data Engineering}}%
{Shell \MakeLowercase{\textit{et al.}}: Bare Demo of IEEEtran.cls for IEEE Journals}
% The only time the second header will appear is for the odd numbered pages
% after the title page when using the twoside option.
% 
% *** Note that you probably will NOT want to include the author's ***
% *** name in the headers of peer review papers.                   ***
% You can use \ifCLASSOPTIONpeerreview for conditional compilation here if
% you desire.

\maketitle

\begin{abstract}
Estimating the frequency of items on the high-volume, fast data stream has been extensively studied in many areas, such as database and network measurement. Traditional sketches provide only coarse estimates under strict memory constraints. Although some learning-augmented methods have emerged recently, they typically rely on offline training with real frequencies or/and labels, which are often unavailable. Moreover, these methods suffer from slow update speeds, limiting their suitability for real-time processing despite offering only marginal accuracy improvements.
To overcome these challenges, we propose UCL-sketch, a practical learning-based paradigm for per-key frequency estimation. 
Our design introduces two key innovations: (i) an online training mechanism based on equivalent learning that requires no ground truth (GT), and (ii) a highly scalable architecture leveraging logically structured estimation buckets to scale to real-world data stream.
The UCL-sketch, which utilizes compressive sensing (CS), converges to an estimator that provably yields a error bound far lower than that of prior works, without sacrificing the speed of processing.
Extensive experiments on both real-world and synthetic datasets demonstrate that our approach outperforms previously proposed approaches regarding per-key accuracy and distribution. Notably, under extremely tight memory budgets, its quality almost matches that of an (infeasible) omniscient oracle. Moreover, compared to the existing equation-based sketch, UCL-sketch achieves an average decoding speedup of nearly 500 times.
To help further research and development, our code is publicly available at \includegraphics[height=1.0em]{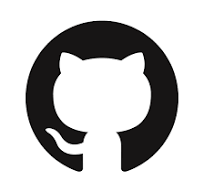}~\href{https://github.com/Y-debug-sys/UCL-sketch}{\texttt{https://github.com/Y-debug-sys/UCL-sketch}}.
\end{abstract}


% \vspace{-0.3cm}

\section{Introduction}

The frequency or volume estimation of unending data streams is a concern in many domains, starting with telecommunications but spreading to social networks, finance, and website engine. In network fields, for example, professionals want to keep track of the activity frequency to identify overall network health and potential anomalies or changes in behavior, which, however, is often challenging because the amount of information may be too large to store in an embedded device or to keep conveniently in fast storage~\cite{cormode2017data}. As a consequence, \textit{sketch}, which is a set of counters or bitmaps associated with hash functions, and a set of simple operations that record approximate information~\cite{yang2018empowering}, has grown in popularity in the context of high-velocity data streams and limited computational resources. Such an approximate algorithm is much faster and more efficient, yet this comes at the expense of unsatisfactory accuracy and cover proportion, especially when facing unbalanced stream characteristics, such as a Zipf or Power-law distribution~\cite{kumar2004data,roy2016augmented}. 

\begin{figure}
    \centerline{\includegraphics[width=1.\linewidth]{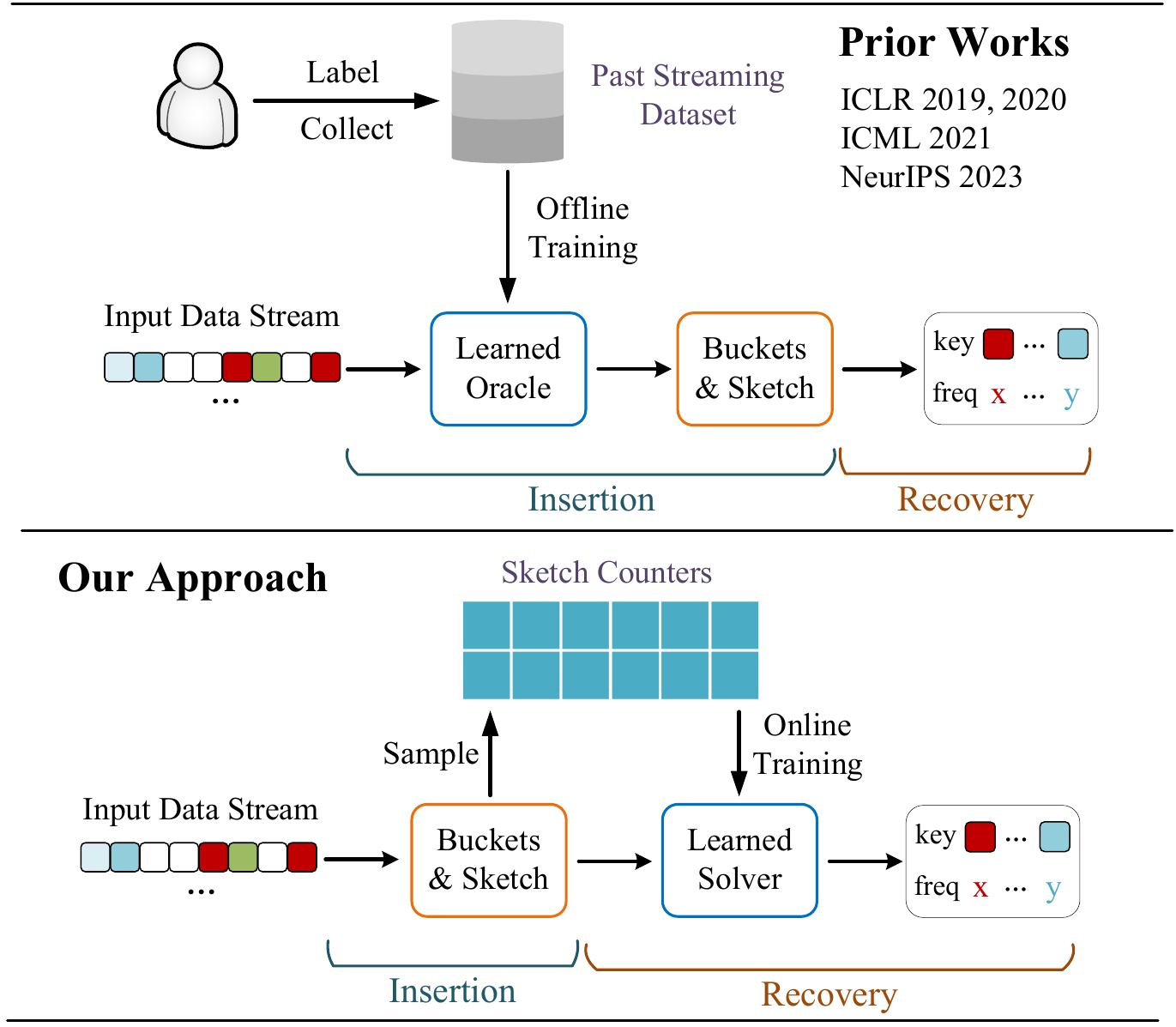}}
    \caption{\textbf{Comparison between the previous learning-augmented sketches and our studied learning-based sketch:} In our approach, we empower the sketch with learning technologies in the recovery phase to improve streaming throughput. The model is online trained using just compressed counters in the sketch, which is much more practical and efficient than the prior works.}
    \label{compare}
    % \vspace{-0.45cm}
\end{figure}

On the other hand, recent years have witnessed the integration of deep learning technology with numerous classic algorithms: index~\cite{kraska2018case}, bloom filter~\cite{mitzenmacher2018model,rae2019meta}, caching~\cite{lykouris2021competitive}, graph optimization~\cite{khalil2017learning} and so on. In particular, the research about learning-augmented streaming algorithms~\cite{hsu2019learning,jiang2020learning,du2021putting,yan2022talentsketch,cheng2023alsketch,aamand2024improved} is receiving significant attention due to the powerful potential of machine learning (ML) to relieve or eliminate the binding of data characteristics and the sketch design. Their typical workflow involves training a heavy hitter oracle, which receives a key and returns a prediction of whether it will be heavy or not, then inserts the most frequent keys into unique buckets and applies a sketch to the remaining keys. Although filtering heavy items has been proven to improve the overall sketch performance on heavy-tailed distribution~\cite{roy2016augmented,hsu2019learning}, these offline and supervised methods could hardly work in real-world applications. First, an unavoidable difficulty in designing algorithms in the learned sketch model is that ground truth like actual frequencies or labels for which key is large are not known in advance~\cite{jiang2020learning}.
% , let alone the actual frequency of all potential ``elephant'' items at that moment.
Moreover, since their models are only fitted on the past data, 
% while key frequencies accumulate and change consistently, 
the prediction performance of the oracle tends to deteriorate rapidly over time. That is to say, the neural network must be retrained with new labeled datasets frequently, therefore, all the above-mentioned sketches face a common problem in terms of updating the out-of-date classifier~\cite{li2020survey}. Besides, the idea of passing a deep model to reduce conflicts also incurs more insertion time and space cost, which may not be advisable for hashing-intense situations by considering data stream is processed sequentially in only one pass.

Building upon the limitations observed in the ``hashing-enhanced'' learning strategy, our study here is primarily motivated by equation-based sketches~\cite{fu2020clustering,sheng2021pr,huang2021toward}, from the perspective of compressed nature of sketching algorithms~\cite{cormode2011sketch}. Specifically, they employed a compressive sensing (CS) approach in the \textit{query phase} to achieve a very low relative error, given counter values and per-key aggregations. While these works showed great applications of the CS theory~\cite{cormode2005towards} to sketch-based estimation, their sensing matrix constructing and iterative-optimization-style recovery operation to get all frequencies of observed keys introduces considerable time and memory complexity even with top-performing solvers~\cite{dalt2022bayesian}. 

Consequently, one cannot help but pose the following intriguing question: \textit{Can we design a learning-based sketch (without ground truth) via the linear system by training and recovering on the fly?} 

The comparison between the question-oriented approach in this work and existing learned sketches is shown in Fig.~\ref{compare}. Without slowing down the conventional sketch insertions, our approach continuously trains the ML model to recover per-key frequencies using only sketch counters. Although promising, the process raises the following two challenges: (1) The first is \textit{self-supervision}. Unsupervised learning, without access to real frequencies or labels, is crucial for overcoming the impracticalities associated with existing learning-based frequency estimation algorithms. (2) The second is \textit{scalability} or \textit{complexity}. Many modern streaming scenarios have evolved into complex systems featuring tens of thousands of distinct items, and the entire key space invariably includes some uncertain keys that will be observed in the future. To decrease the complexity for per-key prediction, models dealing with such data need to be highly scalable as the size of streams grows infinitely. 

To address these challenges, we introduce a new frequency estimation framework called \textbf{UCL-sketch} (\textbf{U}nsupervised \textbf{C}ompressive \textbf{L}earning \textbf{Sketch}), which aims to integrate the advantages of both equation-based and learned sketching approaches. This framework achieves a significant improvement in practical feasibility and accuracy compared to learning-augmented algorithms, while maintaining substantially lower query overhead than equation-based competitors. A notable distinction from prior works is that our model is entirely ground-truth free, relying solely on downsampled frequencies for online training. This property endows UCL-sketch with great flexibility and the capability for quick response to streaming distribution drift. To realize these benefits, we theoretically and empirically demonstrate that the recovery function can be learned from compressed measurements alone using an \textit{equivalent learning} scheme, given per-key aggregations and keys set. Additionally, to mitigate the impact of large-scale and unbounded streams, we adopt the concept of \textit{logical buckets} to split and jointly learn multiple bucket-associated mappings with shared parameters, leading to an efficient and expandable architecture. Experimental results demonstrate the potential of our proposed algorithm through detailed evaluations of the frequency estimation problem. Our contributions can be summarized as follows:

\begin{itemize}
\item We pioneer a new learning-based paradigm in frequency estimation by introducing UCL-sketch—a scalable, fast, and fully unsupervised framework that directly recovers per-key frequencies from sketch counters, advancing the practicality of existing learning-augmented sketches.
\item We present a comprehensive analysis of UCL-sketch. Our theoretical results establish that  its accuracy is fundamentally sound, and so high performance in practice can be achieved by just sampling sufficient “snapshots” of sketch counters. We also provide rigorous justification for the design insights behind each of its key components.
\item Extensive empirical evaluations compare UCL-sketch, in terms of both quality and runtimes, to frequency estimation with state-of-the-art sketches using both publicly available data and synthetic data. Our evaluation covers small and large streams, different skewness of data, and different memory usage. The main results show that \circled{1}  UCL-sketch achieves estimation quality almost matching that of an infeasible oracle with perfect knowledge of future insertions, and its accuracy remains largely unaffected by changes in storage resources and stream conditions. \circled{2} By invoking a learned solver in the query time, UCL-sketch achieves runtimes 1-2 orders of magnitude faster than running a (greedy) CS algorithm. 
\end{itemize} 

We organize the reminder of the paper as follows. Section~\ref{sec:2} reviews related work on sketching algorithms. Section~\ref{sec:3} introduces the system model of equation-based sketches and presents the motivations behind this work. Section~\ref{sec:4} details the design of the proposed frequency estimation framework. Section~\ref{sec:5} provides a comprehensive theoretical analysis to support our design. Section~\ref{sec:6} evaluates the proposed framework through extensive experiments on both real-world and synthetic data streams. Finally, Section~\ref{sec:7} concludes the paper and discusses potential directions for future research.

\begin{figure*}
    \centerline{\includegraphics[width=1.\linewidth]{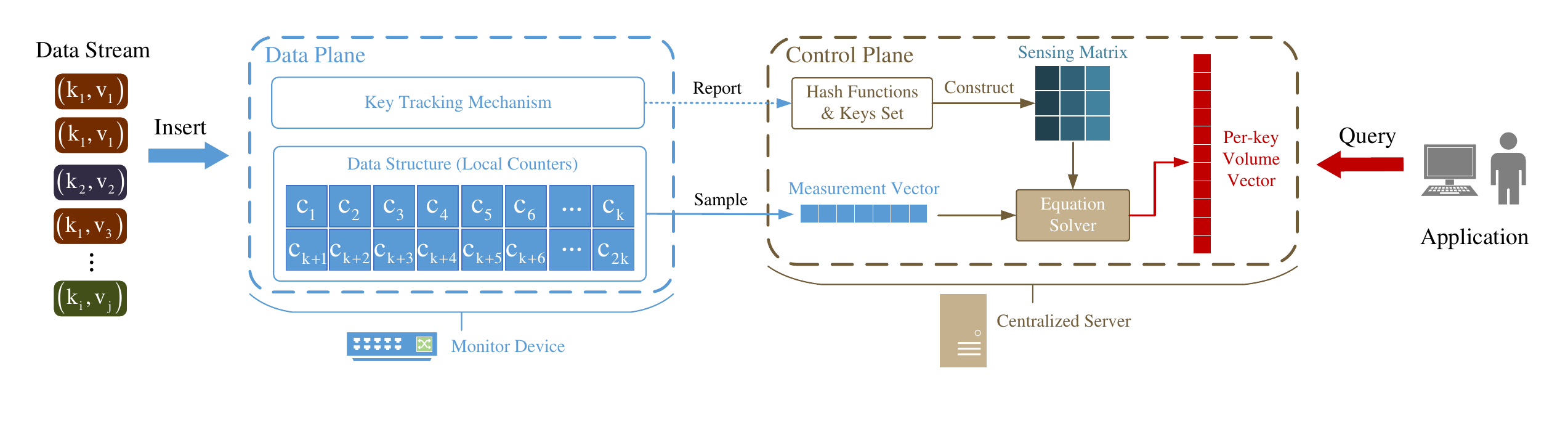}}
    \caption{\textbf{The overall processing framework of equation-based sketch:} In the data plane, it builds a local sketch to record the data stream and a key tracking mechanism for new item identification and reporting. After the centralized server receives sketch counters and keys from the monitor device, the control plane can recover the frequencies through solving an under-constrained equation system.}
    \label{process_overview}
    % \vspace{-0.45cm}
\end{figure*}

% \vspace{-0.2cm}

\section{Related Work}
\label{sec:2}

% This section presents the related works on three types of sketching algorithms: classical, equation-based, and learning-based sketches.

\textbf{Classic Sketch}. A sketch is a compact structure and solution which takes limited space to support approximate frequency queries over high-speed data streams. The most representative algorithms~\cite{cormode2005improved,charikar2002finding,estan2002new,roy2016augmented,liu2016one,yang2018elastic,liu2019nitrosketch} include Count-Min Sketch (CM-sketch), Count Sketch (C-sketch), 
and Augmented Sketch (A-sketch). They adopt a common underlying structure which is essentially a $w \times d$ array of counters for preserving key frequencies. Each of the $d$ rows of the array is associated with a hash function for mapping items to $w$ counters, then they consider the counts of $d$ different buckets array (e.g. minimum for CM-sketch~\cite{cormode2005improved}, median for C-sketch~\cite{charikar2002finding}) to which the data is mapped as the estimation. The CU-sketch~\cite{estan2002new} changed the insertion of the CM-sketch, which only updates the value of the minimum bucket in each insertion process.
The A-sketch~\cite{roy2016augmented} added a filter to the CM-sketch and exchanged data between the filter and the sketch to ensure that hot keys are retained in the filter to reduce hash conflicts.
Elastic Sketch~\cite{yang2018elastic} adopts a pottery shards eviction strategy for optimizing its replacement: If the inserted ID matches the ID stored in the bucket, the positive vote count is incremented; otherwise, the negative vote count is incremented. When the ratio of negative to positive votes exceeds a certain threshold, the stored ID is replaced with the inserted one. Nitrosketch~\cite{liu2019nitrosketch} introduces a sampling mechanism and multi-level control logic. Instead of sampling packets directly, it uses a sample counter array and applies geometric sampling to reduce per-packet processing, thereby saving on update operations of a sketch like C-sketch. For network-wide monitoring, UnivMon~\cite{liu2016one} builds on multi-layer sketches (e.g., C-sketch) with hierarchical sampling and inverse transformation to support accurate traffic analysis~\cite{wang2025survey} with low memory.
Since classic sketches have been proven to deliver high accuracy only with impractical memory consumption, these algorithms are subject to an undesirable compromise between estimation accuracy and memory efficiency.

\textbf{Equation-based Sketch}. Recent works have made progress in mitigating the trade-off by designing advanced query methods of sketch algorithms. \cite{lee2005improving} and~\cite{lu2008counter} first disclose that sketch and compressive sensing are thematically related. The locality-sensitive sketch (LSS)~\cite{fu2020clustering} leverages the relationship between the sketch with the compressed projection, then extends it to a K-means clustering method. The PR-sketch~\cite{sheng2021pr} recovers keys of streaming data by establishing linear equations. 
Specifically, it builds linear equations between counter values and per-key aggregations to improve accuracy, and records keys in the recovery phase to reduce resource usage in the update phase. Finally, PR-sketch fixes the system by a conjugate gradient solver.
The SeqSketch~\cite{huang2021toward} and HistSketch~\cite{he2023histsketch} store a few high-frequency items, then employ a compressed-sensing approach (i,e, OMP) to decode infrequent keys. By solving the linear system, these equation-based sketches compensate for the error introduced by counter sharing, and recovers the complete keys in the shared part with much higher accuracy than the classical sketch. Unfortunately, such global sketches have been shown to suffer from greatly increased time and memory costs of estimation.

\textbf{Learning-based Sketch}. In the last few years, machine learning has taken the world by storm than ever before, which also motivates the design of learning-based frequency estimation algorithms~\cite{yang2018empowering}. 
\cite{yang2018empowering} pioneered the idea of employing machine learning to reduce the dependence of the accuracy of sketches on network traffic characteristics. 
TalentSketch~\cite{yan2022talentsketch} applies a long short-term memory (LSTM) model to network measurement tasks.
In~\cite{hsu2019learning}, the authors first proposed a learned frequency estimation framework by using a trained classifier (or oracle) to store hot and cold items separately. The overall design resembles A-sketch, where cold items are stored in a sketch structure such as CM-sketch or C-sketch; however, the latter incorporates a data exchange mechanism. 
\cite{jiang2020learning} adjusted the learning-augmented sketch, which uses a regression model to directly outputs the predicted frequency of hot keys rather than inserting them in unique buckets.
Then a series of works have also studied theoretical analyses and optimizations under this framework~\cite{jiang2020learning,du2021putting,yan2022talentsketch,cheng2023alsketch,aamand2024improved}. However, these hand-derived methods are excessively dependent on offline models and cannot handle dynamic data distribution. Differently, UCL-sketch provides a new paradigm for learning-enhanced sketch design. It obtains accuracy close to original equation-based sketches while maintaining the efficient query cost by continuously adapting models without ground truth.

% \vspace{-0.2cm}

\section{Preliminaries}
\label{sec:3}

\subsection{Key Ideas of Equation-based Sketch}\label{eqsketch}

We follow prior studies that uncover the linear compression nature of equation-based sketch framework~\cite{fu2020clustering,sheng2021pr,huang2021toward}. Before delving into details, we define a subclass of sketches of interest, termed \textit{linear sketch}: an insertion of a single key-value pair updates only a fixed set of underlying counters. Typically, the linear sketch comprises an \textit{insertion} component that feeds the key-value input to a compact structure that approximates these key-value pairs with one or multiple hash-based counters arrays, a \textit{recovery} component that inverses queried pairs from key-aggregations based on the same set of hash functions. Theorem~\hyperref[theorem1]{1} establishes the equivalence between the sketch with the linear system as follows:

\textbf{Theorem 1.}\label{theorem1} \textit{The goal of frequency estimation based on a linear sketch is equivalent to solving linear equations from the given keys, hash functions, and counters. Let $x\in C^N$ denote the vector of the streaming key-frequency sequence and $y\in C^M$ denote sketch counters, the insertion process corresponds to $y=\boldsymbol{A}x$, while the result of recovery phase corresponds to}
\begin{equation}
    x=\boldsymbol{A}^\dag y+(I-\boldsymbol{A}^\dag \boldsymbol{A})x,
    \label{general_solution}
\end{equation} 
\textit{where $\boldsymbol{A}\in C^{M\times N}$ is an indicator matrix of mapping the vector $x$ to a buckets array $y$, and $\boldsymbol{A}^\dag \in C^{N\times M}$ satisfies $\boldsymbol{A}\boldsymbol{A}^\dag \boldsymbol{A}\equiv \boldsymbol{A}$.}

\textbf{\textit{Remark}}. The detailed proof of Theorem~\hyperref[theorem1]{1} can be found in Appendix, {Considering Theorem~\hyperref[theorem1]{1}, if we model a relation as defining a vector or matrix, then the sketch of this is obtained by multiplying the data by a (fixed) matrix. In this regard, a single update to the underlying volume has the effect of modifying a single entry in the frequency vector. The linear sketch can then be updated by adding to it the result of applying the sensing matrix to this individual change alone. Consequently, the problem of frequency estimation reduces to solving a linear inverse problem.}

As shown in Fig.~\ref{process_overview}, the equation-based sketch needs to build a local linear sketch in the data plane and perform its original update operation. It deploys an additional key tracking mechanism to identify new keys and transfer them to the control plane. What distinguishes the design from other sketches is that it leverages an equation-based approach to compensate for per-key error caused by counter sharing in the recovery phase (or control plane), which can be concluded as three steps: $\left({\rm i}\right)$ Transform sketch counters to the measurement vector $y$. $\left({\rm ii}\right)$ For all distinct stream items, construct a sketch operator $\boldsymbol{A}$ based on the hash functions that map them in the sketch. $\left({\rm iii}\right)$ Fix the system of linear equations by an equation solver. We now proceed to formally define the problem setting in the next subsection.

% \vspace{-0.3cm}

\subsection{Problem Statement}

Simply speaking, we formulate our frequency estimation algorithms via compressive sensing (CS), like the equation-based sketch introduced in Section~\ref{eqsketch}. Let a data stream of running length $n$ be a sequence of $n$ tuples. The $t$-th tuple is denoted as $(k_t, v_t)$, where $k_t$ is a data-item key used for hashing and $v_t$ is a frequency value associated with the item. For each item, the insertion process applies an update to sketch counters (vector) $y$ of length $M$, with a sensing matrix $\boldsymbol{A}$ defined by hash functions and the key value. Then given the collected $y$ and $\boldsymbol{A}$, the output of recovery phase is a list of estimated frequencies, i.e., ground-truth (GT) vector $x$. Also note that we assume the size of possible keys space $N > M$ in this work, because keys are usually drawn from a large domain (e.g, IP addresses, URLs) while available space in data plane is limited, leading to an ill-posed system. Formally, the recovery phase builds an optimization problem: $\mathop {\max }\limits_{{x}} \log p\left( {{x}|{y}} \right), \  {\rm {s.t.}} \ {y}={\boldsymbol{A}}{x}$.

% \vspace{-0.35cm}

\subsection{Motivations}\label{motivation}

\textbf{Limitation of Baseline Solution.} There have been many implementations and extensions of equation-based sketch. Among these sketching solutions, the PR-sketch~\cite{sheng2021pr} and SeqSketch~\cite{huang2021toward} represent the most recent examples. These methods involve key tracking mechanisms to collect distinct keys in the stream and apply optimization techniques to solve the linear system. However, the computational cost of streaming problems grows significantly with the number of keys, making iterative process of ``decode" algorithms less feasible for modern streams. As a result, they have remarkably increased per-key accuracy as well as computation time and peak memory consumption at query time.

\textbf{Impact of a Learned Equation Solver.} One idea here to eliminate the need for iterative optimization is a learned equation solver that directly maps measurement $y$ to frequencies $x$. While such a deep solver incurs additional training overhead, this cost can be amortized in parallel with ongoing stream processing. In return, it significantly reduces query processing latency by enabling one-shot prediction, while preserving the strong estimation accuracy of traditional (baseline) solvers—mirroring the success of learned approaches in the field of compressive sensing~\cite{gregor2010learning,pang2020self,chen2021equivariant}.

\textbf{Challenges of Learning-based Solution.} Building on the above discussion, our primary objective is to develop equation-based sketches that leverage learned solvers to automate per-key frequency recovery. However, this goal introduces two key challenges: (1) enabling online training without access to ground-truth frequencies, as acquiring real-time exact counts for all possible keys is impractical; and (2) maintaining alignment with large-scale data streams under strict constraints on parameter overhead, especially as the solver’s complexity grows over time. To address these issues, we present in the following section a detailed description of our proposed self-supervised learning framework for per-key frequency recovery.

% \vspace{-2mm}
% \vspace{-0.2cm}

\section{Methodology}
\label{sec:4}

In this section, we present our method, called UCL-sketch, and elaborate on its design. At a high level, UCL-sketch inherits the architectural framework of equation-based sketches but replaces traditional iterative decoding algorithms with a learned solver. To overcome the challenge of GT-free online training, we introduce the concept of equivalent learning and develop a solver trained using windowed sampled counters. Moreover, to support large-scale data streams with dynamically expanding key spaces, we incorporate position encoding with parameter sharing, enabling logically unbounded scalability of the key space.

% \vspace{-2mm}
% \vspace{-0.2cm}

\subsection{Basic Design}

\begin{figure*}
    \subfigure[Data structure of UCL-sketch\label{data_structure_graph}]{
        \centering
        \includegraphics[width=0.4\linewidth]{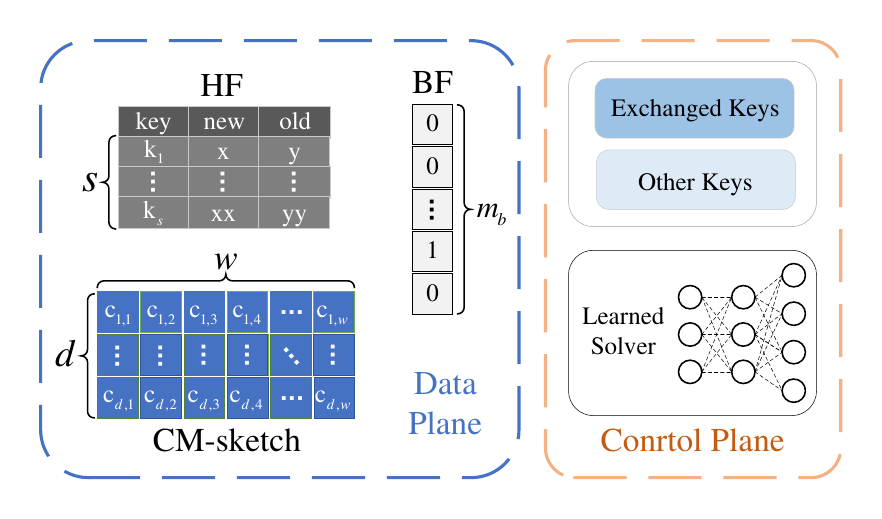}
    }
    \hfill
    \subfigure[Examples of update processing by UCL-sketch\label{insertion}]{
        \centering
        \includegraphics[width=0.59\linewidth]{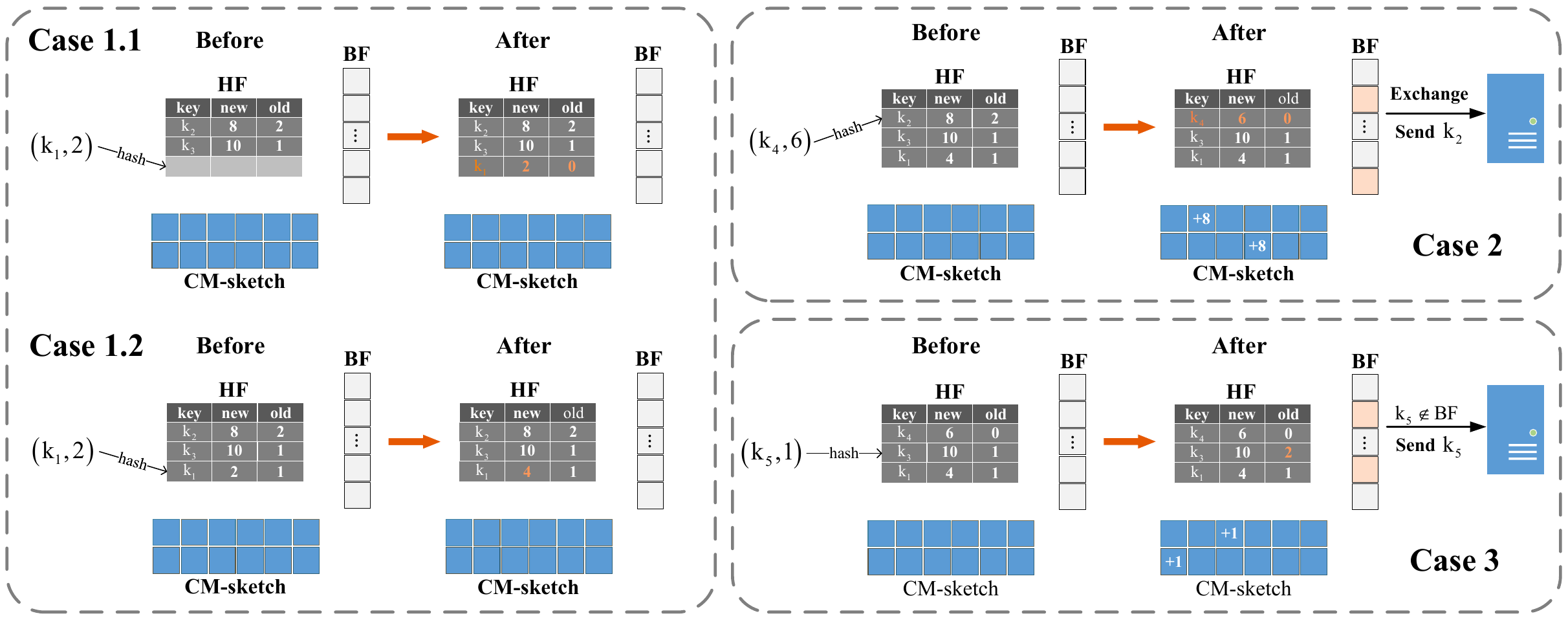}
    }
    \caption{\textbf{The basic design of UCL-sketch:} (a) UCL-sketch maintains a hash table, a CM-sketch, and a Bloom filter in the data plane, while deploying two key sets in the control plane (storing keys evicted from the hash table and other keys, respectively), along with a learning-based solver. (b) During stream processing in the data plane, each incoming item is first checked against the hash table; if the insertion fails, the item is inserted into the CM-sketch instead, and its key is reported to the control plane if it was either recently evicted from the hash table or not found in the Bloom filter.}
    % \vspace{-0.3cm}
\end{figure*}

\subsubsection{Key Ideas} To mitigate the extra bandwidth overhead used for transmitting keys, the UCL-sketch employs a Bloom Filter during the update phase, ensuring that each unique key is identified and reported at most once. Moreover, we filter hot keys twice, storing them separately in a hash table and an array, which has been proven beneficial for highly skewed data streams~\cite{roy2016augmented,huang2021toward}.

\subsubsection{Data Structure} Fig.~\ref{data_structure_graph} depicts the data structure of UCL-sketch. In the data plane, it has three types of data structures: (1) a heavy filter (hash table) \textit{HF} to track frequent key pairs, (2) a sketch to record the remaining items, and (3) a Bloom Filter \textit{BF} for key identification. Each slot in the HF consists of three fields. In addition to a key identifier, the slot contains two counters: \textit{new count}, which tracks the values associated with the key, and \textit{old count}, which records the values not attributed to the key. 
As discussed in Section~\ref{sec:3}, the sketch used here can be any linear sketch. In this work, particularly, we adopt the CM-sketch due to its simple and fast insertion (see Appendix for other implementations).
For the control plane, apart from a learned solver, we maintain two non-repeating and non-overlapping arrays to record inserted keys and exchanged keys from HF in our sketch, respectively.

\subsubsection{Update Operation} The procedure of inserting an item in our UCL-sketch is very similar to previous equation-based sketch, e.g. SeqSketch~\cite{huang2021toward}. The main difference is the exchanged keys that are supposed to be relatively hot are stored separately in preparation for subsequent training phase.

Algorithm~\hyperref[alg1]{1} outlines the procedure of inserting a key-value pair ($k, v$). We first compute its hash position in HF, then as Fig.~\ref{insertion} shows, there are overall three cases:

\begin{algorithm}[htb] 
\caption{Update Operation on the UCL-sketch} 
\label{alg1} 
\begin{algorithmic}[1] 
\REQUIRE key-value pair $(k, v)$
\STATE $i \leftarrow \mathrm{hash}(k)$;
\IF{HF[$i$].key == Null}
    \STATE HF[$i$].key $\leftarrow k$; HF[$i$].new $\leftarrow v$; HF[$i$].old $\leftarrow 0$;
\ELSIF{HF[$i$].key == $k$}
    \STATE HF[$i$].new $\leftarrow$ HF[$i$].new + $v$;
\ELSE
    \STATE HF[$i$].old $\leftarrow$ HF[$i$].old + $v$;
    \IF{HF[$i$].new $<$ HF[$i$].old}
        \STATE insert CM-sketch with (HF[$i$].key, HF[$i$].new); 
        \STATE insert BF with HF[$i$].key; 
        \STATE report hot key HF[$i$].key to control plane; 
        \STATE HF[$i$].key $\leftarrow k$; HF[$i$].new $\leftarrow v$; HF[$i$].old $\leftarrow 0$;
    \ELSE
        \STATE insert CM-sketch with $(k, v)$;
        \IF{$k \notin$ BF}
            \STATE insert BF with $k$;
            \STATE report cold key $k$ to the control plane;
        \ENDIF
    \ENDIF
\ENDIF
\end{algorithmic}
\end{algorithm}

\textit{Case 1}: The slot is empty or the existing entry has the same key. We insert the item into the position or just increment the new count by its value.

\textit{Case 2}: The current position does not have the same key and $($new count - old count$)$ $\le$ 0 after incrementing the old count by the item's value. We replace the existing entry with the new item and evict the old entry into the sketch. Then we transfer the exchanged key with the ``hot'' flag to the control plane, and update the BF.

\textit{Case 3}: The current position does not have the same key and $($new count - old count$)$ $>$ 0 after incrementing the old count by the item's value. We insert the item into the sketch. After inserting the BF, only if it is identified as a new key, UCL-sketch sends the key to the control plane.

Given a heavy-tailed distribution of stream data, Case 1 constitutes a large portion while Case 2 represents the opposite. Thus after filtering by the hash table, UCL-sketch is memory efficient as the number of exchanges is usually very small~\cite{roy2016augmented}.

\subsection{Training Strategy}

\begin{figure*}
    \centerline{\includegraphics[width=1.\linewidth]{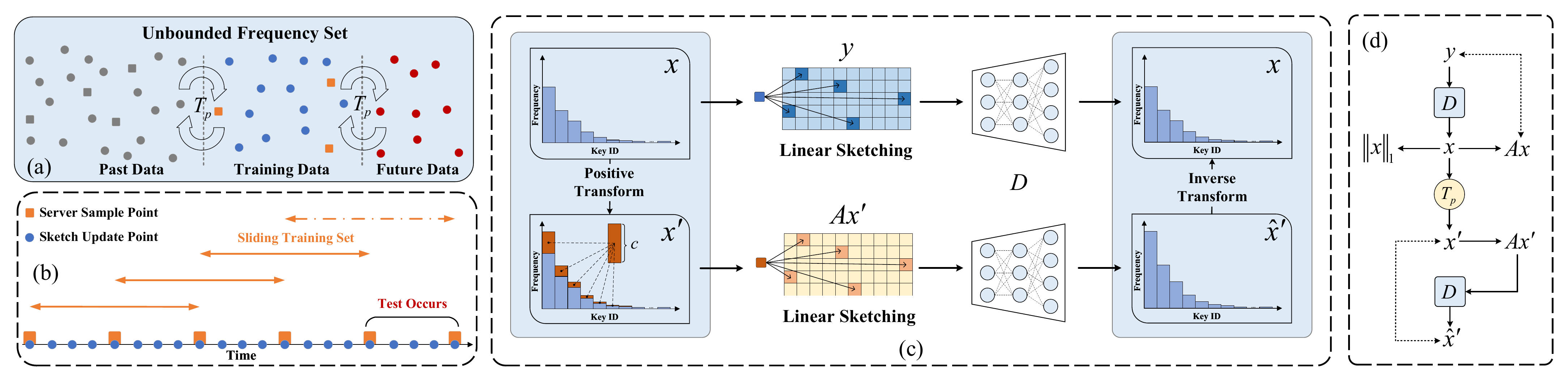}}
    \caption{\textbf{Main ideas of training strategy:} (a) The unbounded set of frequency vectors is tolerant of certain Zipfian transformations. (b) Continually adapting the model using sampled counters in a sliding window for online training. (c) The learned solver $y \rightarrow x$ should also be invariant to these natural transformations. (d) Illustration of our self-supervised equivalent loss design.}
    \label{learning}
    % \vspace{-0.45cm}
\end{figure*}

\begin{figure*}
    \centerline{\includegraphics[width=1.\linewidth]{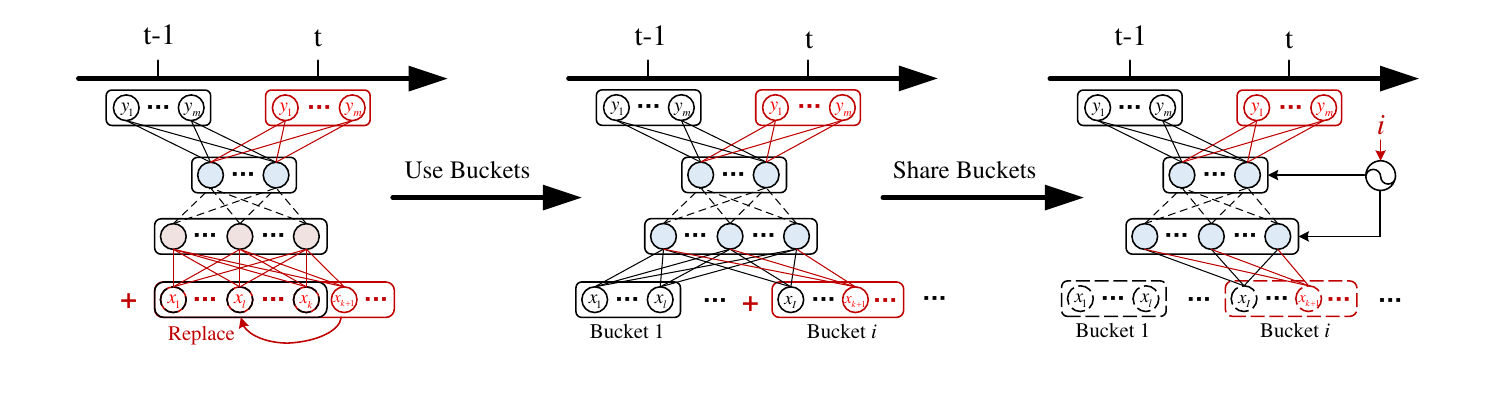}}
    \caption{\textbf{Expansion design of a learned solver:} \textbf{Left:} \textit{Redesigning}. Whenever a new key emerges, the entire output layer learned on previous streams is replaced and retrained. \textbf{Center:} \textit{Partial-redesigning with buckets}. The keys are separated into independent buckets, so the solver is affected only by newly updated bucket. \textbf{Right:} \textit{Non-redesigning with logical buckets}. By sharing buckets, the solver can be adapted to varying key space, and greatly reduces the number of parameters.}
    \label{expansion}
    % \vspace{-0.45cm}
\end{figure*}

\subsubsection{Goal and intuitions} As mentioned in Section~\ref{motivation}, we consider a challenging but reasonable setting in which only sampled measurement vector $y$, and the sketch sensing matrix $\boldsymbol{A}$ are available for on-line training our solver $D: y=\boldsymbol{A}x \rightarrow x$. As shown in Eq.~\ref{general_solution}, the root problem is a non-trivial null space defined by $(I-\boldsymbol{A}^\dag \boldsymbol{A})x$ while $y$ only provides the information of range space of $\boldsymbol{A}$, i.e., its pseudo-inverse $\boldsymbol{A}^\dag$. Therefore, a simple solver without additional constraints is not efficient enough to resolve the GT ambiguity. Our intuition for achieving unbiasedness is the distribution of item frequencies follow approximate Zipf's law, then the output of the solver should be invariant to certain groups of transformations that allow us to learn beyond the range space.

\subsubsection{Online Training} First of all, we present an online training procedure for continually adapting the learned solver. When analyzing or processing continuous streaming data, one only needs to keep track of recent stream because queries always occur in the future, while the useful information contained in past streaming data is diminishing over time. Therefore, as shown in Fig.~\ref{learning}~(b), our approach for dealing with non-stationary data streams is to adopt a sliding window (SW) mechanism which retains a fixed number of sampled ``snapshots'' of sketch counters in memory, instead of training on the entire history. The concept revolves around maintaining a ``window" that slides with time, capturing only the recent state of the unbounded stream. The sample point depends on the times of sketch updates, for instance, these counters are transmitted to the control plane after every 1,000 insertions.

\subsubsection{Equivariant Learning} A straightforward practice of unsupervised frequency recovery is to impose the measurement consistency on the model, using a range space loss of form like $\left\| {\boldsymbol{A}x - y} \right\|_2^2$. However, the sketch operator $\boldsymbol{A}$\footnote{Note that we do not need to explicitly maintain the 0-1 sparse sensing matrix $\boldsymbol{A}$ in memory, although the control plane has sufficient space. Instead, we generate its elements on demand.} has a null space, which means the model converges freely to the biased solution $\boldsymbol{A}^\dag y + H(y)$ with $\boldsymbol{A}H(y) = 0$. This will cause unstable reconstructions without meeting ground truth data or prior information. According to CS theory, one direct way to alleviate the problem is ``$\ell_1$-minimization'' by adding a regularization $\left\| {x} \right\|_1$ penalizes against the lack of sparsity~\cite{huang2021toward}. Fortunately, it makes sense in streaming algorithms because heavy-tailed data such as network traffic exhibits high sparsity, but it is not enough to eliminate the impact of null space, since the accuracy of particular items is still unsatisfactory~\cite{li2023cs}.

\begin{wrapfigure}[10]{r}{0.3\textwidth}
  \vspace{-0.6cm}
  \setlength{\abovecaptionskip}{-0.1cm}
  \begin{center}
  \includegraphics[width=1.\linewidth]{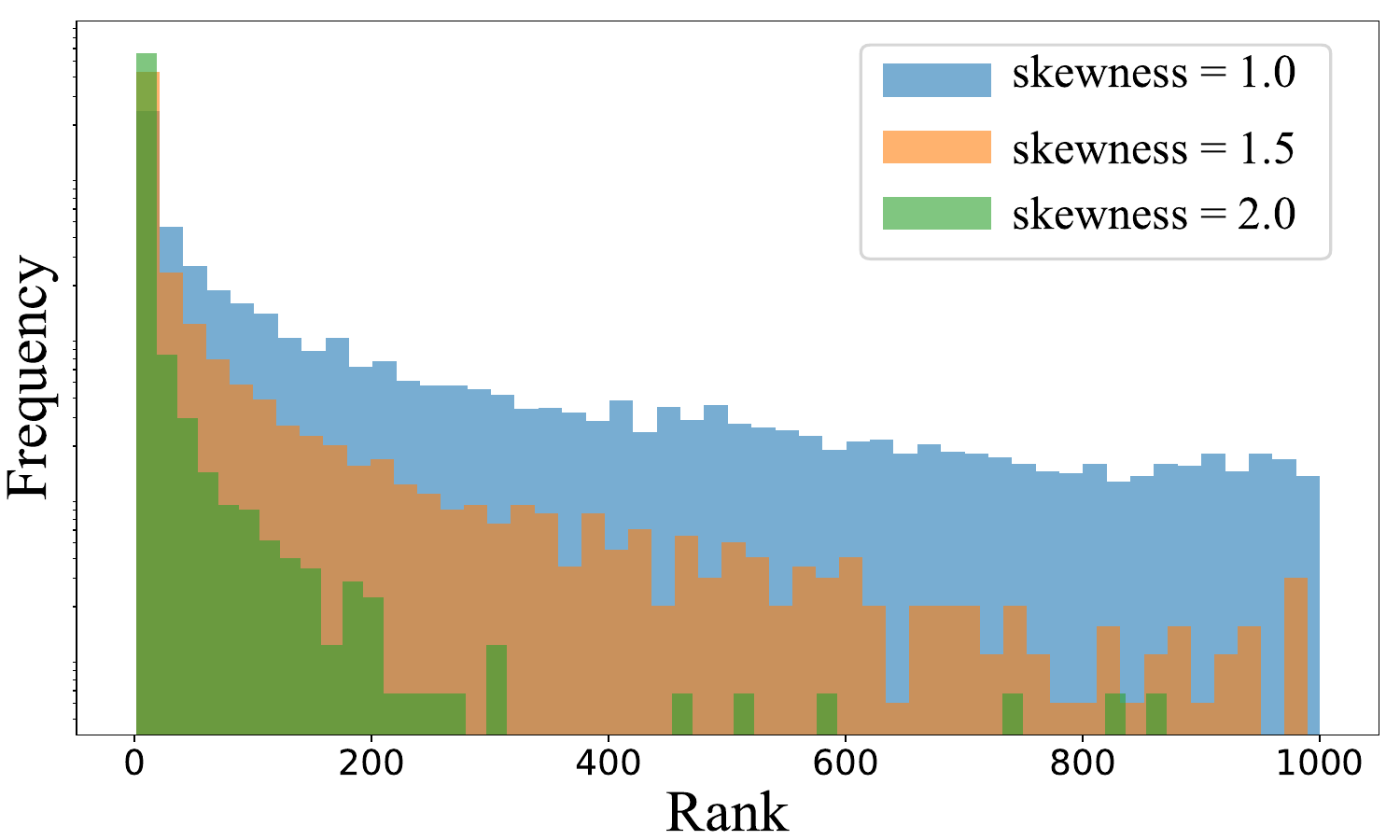}
\caption{Example of the Zipfian distribution with different skewness.}
\label{zipfian}
\end{center}
\end{wrapfigure}

In order to learn more knowledge beyond the range space of $\boldsymbol{A}$, we draw on ideas from the Zipf Law prior of streaming item frequencies, and this is a common and natural reoccurring pattern in real-world data~\cite{aamand2024improved}. At their simplest, zipfian models are based on the assumption of a simple proportionality relationship: ${f(k_j)} \propto \frac{1}{j}$, where $f(k_j)$ is the frequency of $j$-th key $k_j$ in keys set sorted by volume as shown in Fig.~\ref{zipfian}. Furthermore, we introduce an additional assumption of ``temporal smoothness'' in the heavy-tailed distribution—a well-established principle in real-world measurement studies~\cite{zhang2009spatio}. This assumption captures the empirical observation that frequencies within \textit{adjacent} temporal sample points tend to express similar zipfian behavior: a few frequent items exhibit approximate proportional growth, while other infrequent items remain their original size.

% In order to learn more knowledge beyond the range space of $\boldsymbol{A}$, we draw on ideas from the Zipf Law prior of streaming item frequencies, and this is a common and natural reoccurring pattern in real-world data~\cite{aamand2024improved}. At their simplest, zipfian models are based on the assumption of a simple proportionality relationship: ${f(k_j)} \propto \frac{1}{j}$, where $f(k_j)$ is the frequency of $j$-th key $k_j$ in keys set sorted by volume as shown in Fig.~\ref{zipfian}. Furthermore, we posit an additional assumption of ``temporal smoothness'' in the heavy-tailed prior, which is a well-established assumption in the field of real-world measurement~\cite{zhang2009spatio}, observing that frequencies within \textit{adjacent} temporal sample points tend to express similar zipfian behavior: a few frequent items exhibit approximate proportional growth, while other infrequent items remain their original size.

Under this mild prior information, we define a group of transformations $\mathcal{P}=\left\{ p_{1,\dots,|\mathcal{P}|} \right\}$, in which arbitrary $p_i$ can be summarized in the following steps: Given a positive integer $c$ which usually takes value around the sampling interval, $p_i$ allocates it proportionally based on the volume of exchanged (or hot) keys in the input frequency vector. Next, the transformation randomly selects a small number of frequencies, i.e. 5$\%$ from the remaining non-hot keys, and increments them by minimum update unit. As Fig.~\ref{learning}~(a) shows, for all possible $x$ in the unbounded frequency set $\mathcal{X}$, the equivalent relationship $T_{p}x \in \mathcal{X}, \ \forall p \in \mathcal{P}$ holds, where $T_p \in \mathbb{R}^{N \times N}$ is the corresponding transformation matrix of $p$. Then, our learned solver $D$ should also capture such invariant proportional property of the target domain, that is, $D(\mathbf{A}T_px)=T_pD(\mathbf{A}x)$. This additional constraint on the mapping allows the model to learn beyond the range space (see details in Theorem~\hyperref[theorem3]{3}). If the incremental component can be handled by the solver, then the ambiguity in null space recovery can be effectively mitigated during learning. As shown in Fig.~\ref{learning} (c) and (d), the network weights are updated by minimizing the following objective:
\begin{equation}
    \begin{aligned}
    \mathop {\arg \min }\limits_{{\theta}} {\mathbb{E}_{y \in \boldsymbol{A}\mathcal{X}, p \in \mathcal{P}}} & \{ \left\| {\boldsymbol{A}D\left( y \right) - y} \right\|_2^2 + \lambda {{\left\| {D\left( y \right)} \right\|}_1} \\
    &+ \left\| {D\left( {\boldsymbol{A}{T_p}D\left( y \right)} \right) - {T_p}D\left( y \right)} \right\|_2^2 \},
    \end{aligned}
    \label{objective}
\end{equation}
where the first term enforces measurement consistency, the second term imposes sparse constraint, and the third term enforces system equivariance, and $\lambda$ is a trade-off coefficient. The training pseudo-code is exhibited in Algorithm~\hyperref[alg2]{2}.

\begin{table*}[htb]
\normalsize
\resizebox{\textwidth}{!}{
\begin{tabular}{l}
\toprule
\textbf{Algorithm 2} One-Epoch Training Algorithm of UCL-sketch\\
\midrule
\begin{minipage}{0.6\linewidth}\label{alg2}
    \centering
    \begin{algorithmic}[1]
    \REQUIRE $\eta$, $\lambda$, sketching matrix $\boldsymbol{A}$, exchanged key ids $h$,\\
    ~~~~~~~measurement set $\boldsymbol{Y}$, and number of keys $n$
    \ENSURE trained model $D$
    \FOR{measurement vector $y$ in $\boldsymbol{Y}$}
         \STATE $x \leftarrow$ Per$\_$Key$\_$Recovery($D, y, n$);
         % \STATE $\hat y \leftarrow \boldsymbol{A}x$;
         \STATE $x' \leftarrow$ Positive$\_$Transform($x, h, n$);
         \STATE $\hat x' \leftarrow$ Per$\_$Key$\_$Recovery($D, \boldsymbol{A}x', n$);
         \STATE Gradient descent on $\eta {\nabla _{{\theta _D}}} ( {\left\| {\boldsymbol{A}x - y} \right\|_2^2 + \left\| {\hat x' - x'} \right\|_2^2 + \lambda \left\| {x} \right\|_1} )$;
    \ENDFOR
    \RETURN $D$;
    \end{algorithmic}
\end{minipage} \\
\bottomrule
\end{tabular}
\begin{tabular}{l}
\toprule
\textbf{Algorithm 3} Query Operation on the UCL-sketch\\
\midrule
\begin{minipage}{0.5\linewidth}\label{alg3}
    \centering
    \begin{algorithmic}[1]
    \REQUIRE learned model $D$, keys set $\boldsymbol{\Omega}$, bucket length $L$,\\
    ~~~~~~~~current measurement vector $y$, and key $k$ 
    \ENSURE estimated frequency $x_k$
    \STATE position $\leftarrow $ Get$\_$Key$\_$Position($\boldsymbol{\Omega}, k$);
    \STATE bucket{$\_$}id $\leftarrow$ position // $L$;
    \STATE inner{$\_$}id $\leftarrow$ position - bucket{$\_$}id $\times$ $L$;
    \STATE $x \leftarrow$ $D$($y$, bucket${\_}$id);
    \STATE $x_k^s \leftarrow x$[inner$\_$id];
    \STATE $x_k \leftarrow x_k^s +$ Heavy$\_$Filter$\_$Query($k$);
    \RETURN $x_k$;
    \end{algorithmic}
\end{minipage} \\
\bottomrule
\end{tabular}}
% \label{alg}
% \vspace{-0.3cm}
\end{table*}

\begin{figure*}
    \centerline{\includegraphics[width=1.\linewidth]{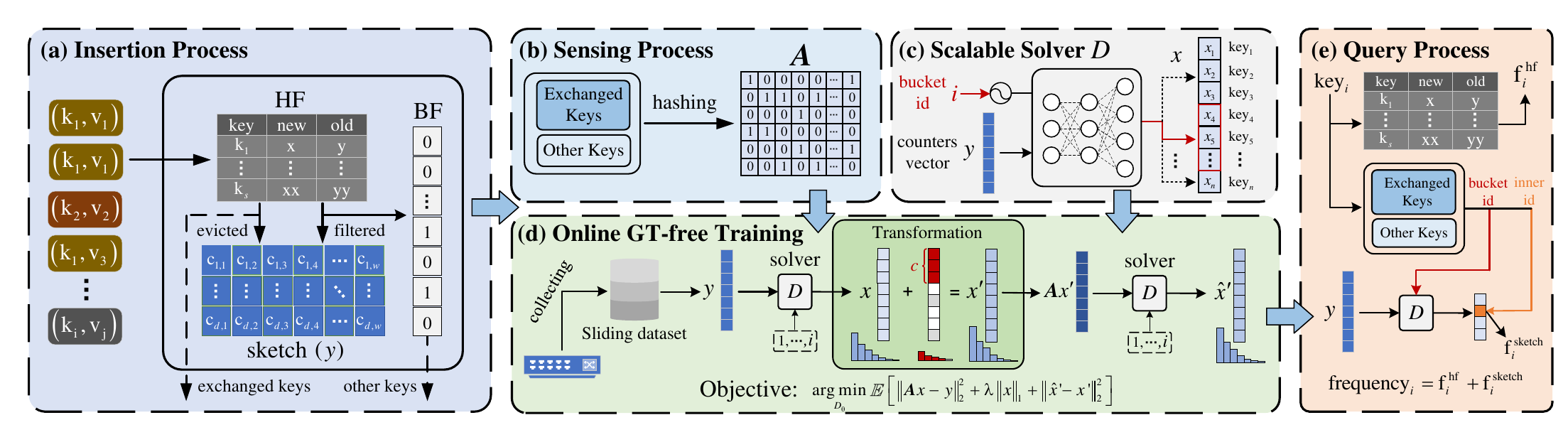}}
    \caption{The overview of complete version of our UCL-sketch.}
    \label{final_version}
    % \vspace{-0.45cm}
\end{figure*}

% \vspace{-0.3cm}

\subsection{Scalable Architecture}

\subsubsection{Goal and intuitions} UCL-sketch is explicitly designed to allow for sketching large-scale data steam, where an unknown number of new items arrive at the monitor device in sequence, rather than designed for a fixed or small key space. Specifically, our goal here is to let the learned solver dynamically expand its capacity once new elements arrive, while achieving efficiency in parameters. The intuition for designing such a lifelong network is to incrementally adapt to new items while retaining acquired frequencies of previous items, so parameter sharing is a good and natural choice.

\subsubsection{Network Expansion} A most naive way to design the network for a sequence of items would be retraining the output layer(s) every time a new item emerges. However, such retraining would incur significant costs for a deep neural network. Instead, we suggest dividing keys set into small buckets, where each bucket is maintained with its own parameters, thereby reducing the extra expanding overhead. Given the information collected about the observed stream, this design transforms the problem into a maximum-a-posteriori (MAP) estimate of the bucket-associated frequencies~\cite{dalt2022bayesian}. Unfortunately, it is still very intensive in terms of memory usage since the network's size scales with the observed keys, under high-speed streams where the computational cost is a significant concern. 

Notice that the statistical properties of sketch-based estimation should be stationary over buckets, as it implies that the similar posteriori transformation can be applied at each key (or bucket) inserted in the sketch (see details in Theorem~\hyperref[theorem4]{4}). Thus, a better solution in such a case is sharing parameters across these buckets. Specifically, we train a solver to model the mapping $(y,i) \rightarrow x_i$, in which $i$ denotes the index of our selected bucket and $x_i \subset x$ is predicted frequencies in the logical bucket. Borrowing ideas from literature in diffusion models~\cite{ho2020denosing}, we learn the bucket-shared network using the sinusoidal position embedding~\cite{ashish2017attention}. Our model details can be found in Appendix. After this bucket sharing and logification, the network needs to update its weights by repeatedly forward propagation since the split changes the overall structure. Fortunately, in practice, this operation can be performed for all logical buckets in parallel. Fig.~\ref{expansion} illustrates our dynamically scalable network architecture and shows what happens when the set is partitioned into independent buckets.

\subsubsection{Query Operation} When querying an item $k$, we initially locate its index in the keys set such that we can determine the bucket id and relative position of $k$. Then we query the hash table in the data plane to obtain its filtered frequency, and return 0 if the key is not in it. The partial result and sampled sketch counters will be reported to the server together. With the previously acquired position, we predict the remaining frequency of $k$ by inputting counters and bucket id into the learned solver. The final estimated frequency is a sum of the two parts. The process is depicted in Algorithm~\hyperref[alg3]{3}.

% \vspace{-0.3cm}

\subsubsection{Coping with Sketch Changes}

\begin{figure}
    \centerline{\includegraphics[width=1.\linewidth]{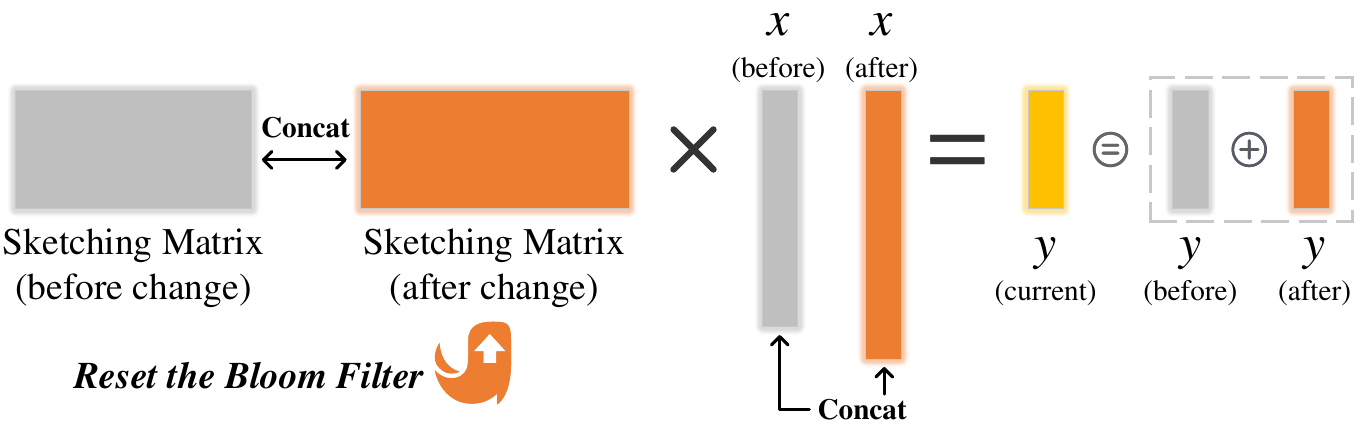}}
    \caption{Scalable UCL-sketch is robust to changes in its underlying sketch.}
    \label{change_hash}
    % \vspace{-0.45cm}
\end{figure}

In practice, sketch parameters are often updated, such as resizing memory or changing hash functions, which can render previously acquired knowledge invalid. We focus on the most challenging case (others are in the Appendix): switching the hash function midway through recording. A conventional sketch (e.g., CM-sketch) would lose earlier traces and quickly fail, whereas UCL-sketch adapts by re-recording post-change keys while retaining and marking the originals, effectively appending columns to the sketching matrix, as illustrated in Fig.~\ref{change_hash}. Online learning then enables rapid recovery despite the loss of prior knowledge.

% \vspace{-0.4cm}

\subsection{Putting It Together}

{We now put the basic design and our optimizations together, to build the final version of UCL-sketch. Fig.~\ref{final_version} gives overview of the complete version. Fig.~\ref{final_version} (a) shows the procedure of insertion process in data plane. Then in Fig.~\ref{final_version} (b) and (c), we construct the corresponding sketch sensing matrix $\boldsymbol{A}$ and present scalable inference of our bucket-wise network, respectively. To enable all buckets to share a single model, the model includes a bucket id $i$-indicator to guide the solver in selecting the corresponding logical bucket for estimation. As for training, since no real frequency is sent to the control plane, as illustrated in Fig.~\ref{final_version} (d), our goal (or loss function) is to reconstruct per-key frequencies $x$ conforms to three constraints: measurement, sparsity and invariant proportional property of Zipfian distribution. To query the frequency of keys, we sum up the results from the learned solver and the hash table as shown in Fig.~\ref{final_version} (e).

% \vspace{-0.3cm}

\section{Theoretical Analysis}
\label{sec:5}

In this section, we analyze the proposed UCL-sketch, including complexity, keys coverage, error bound, and requirement for unbiased estimation during training. Due to space constraints, we only list the conclusions and remarks here. Detailed proofs can be found in Appendix. For convenience, we list the notations frequently used in this part in Table~\ref{tab:notations}.

% \vspace{-0.3cm}

 \begin{table}[ht]
\centering
\caption{Major notations used in our analysis}
% \vspace{-0.2cm}
\label{tab:notations}
\begin{tabular}{|c|l|}
\hline
\textbf{Notation} & \textbf{Meaning} \\
\hline
\addlinespace[2pt] % Add space between the first and second row
\hline
$e$ & base of the natural logarithm \\
\hline
$F$ & total frequencies stored in the entail system \\
\hline
$f$ & filtered frequencies stored in the sketch part\\
\hline
$N$ & total number of distinct items \\
\hline
$K$ & total number of current distinct items \\
\hline
$s$ & number of slots in HF \\
\hline
$b$ & number of flag bits for each pair in HF \\
\hline
$m_b$ & number of bits in BF \\
\hline
$(\varepsilon_b,\delta_b)$ & coefficient and error probability in BF \\
\hline
$(\varepsilon_c,\delta _c)$ & coefficient and error probability in sketch \\
\hline
$w$ & $\left\lceil {{e \mathord{\left/ {\vphantom {e {{\varepsilon _c}}}} \right. \kern-\nulldelimiterspace} {{\varepsilon _c}}}} \right\rceil$, sketch array width \\
\hline
$d$ & $\left\lceil {\ln \left( {{1 \mathord{\left/{\vphantom {1 {{\delta _c}}}} \right.\kern-\nulldelimiterspace} {{\delta _c}}}} \right)} \right\rceil$, sketch array depth \\
\hline
$k_b$ & $\log \left( {{1 \mathord{\left/ {\vphantom {1 {{\varepsilon _b}{\delta _b}}}} \right. \kern-\nulldelimiterspace} {{\varepsilon _b}{\delta _b}}}} \right)$, number of hash functions in BF \\
\hline
$\boldsymbol{A}$ & sketching matrix of a linear sketch, i.e., CM-sketch \\
\hline
$x$ & ground-truth frequency data vector \\
\hline
$y$ & sketch counters measurement vector \\
\hline
$\sigma_s$ & $s$-restricted isometry constant of $\boldsymbol{A}$ \\
\hline
$x_T$ & vector satisfying $\left\{ x\left( i \right): i \in T; 0: i \notin T \right\}$ \\
\hline
$(\cdot)^c$ & complement of a set $(\cdot)$ \\
\hline
$T_{p_{(\cdot)}}$ & matrix form of a transformation operation $p_{(\cdot)}$ \\
\hline
\end{tabular}
\end{table}

\textbf{Definition 1}\label{def} \textit{(s-restricted Isometry Constant~\cite{candes2005decoding}). For every integer $s = 1,2, \ldots $, we define the s-restricted isometry constants $\sigma_s$ of a matrix $\boldsymbol{A}$ as the smallest quantity such that}
\begin{equation}
    \left( {1 - {\sigma _s}} \right)\left\| x \right\|_2^2 \le \left\| {\boldsymbol{A} x} \right\|_2^2 \le \left( {1 + {\sigma _s}} \right)\left\| x \right\|_2^2
    \label{RIC}
\end{equation}
\textit{for all $s$-sparse vectors, where a vector is said to be $s$-sparse if it has at most $s$ nonzero entries.}

Lemma~\hyperref[lemma1]{1} shows the complexities of memory space, and update time of UCL-sketch.

\textbf{Lemma 1}. \label{lemma1} \textit{The space complexity of UCL-sketch is ${\mathcal O}\left( m_b + \frac{e}{{{\varepsilon _c}}}\ln \frac{1}{{{\delta _c}}} + bs \right)$, and the time complexity of update operation is ${\mathcal O}\left( {\log \frac{1}{{{\varepsilon _b}{\delta _b}}} + \ln \frac{1}{{{\delta _c}}}} \right)$ in the data plane.}

Lemma~\hyperref[lemma2]{2} guarantees the error bound for missing keys\footnote{For these missing keys, we report zeros as their volumes, as truncating low-frequency estimates to 0 will not impact the system's overall performance~\cite{aamand2024improved}.} in the recovery phase of UCL-sketch.

\textbf{Lemma 2}. \label{lemma2} \textit{The keys coverage of the Bloom Filter in UCL-sketch obeys}
\begin{equation}
    \Pr \left(Y \ge K_y \right) \le \frac{K}{K_y}{\left( {1 - {e^{- \frac{{k_b}{K}}{m_b}}}} \right)},
    % \label{coverage}
\end{equation}
\textit{where variable $Y$ denotes the number of keys that are not covered but viewed as covered.}

We show UCL-sketch's worst-case error bound of per-key recovery (without equivalent loss) from sketch counters as shown in Theorem~\hyperref[theorem2]{2}.

\textbf{Theorem 2}. \label{theorem2} \textit{Let $f=\left(f(1), f(2), \dots, f(n)\right)$ be the real volume vector of a stream that is stored in the sketch, where $f(i)$ denotes the volume of $i$-th distinct item. Consider $T_0$ as the locations of the $s$ largest volume of $f$. Assume that the reported volume vector $f^*$ is the optimal solution that minimizes the first two objective in Eqn.~\ref{objective}. Then the worst-case frequency estimation error is bounded by}
\begin{equation}
    {\left\| {{f^*} - f} \right\|_1} \le \frac{{ {{2 + \left( {2\sqrt 2  + 2} \right){\sigma _{2s}}}}}}{{1 - \left( {\sqrt 2  + 1} \right){\sigma _{2s}}}}{\left\| {{f_{T_0^c}}} \right\|_1}
    % \label{CS_Bound}
\end{equation}

\textbf{\textit{Remark}}. We list theoretical comparison between UCL-sketch and six methods in Table~\ref{anlysis_table}. Although some classic sketches have slightly lighter structure than ours, they obtain more inaccurate error bound. Specifically, it is worthy noting that $\frac{1}{K} \left\|{{f_{T_0^c}}} \right\|_1 \ll \varepsilon _c \left\|{{f}} \right\|_1$ in most scenarios due to heavy-tailed (sparse) property of real-world streams. It has also been proven in~\cite{yang2018elastic} that the bound of Elastic Sketch is lower than that of C-sketch and CM-sketch, e.g., $\varepsilon _c \left\|{{f}} \right\|_1 \le \varepsilon _c \left\|{{F}} \right\|_1 \le \varepsilon _c \left\|{{F}} \right\|_2$. Therefore, our equation-based algorithm significantly outperforms these three competitors regarding estimation accuracy. Meanwhile, 
Univmon requires $\log N$ times the space in the update phase. Note that the effect of equivalent learning was not considered in our analysis. Consequently, the practical performance achieved by our algorithms may surpass the theoretical result, which has been substantiated by the empirical results presented in Section~\ref{sec:6}.
% \vspace{-0.3cm}
\begin{table}[htbp]
  \centering
  \caption{Theoretical comparison with state-of-the-art sketches}
  % \vspace{-0.2cm}
  \begin{threeparttable}
    \resizebox{0.5\textwidth}{!}{\begin{tabular}{c|c|c|c|c}
    \toprule
    \textbf{Algorithm} & \textbf{Space Complexity} & \textbf{Time Complexity} & \textbf{Average Error} & \textbf{Prob.} \\
    \midrule
    CM-sketch & ${\mathcal{O}}\left(\frac{e}{{{\varepsilon _c}}}\ln \frac{1}{{{\delta _c}}}\right)$ & ${\mathcal{O}}\left(\ln \frac{1}{{{\delta _c}}}\right)$ & ${\mathcal{O}}\left( \varepsilon _c \left\|{{F}} \right\|_1 \right)$ & $1-\delta _c$ \\
    C-sketch & ${\mathcal{O}}\left(\frac{e}{{{\varepsilon^2 _c}}}\ln \frac{1}{{{\delta _c}}}\right)$ & ${\mathcal{O}}\left(\ln \frac{1}{{{\delta _c}}}\right)$ & ${\mathcal{O}}\left( \varepsilon _c \left\|{{F}} \right\|_2 \right)$ & $1-\delta _c$ \\
    Elastic Sketch & ${\mathcal{O}}\left(\frac{-s\log K}{\ln (1-{{\varepsilon _b}{\delta _b}})} + \frac{e}{{{\varepsilon^2 _c}}}\ln \frac{1}{{{\delta _c}}}\right)$ & ${\mathcal O}\left(\ln \frac{1}{{{\delta _c}}}\right)$ & ${\mathcal{O}}\left(\varepsilon _c \left\|{{f}} \right\|_1 \right)$ & $1-\delta _c$ \\
    Univmon & ${\mathcal O}\left(\log N (\frac{e}{{{\varepsilon _c}}}\ln \frac{1}{{{\delta _c}}} + s \log N)\right)$ & ${\mathcal O}\left(\log N \ln \frac{s}{{{\delta _c}}}\right)$ & / & / \\
    \midrule
    UCL-sketch & ${\mathcal{O}}\left( m_b + \frac{e}{{{\varepsilon _c}}}\ln \frac{1}{{{\delta _c}}} + bs \right)$ & ${\mathcal{O}}\left( {\log \frac{1}{{{\varepsilon _b}{\delta _b}}} + \ln \frac{1}{{{\delta _c}}}} \right)$ & ${\mathcal{O}}\left( \frac{1}{K} \left\| {{f_{T_0^c}}} \right\|_1 \right)$ & $\approx$ 1 \\
    \bottomrule
    \end{tabular}}
    \begin{tablenotes}
        \footnotesize
        \item {In the table, complexity analyses of all baselines are from~\cite{sheng2021pr}.}
    \end{tablenotes}
    \end{threeparttable}
  \label{anlysis_table}
  % \vspace{-0.3cm}
\end{table}

Theorem~\hyperref[theorem3]{3} gives the necessary condition of UCL-sketch to achieve full accuracy in the training set with our GT-free equivalent learning strategy.

\textbf{Theorem 3}. \label{theorem3} \textit{A necessary condition for recovering the true volume from compressed counters is that the following linear system has a unique solution:}
\begin{equation}
    \boldsymbol{B}x = \left( {\begin{array}{*{20}{c}}
\boldsymbol{A}\\
{\boldsymbol{A}{T_{{p_1}}}}\\
 \vdots \\
{\boldsymbol{A}{T_{{p_{\left| \mathcal{P} \right|}}}}}
\end{array}} \right)x = \left( {\begin{array}{*{20}{c}}
y\\
{{y^{{1}}}}\\
 \vdots \\
{{y^{{{\left| \mathcal{P} \right|}}}}}
\end{array}} \right),
\end{equation}
where $y^{\left(\cdot\right)}$ is the measurement corresponding to the transformation $p_{\left(\cdot\right)}$. Rigorously, rank$\left(\boldsymbol{B}\right) = n$.

\textbf{\textit{Remark}}. Due to the set invariance of zipfian streaming data, {Theorem~\hyperref[theorem3]{3} states the requirement to learn a deep solver without ground truth, is that numerous virtual sketch operators $\{{\boldsymbol{A}{T_{{p_{(i)}}}}}\}_{i=1,2,\dots,|\mathcal{P}|}$ have enough different range space to determine unique per-key frequencies $x$. That means the choice of transformation group $\mathcal{P}$ in above system is not arbitrary, but needs to be rank $n$ or at least $> m$ so that the model is guaranteed to learn from the null space of $\boldsymbol{A}$. Critically though, much room beyond $\boldsymbol{A}^\dag$ will be filled after a large number of random transformations during training. Thus, it is possible to train the solver from only sketch counters using our scheme, but also note that the group $\mathcal{P}$ might in some cases not be sufficient since the target $x$ is rapidly and continuously extending in complex streams.}

In Theorem~\hyperref[theorem4]{4}, we give a unified MAP estimate result of all keys given the information data structure, which helps to explain why the statistical property of sketch are similar over different keys and buckets, creating the probability of sharing bucket-wise estimations in our deep solver.

\textbf{Theorem 4}. \label{theorem4} \textit{Assume that the prior of each $k_i$-associated frequency in per-key GT vector $x$ satisfies $\Pr \left( x(i) = a \right) = \mathcal{N}\left( a; \mu_i, \sigma_i^2 \right)$, then for $\forall k_i$ and ${\left\| x \right\|}_1 \rightarrow \infty$, through asymptotic approximation, its posterior estimate based on sketch counters $y$ has the following general form:}
\begin{equation}
    \hat x(i) = \frac{{{\mu _i}\left\| x \right\|_2^2 + \sigma _i^2\left( {w\sum\nolimits_{j = 1}^d {{y_{j \times {H_j}\left( {{k_i}} \right)}} - d{{\left\| x \right\|}_1}} } \right)}}{{\left\| x \right\|_2^2 + \sigma _i^2d(w - 1)}}
    \label{map}
\end{equation}
\textit{where $H_j$ is the indep. hash function: $\boldsymbol{\Omega} \rightarrow \{1,\dots,w\}$.}

\textbf{\textit{Remark}}. Theorem~\hyperref[theorem4]{4} can be easily extended to the estimation of combinations involving multiple keys (i.e., bucket). From Eqn.~\ref{map}, we see that once the system input (i.e., $y$ and $\boldsymbol{A}$) is provided, all parameters are shared and fixed, except for $(\mu_i, \sigma_i)$ which is unknown for particular keys. Alternatively, approaches using neural network are able to consist of jointly learning different estimation buckets with a $i$-indicator. While the UCL-sketch can thus be viewed as a number of approximations, our empirical evaluation in Section~\ref{sec:6} shows that the UCL-sketch yields highly accurate estimates, confirming the validity of the shared logical buckets.

% \vspace{-0.3cm}

\section{Experiments}
\label{sec:6}

We next report key results compared to state-of-the-art methods with both real-world and synthetic steam datasets. 

% \vspace{-0.3cm}

\subsection{Experimental Setup} \label{begin}

\textbf{Datasets.}
We conduct experiments on three real-world datasets. The first is \textit{CAIDA}~\cite{caida}, real traffic data collected on a backbone link between Chicago and Seattle in 2018. We form 13-byte keys with five fields: source and destination IP addresses, source and destination ports, and protocol. In our experiments, we use 1 million packets of it with around 100K distinct keys. The second is \textit{Kosarak}~\cite{kosarak}, consists of anonymized click-stream data from a Hungarian online news portal. We also extract a segment of data with a length of 1 million for our experiments, where about 25K unique keys are in it. The third is \textit{Retail}~\cite{brijs1999using}, which contains retail market basket data supplied by an anonymous Belgian retail supermarket store. There are nearly 910K items in this stream, and we utilize the entire dataset comprising a total of 16K unique keys. Each key is 4-byte long in the above two datasets. To evaluate the robustness of the proposed algorithm, we also synthesize four datasets that satisfy \textit{Zipf}’s law, where the skewness varies in [1.2, 1.3, 1.4, 1.5] and keys with length of 4-byte distinguish items in these datasets. There are 2M elements in each dataset, with around 22K $\sim$ 214K total distinct items depending on the skewness.

\textbf{Metrics.}
For comparing the accuracy of frequency estimations, we leverage \textit{Average Absolute Error (AAE)} and \textit{Average Relative Error (ARE)}. Additionally, we use \textit{Weighted Mean Relative Difference (WMRD)} and \textit{Entropy Relative Error} to evaluate the accuracy of the per-key distribution. Formally, the detailed descriptions are given below:

$(1)$ {\textit{AAE}}: It equals $\frac{1}{n}\sum\nolimits_{i = 1}^n {\left| {f(i) - {f^*}(i)} \right|}$, where $f(\cdot)$ and $f^*(\cdot)$ are real and estimated frequency respectively. 

$(2)$ {\textit{ARE}}: It equals $\frac{1}{n}\sum\nolimits_{i = 1}^n {\frac{{\left| {f(i) - {f^*}(i)} \right|}}{{f(i)}}}$, where $f(\cdot)$ and $f^*(\cdot)$ are the same as those defined above. 

$(3)$ {\textit{WMRD}}~\cite{sheng2021pr}: It can be written as $\frac{{\sum\nolimits_{i = 1}^z {\left| {n(i) - {n^*}(i)} \right|} }}{{\sum\nolimits_{i = 1}^z {\frac{{n(i) + {n^*}(i)}}{2}} }}$, where $z$ is the maximum single-key frequency, and $n(i)$ and $n^*(i)$ are the real and estimated number of keys with frequency $i$ respectively. 

$(4)$ {\textit{Entropy Absolute Error}}: We calculate the entropy $e$ based on a frequency set as $- \sum\nolimits_{i = 1}^z {\left( {i \times \frac{{n(i)}}{{\sum\nolimits_{i = 1}^z {n(i)} }}{{\log }_2}\frac{{n(i)}}{{\sum\nolimits_{i = 1}^z {n(i)} }}} \right)}$, then the absolute error is ${\left| {e - {e^*}} \right|}$ where $e$ and $e^*$ are the true and estimated entropy. Here we define $0\log(0) = 0$.

\begin{figure}[htbp]
\centering
\subfigure[CAIDA Dataset]{
\includegraphics[width=1.\linewidth]{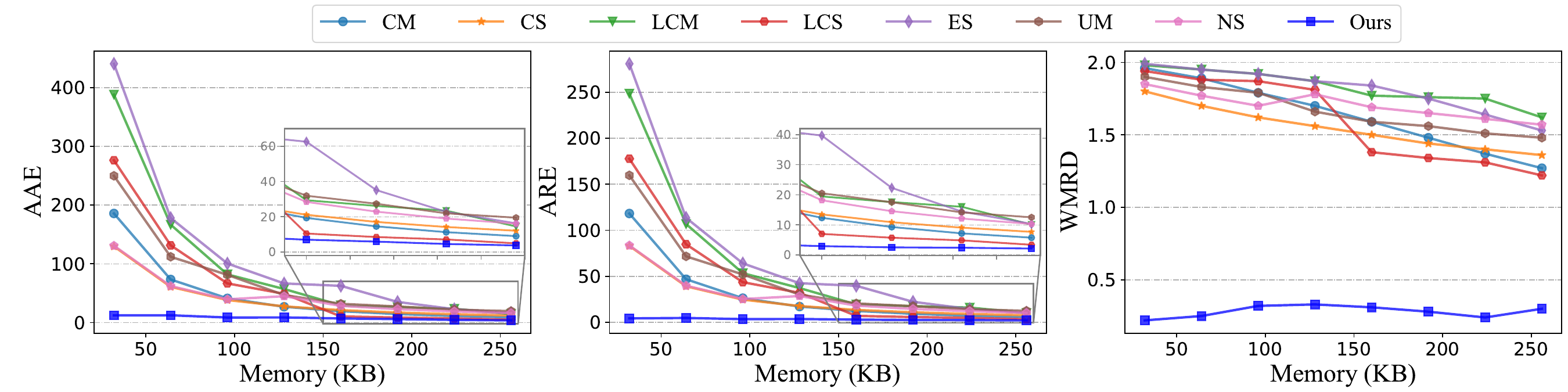}
}
\\
% \vspace{-0.2cm}
\subfigure[Kosarak Dataset]{
\includegraphics[width=1.\linewidth]{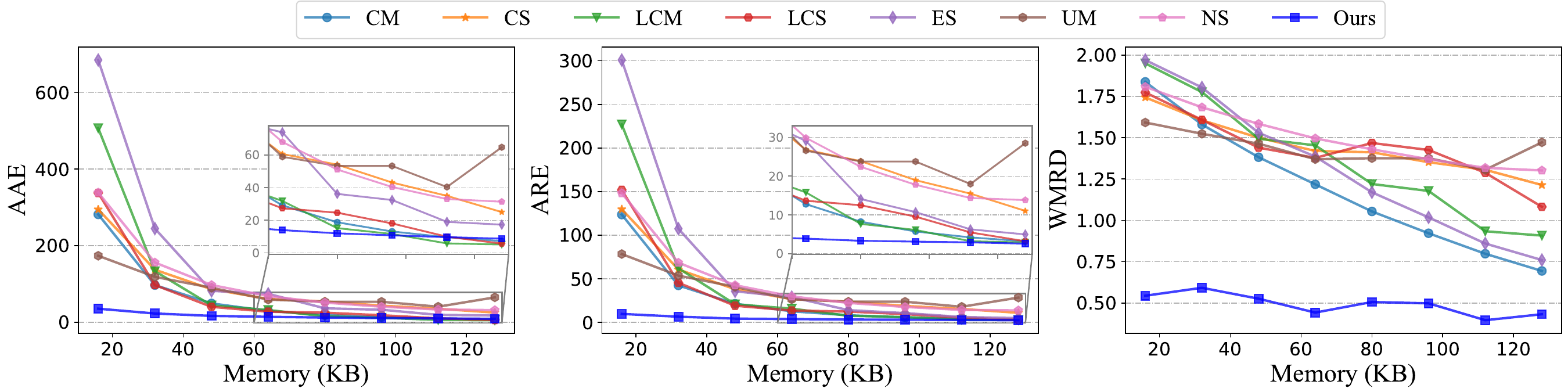}
}
\\
% \vspace{-0.2cm}
\subfigure[Retail Dataset]{
\includegraphics[width=1.\linewidth]{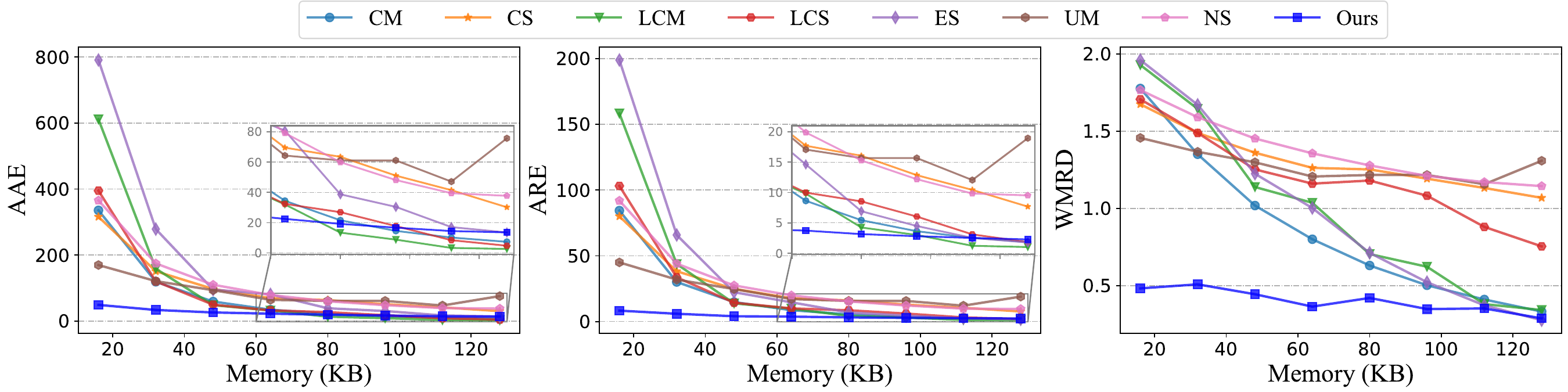}
}
\caption{Performance comparison between our UCL-sketch and existing state-of-the-art sketches.}
\label{performance}
% \vspace{-0.45cm}
\end{figure}

\textbf{Algorithm Comparisons.}
For comparison, we implement ten existing frequency estimation algorithms\footnote{The implementation of all sketches (in Python) can be found in our codebase, which is written mainly based on a C++ Github repository at \href{https://github.com/N2-Sys/BitSense/tree/main/simulator/src/sketch}{\texttt{https://github.com/N2-Sys/BitSense}}.}, such as \textit{CM-sketch (CM)}~\cite{cormode2005improved}, \textit{C-sketch (CS)}~\cite{charikar2002finding}, \textit{Elastic Sketch (ES)}~\cite{yang2018elastic}, \textit{UnivMon (UM)}~\cite{liu2016one}, \textit{Nitrosketch (NS)}~\cite{liu2019nitrosketch} and two ideally learned sketches~\cite{hsu2019learning}: \textit{Learned CM-sketch (LCM)} and \textit{Learned C-sketch (LCS)}}. In addition, we compare our UCL-sketch with equation-based sketches, i.e., \textit{FlowRadar}~\cite{li2016flowradar}, \textit{PR-sketch}~\cite{sheng2021pr} and \textit{SeqSketch}~\cite{huang2021toward}, in our ablational experiments (detailed in Appendix). We give them the same local memory of buckets $\&$ sketch as ours. To be more specific, the number of hash functions has been fixed at 4 across all methods. We fix levels as 2 in the universal sketch of UnivMon. For Elastic Sketch, we allocate 25$\% \sim$ 50$\%$ memory for its hash table. In particular, the neural network oracle in two learned sketches is replaced with an ideal oracle that knows the identities of the heavy hitters, whose target domain depends on the number of its unique buckets which take up around 50$\%$ memory. Also note that the memory of the learned oracle is not computed in the memory cost reported in this section. Finally, the detailed description of our parameter settings can be found in Appendix.

\begin{table}[htbp]
  \centering
  \caption{{Entropy absolute error on different streaming data sets \\ (\textbf{bold} indicates best performance)}}
  % \vspace{-0.2cm}
    \resizebox{0.5\textwidth}{!}{\begin{tabular}{c|c|c|c|c|c|c|c|c|c}
    \toprule
    Datasets & Memory & \cellcolor[rgb]{ .851,  .851,  .851}Ours & CM  & CS  & LCM  & LCS  & ES  & UM & NS \\
    \midrule
    \multirow{9}[16]{*}{\begin{sideways}CAIDA\end{sideways}} & 32KB & \cellcolor[rgb]{ .851,  .851,  .851}\textbf{9.68} & 1550.48 & 1252.46 & 2475.28 & 1042.73 & 3969.5 & 1196.32 & 510.67 \\
\cmidrule{2-10}          & 64KB & \cellcolor[rgb]{ .851,  .851,  .851}\textbf{7.54} & 543.85 & 550.84 & 806.84 & 380.76 & 1459.49 & 571.56 & 212.99 \\
\cmidrule{2-10}          & 96KB    & \cellcolor[rgb]{ .851,  .851,  .851}\textbf{4.51} & 284.80  & 385.78 & 394.74 & 341.08 & 774.28 & 429.74 & 131.79 \\
\cmidrule{2-10}          & 128KB   & \cellcolor[rgb]{ .851,  .851,  .851}\textbf{3.39} & 176.97 & 236.29 & 227.62 & 124.86 & 490.86 & 557.75 & 105.86 \\
\cmidrule{2-10}          & 160KB   & \cellcolor[rgb]{ .851,  .851,  .851}\textbf{3.12} & 122.43 & 185.47 & 168.27 & 71.54 & 455.09 & 291.04 & 100.44 \\
\cmidrule{2-10}          & 192KB    & \cellcolor[rgb]{ .851,  .851,  .851}\textbf{2.82} & 89.00 & 147.47 & 142.04 & 55.93 & 237.98 & 246.48 & 82.09 \\
\cmidrule{2-10}          & 224KB    & \cellcolor[rgb]{ .851,  .851,  .851}\textbf{2.90} & 67.45 & 120.76 & 121.82 & 45.29 & 146.81 & 194.93 & 68.26 \\
\cmidrule{2-10}          & 256KB    & \cellcolor[rgb]{ .851,  .851,  .851}\textbf{1.42} & 52.95 & 101.58 & 69.13 & 29.54 & 100.88 & 169.14 & 56.25 \\
    \midrule
    \multirow{9}[16]{*}{\begin{sideways}Kosarak\end{sideways}} & 16KB & \cellcolor[rgb]{ .851,  .851,  .851}\textbf{58.24} & 2620.64 & 3270.70 & 4506.12 & 3469.50 & 6537.66 & 1875.06 & 3575.26  \\
\cmidrule{2-10}          & 32KB    & \cellcolor[rgb]{ .851,  .851,  .851}\textbf{35.10} & 841.50 & 1438.20 & 1008.47 & 846.16 & 2268.03 & 1222.37 & 1532.43 \\
\cmidrule{2-10}          & 48KB    & \cellcolor[rgb]{ .851,  .851,  .851}\textbf{21.63} & 404.67 & 876.32 & 295.55 & 317.96 & 703.97 & 902.99 & 914.97 \\
\cmidrule{2-10}          & 64KB   & \cellcolor[rgb]{ .851,  .851,  .851}\textbf{16.85} & 231.51 & 608.17 & 196.10 & 202.46 & 643.72 & 565.21 & 625.38 \\
\cmidrule{2-10}          & 80KB   & \cellcolor[rgb]{ .851,  .851,  .851}\textbf{14.29} & 144.01 & 542.87 & 86.35 & 173.80 & 299.46 & 524.43 & 462.48 \\
\cmidrule{2-10}          & 96KB    & \cellcolor[rgb]{ .851,  .851,  .851}\textbf{13.19} & 98.37 & 430.44 & 60.62 & 119.25 & 274.27 & 483.68 & 361.92 \\
\cmidrule{2-10}          & 112KB    & \cellcolor[rgb]{ .851,  .851,  .851}\textbf{10.79} & 70.56 & 347.66 & 27.47 & 61.44 & 155.77 & 393.32 & 294.88 \\
\cmidrule{2-10}          & 128KB    & \cellcolor[rgb]{ .851,  .851,  .851}\textbf{10.18} & 52.63 & 247.36 & 23.44 & 34.50 & 144.44 & 654.84 & 288.81 \\
    \midrule
    \multirow{9}[15]{*}{\begin{sideways}Retail\end{sideways}} & 16KB & \cellcolor[rgb]{ .851,  .851,  .851}\textbf{69.16} & 3166.01 & 3468.11 & 5527.20 & 4129.77 & 7601.40 & 1766.46 & 3857.67 \\
\cmidrule{2-10}          & 32KB    & \cellcolor[rgb]{ .851,  .851,  .851}\textbf{41.02} & 1066.05 & 1569.19 & 1265.28 & 1097.03 & 2621.57 & 1197.95 & 1734.13 \\
\cmidrule{2-10}          & 48KB    & \cellcolor[rgb]{ .851,  .851,  .851}\textbf{27.14} & 511.37  & 984.25 & 346.69 & 415.05 & 850.06 & 894.09 & 1051.95 \\
\cmidrule{2-10}          & 64KB   & \cellcolor[rgb]{ .851,  .851,  .851}\textbf{22.20} & 291.21 & 690.76 & 211.31 & 254.86 & 742.85 & 582.79 & 740.24 \\
\cmidrule{2-10}          & 80KB   & \cellcolor[rgb]{ .851,  .851,  .851}\textbf{18.87} & 179.20 & 643.02 & 83.16 & 208.78 & 344.02 & 591.83 & 554.35 \\
\cmidrule{2-10}          & 96KB    & \cellcolor[rgb]{ .851,  .851,  .851}\textbf{15.07} & 119.71 & 516.00 & 50.66 & 130.63 & 282.78 & 494.88 & 441.95 \\
\cmidrule{2-10}          & 112KB    & \cellcolor[rgb]{ .851,  .851,  .851}\textbf{12.56} & 82.79 & 419.62 & 17.72 & 60.03 & 156.78 & 448.04 & 362.26 \\
\cmidrule{2-10}          & 128KB    & \cellcolor[rgb]{ .851,  .851,  .851}\textbf{10.56} & 59.45 & 305.66 & 13.67 & 32.82 & 129.85 & 773.58 & 358.09 \\
\bottomrule
    \end{tabular}}
  \label{entropy}
  \vspace{-0.3cm}
\end{table}

\begin{figure}[htbp]
\centering
\subfigure[Skewness = 1.2]{
\includegraphics[width=1.\linewidth]{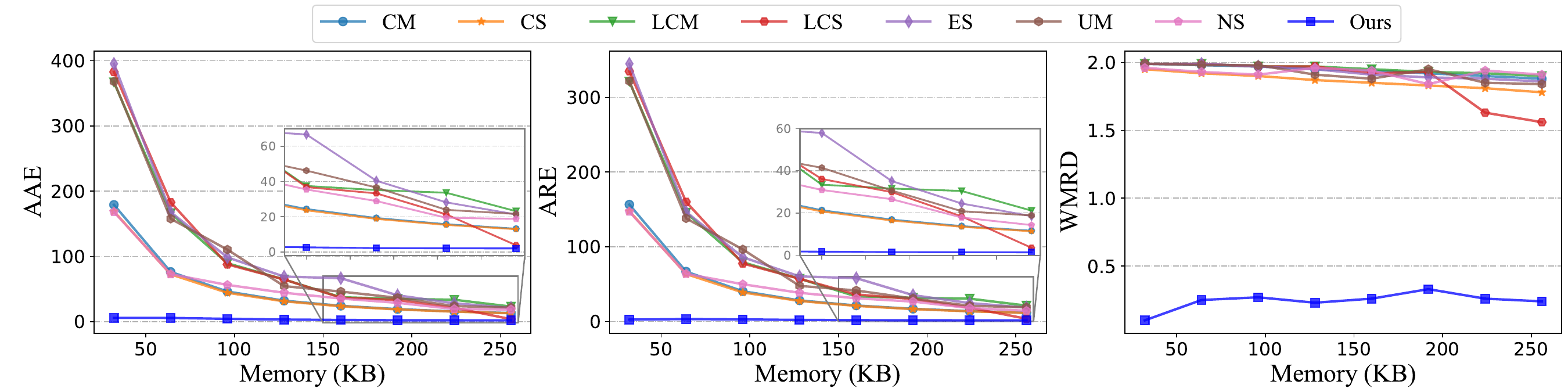}
}
\\
% \vspace{-0.2cm}
\subfigure[Skewness = 1.3]{
\includegraphics[width=1.\linewidth]{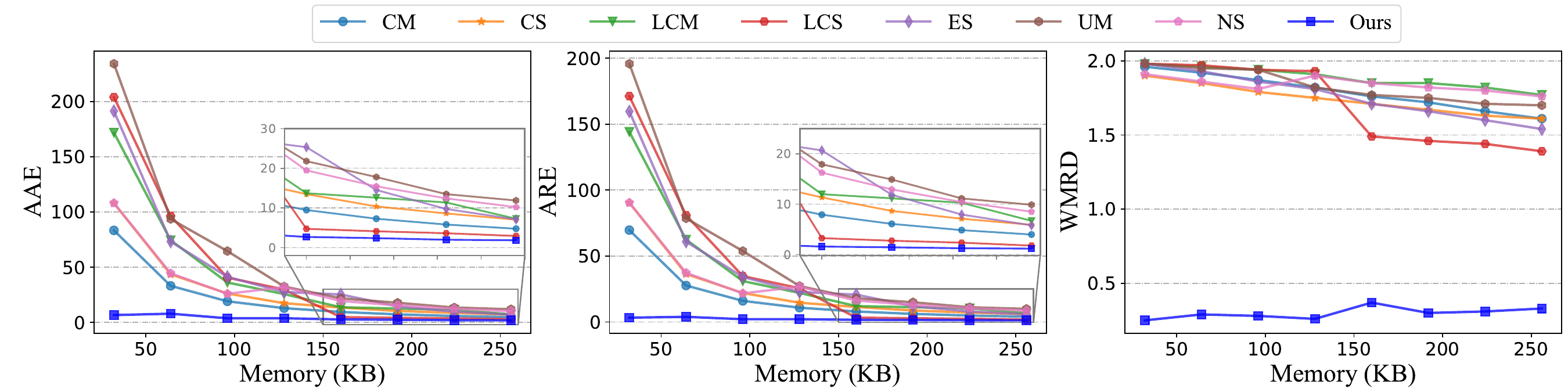}
}
\\
% \vspace{-0.2cm}
\subfigure[Skewness = 1.4]{
\includegraphics[width=1.\linewidth]{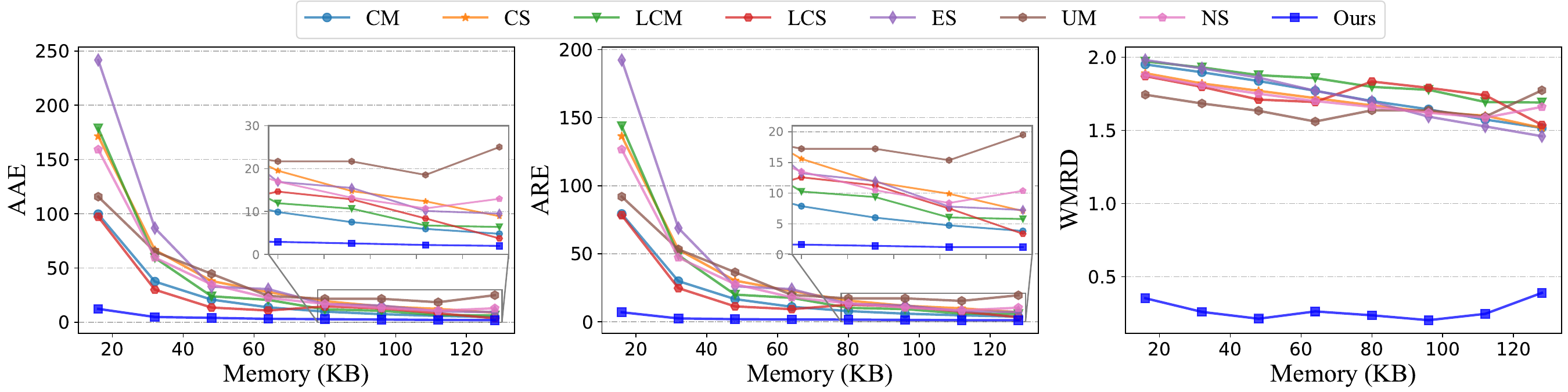}
}
\\
% \vspace{-0.2cm}
\subfigure[Skewness = 1.5]{
\includegraphics[width=1.\linewidth]{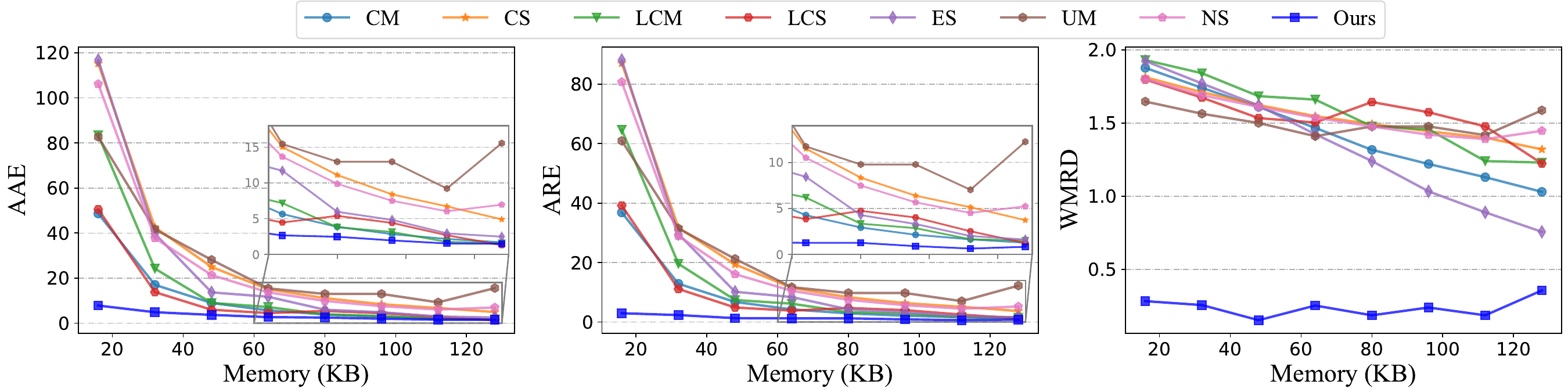}
}
\caption{Performance comparison between our UCL-sketch and existing state-of-the-art sketches on synthetic Zipfian datasets with different skewness.}
\label{performance_zipf}
\vspace{-0.3cm}
\end{figure}

\begin{table}[htbp]
  \centering
  \caption{Entropy absolute error on different Zipfian datasets \\ (\textbf{bold} indicates best performance)}
  % \vspace{-0.2cm}
    \resizebox{0.5\textwidth}{!}{\begin{tabular}{c|c|c|c|c|c|c|c|c|c}
    \toprule
    Skewness & Memory & \cellcolor[rgb]{ .851,  .851,  .851}Ours & CM  & CS  & LCM  & LCS  & ES  & UM & NS \\
    \midrule
    \multirow{9}[16]{*}{1.2} & 32KB & \cellcolor[rgb]{ .851,  .851,  .851}\textbf{6.23} & 1354.36  & 1869.68 & 2290.03 & 1008.36 & 3369.08 & 1405.47 & 1109.91 \\
\cmidrule{2-10}          & 64KB & \cellcolor[rgb]{ .851,  .851,  .851}\textbf{2.87} & 502.69 & 724.66 & 830.31 & 369.63 & 1259.31 & 556.77 & 433.10 \\
\cmidrule{2-10}          & 96KB    & \cellcolor[rgb]{ .851,  .851,  .851}\textbf{3.56} & 277.20 & 431.75 & 448.20 & 497.78 & 675.75 & 537.01 & 250.64  \\
\cmidrule{2-10}          & 128KB   & \cellcolor[rgb]{ .851,  .851,  .851}\textbf{3.01} & 180.74 & 242.73 & 288.44 & 133.06 & 443.37 & 694.09 & 263.25 \\
\cmidrule{2-10}          & 160KB   & \cellcolor[rgb]{ .851,  .851,  .851}\textbf{3.74} & 129.94 & 206.26 & 185.51 & 94.36 & 428.60 & 330.40 & 226.99 \\
\cmidrule{2-10}          & 192KB    & \cellcolor[rgb]{ .851,  .851,  .851}\textbf{2.90} & 98.99 & 158.24 & 167.02 & 193.81 & 237.88 & 401.55 & 116.56 \\
\cmidrule{2-10}          & 224KB    & \cellcolor[rgb]{ .851,  .851,  .851}\textbf{2.70} & 78.23 & 126.25 & 156.93 & 78.60 & 156.00 & 206.78 & 175.68 \\
\cmidrule{2-10}          & 256KB    & \cellcolor[rgb]{ .851,  .851,  .851}\textbf{1.53} & 63.73 & 103.69 & 100.37 & 42.89 & 112.46 & 185.67 & 117.76 \\
    \midrule
    \multirow{9}[16]{*}{1.3} & 32KB & \cellcolor[rgb]{ .851,  .851,  .851}\textbf{3.26} & 592.36 & 1192.73 & 913.99 & 476.62 & 1569.04 & 1047.64 & 816.45 \\
\cmidrule{2-10}          & 64KB & \cellcolor[rgb]{ .851,  .851,  .851}\textbf{1.41} & 201.90 & 435.70 & 294.59 & 159.95 & 509.99 & 391.56 & 296.26 \\
\cmidrule{2-10}          & 96KB    & \cellcolor[rgb]{ .851,  .851,  .851}\textbf{2.29} & 105.46 & 248.73 & 147.04 & 192.65 & 263.63 & 373.41 & 167.69 \\
\cmidrule{2-10}          & 128KB   & \cellcolor[rgb]{ .851,  .851,  .851}\textbf{1.60} & 66.54 & 141.37 & 88.00 & 52.50 & 164.57 & 416.50 & 167.98 \\
\cmidrule{2-10}          & 160KB   & \cellcolor[rgb]{ .851,  .851,  .851}\textbf{2.13} & 46.66 & 120.21 & 57.10 & 21.95 & 154.25 & 206.73 & 145.24 \\
\cmidrule{2-10}          & 192KB    & \cellcolor[rgb]{ .851,  .851,  .851}\textbf{1.76} & 34.75 & 89.10 & 50.18 & 17.72 & 80.89 & 161.93 & 115.41 \\
\cmidrule{2-10}          & 224KB    & \cellcolor[rgb]{ .851,  .851,  .851}\textbf{1.46} & 26.97 & 73.72 & 43.61 & 14.52 & 51.25 & 118.25 & 88.42 \\
\cmidrule{2-10}          & 256KB    & \cellcolor[rgb]{ .851,  .851,  .851}\textbf{0.81} & 21.50 & 58.37 & 25.51 & 10.30 & 35.84 & 102.04 & 70.00 \\
    \midrule
    \multirow{9}[15]{*}{1.4} & 16KB & \cellcolor[rgb]{ .851,  .851,  .851}\textbf{34.26} & 777.78 & 2015.95 & 1303.13 & 834.16 & 2106.11 & 1437.18 & 1528.59 \\
\cmidrule{2-10}          & 32KB    & \cellcolor[rgb]{ .851,  .851,  .851}\textbf{1.71} & 249.71 & 724.19  & 338.55 & 205.26 & 660.07 & 772.68 & 525.88 \\
\cmidrule{2-10}          & 48KB    & \cellcolor[rgb]{ .851,  .851,  .851}\textbf{3.65} & 126.36 & 394.16 & 116.05 & 83.62 & 213.24 & 517.90 & 278.60 \\
\cmidrule{2-10}          & 64KB   & \cellcolor[rgb]{ .851,  .851,  .851}\textbf{0.53} & 77.86 & 289.24  & 93.28 & 62.56 & 200.60 & 265.29 & 174.61 \\
\cmidrule{2-10}          & 80KB   & \cellcolor[rgb]{ .851,  .851,  .851}\textbf{0.30} & 52.84 & 188.41 & 48.16 & 76.59 & 100.88 & 216.73  & 131.69 \\
\cmidrule{2-10}          & 96KB    & \cellcolor[rgb]{ .851,  .851,  .851}\textbf{0.89} & 38.47 & 137.07 & 41.25 & 64.79 & 94.11 & 216.73 & 96.45 \\
\cmidrule{2-10}          & 112KB    & \cellcolor[rgb]{ .851,  .851,  .851}\textbf{1.26} & 29.64 & 115.87 & 23.69  & 39.95 & 58.06 & 195.21 & 81.31 \\
\cmidrule{2-10}          & 128KB    & \cellcolor[rgb]{ .851,  .851,  .851}\textbf{0.61} & 23.46 & 83.55 & 22.19 & 18.21 & 55.11 & 244.32 & 102.76 \\
\midrule
    \multirow{9}[15]{*}{1.5} & 16KB & \cellcolor[rgb]{ .851,  .851,  .851}\textbf{24.69} & 357.16 & 1335.87 & 554.13 & 398.21 & 966.71 & 991.13 & 993.50 \\
\cmidrule{2-10}          & 32KB    & \cellcolor[rgb]{ .851,  .851,  .851}\textbf{2.94} & 105.53 & 460.32 & 117.42 & 83.63 & 286.85 & 467.95 & 339.90 \\
\cmidrule{2-10}          & 48KB    & \cellcolor[rgb]{ .851,  .851,  .851}\textbf{4.86} & 50.26 & 264.98 & 36.07 & 32.43 & 83.11 & 309.66 & 179.28 \\
\cmidrule{2-10}          & 64KB   & \cellcolor[rgb]{ .851,  .851,  .851}\textbf{4.25} & 29.70 & 149.75 & 26.13 & 22.49 & 73.99 & 162.53 & 103.97 \\
\cmidrule{2-10}          & 80KB   & \cellcolor[rgb]{ .851,  .851,  .851}\textbf{1.71} & 19.52  & 107.31 & 12.23 & 24.34 & 34.8 & 129.13 & 75.11 \\
\cmidrule{2-10}          & 96KB    & \cellcolor[rgb]{ .851,  .851,  .851}\textbf{4.02} & 13.76 & 78.85 & 9.47 & 19.31 & 29.52 & 129.13 & 56.80 \\
\cmidrule{2-10}          & 112KB    & \cellcolor[rgb]{ .851,  .851,  .851}\textbf{4.46} & 10.51 & 62.80 & 4.73 & 11.19 & 17.22 & 89.86 & 46.64 \\
\cmidrule{2-10}          & 128KB    & \cellcolor[rgb]{ .851,  .851,  .851}\textbf{0.79} & 7.98 & 46.58 & 4.22 & 5.88 & 15.01 & 159.09 & 58.71 \\
    \bottomrule
    \end{tabular}}
  \label{entropy_zipf}
  \vspace{-0.3cm}
\end{table}

% \vspace{-0.3cm}

\subsection{Performance Comparison}\label{main_results}

\textbf{Estimation Results on Real-world Datasets.} First, we compare the AAE, ARE, and WMRD metrics for all seven sketches, by varying the local memory budgets from 32KB to 256KB on CAIDA dataset and 16KB to 128KB on other relatively smaller datasets. In the first two columns of Fig.~\ref{performance}, all methods achieve smaller AAE and ARE by increasing the space budget, whereas the UCL-sketch nearly always performs the best. In particular, when the memory cost is lower than 64KB, UCL-sketch achieves 6$\sim$20 times smaller error rate, compared to the best algorithm. On the stream data with much larger key space, i.e. CAIDA dataset, there is a more remarkable accuracy gap between our method and other sketches in all memory settings. For example, AREs of CM, CS, LCM, LCS, ES, UM, NS are 7.91 times, 6.98 times, 16.39 times, 11.13 times, 19.11 times, 12.67 times, and 8.11 times of that of ours on average, respectively. Despite the additional BF to keep tracking unique items, the reason is that existing algorithms cannot estimate per-key aggregations accurately without leveraging the linear system of sketching. 

Regarding frequency distribution alignment, we find that UCL-sketch still achieves better accuracy than the state-of-the-art sketches. As shown in the last column in Fig.~\ref{performance}, we observe that UCL-sketch maintains the WMRD value lower than 0.55 in all settings on three datasets. Unfortunately, such a stable performance does not persist in other competitors. We see that they are significantly less precise than UCL-sketch, although the metric drops as memory increases in general. Even worse, the baselines only achieve WMRDs over 1.5 with 128KB of memory on CAIDA, because the trace set contains too many keys to reliably measure the distribution of the stream with traditional point-wise methods, and their desired resources exceed the hardware capacity. Moreover, we list entropy relative errors of all estimation algorithms in Table~\ref{entropy}. As expected, similar trends to WMRD on the frequency entropy can be observed in the table, where our algorithm consistently achieves the smallest absolute error, substantially outperforming the second-best. Overall, UCL-sketch achieves both high performance in frequency query and distributional accuracy.

% \vspace{-0.3cm}

\textbf{Estimation Results on Toy Zipfian Datasets.} Fig.~\ref{performance_zipf} plots the average results regarding AAE, ARE, and WMRD, while frequency entropy relative errors are listed in Table~\ref{entropy_zipf}. The results confirm the conclusion shown in the real-world results: We observe that the performance of most methods increases as skewness rises. For example, in the case of skewness=1.2, a remark performance decline across all baselines. However, of particular interest is an inversely proportional trend observed in UCL-sketch's accuracy, as with the lowest skewness, the proposed algorithm can achieve even smaller deviation, e.g. average entropy absolute errors never exceeding 9 in that case, beating the other algorithms with a significant advantage. Upon further investigation, we have identified that this is because the heavy-tailed distribution leads to much fewer distinct keys given the same total count, then the average effect offsets and even reverses the growth of our errors. Besides, we find that UCL always achieves better accuracy than the state-of-the-art algorithm. When utilizing 64KB of memory, from Fig.~\ref{performance_zipf} (a), it becomes evident that UCL-sketch achieves a substantial reduction in AAE compared to the CM, CS, LCM, LCS, ES, UM and NS by factors of 10.94, 10.44, 22.48, 21.09, 24.77, 22.35 and 12.98, respectively. In terms of WMRD, our experimental results show that CM, CS, LCM, LCS, ES, UM and NS are 8.00, 7.67, 8.04, 7.71, 7.96, 7.88 and 7.92 times higher than that of the UCL-sketch in average.

% \vspace{-0.2cm}

\begin{figure}[htbp]
\setlength{\abovecaptionskip}{-0.01cm}
\setlength{\belowcaptionskip}{-0.2cm}
\centering
\subfigure[Stream Insertion Throughput]{
\includegraphics[width=0.465\linewidth]{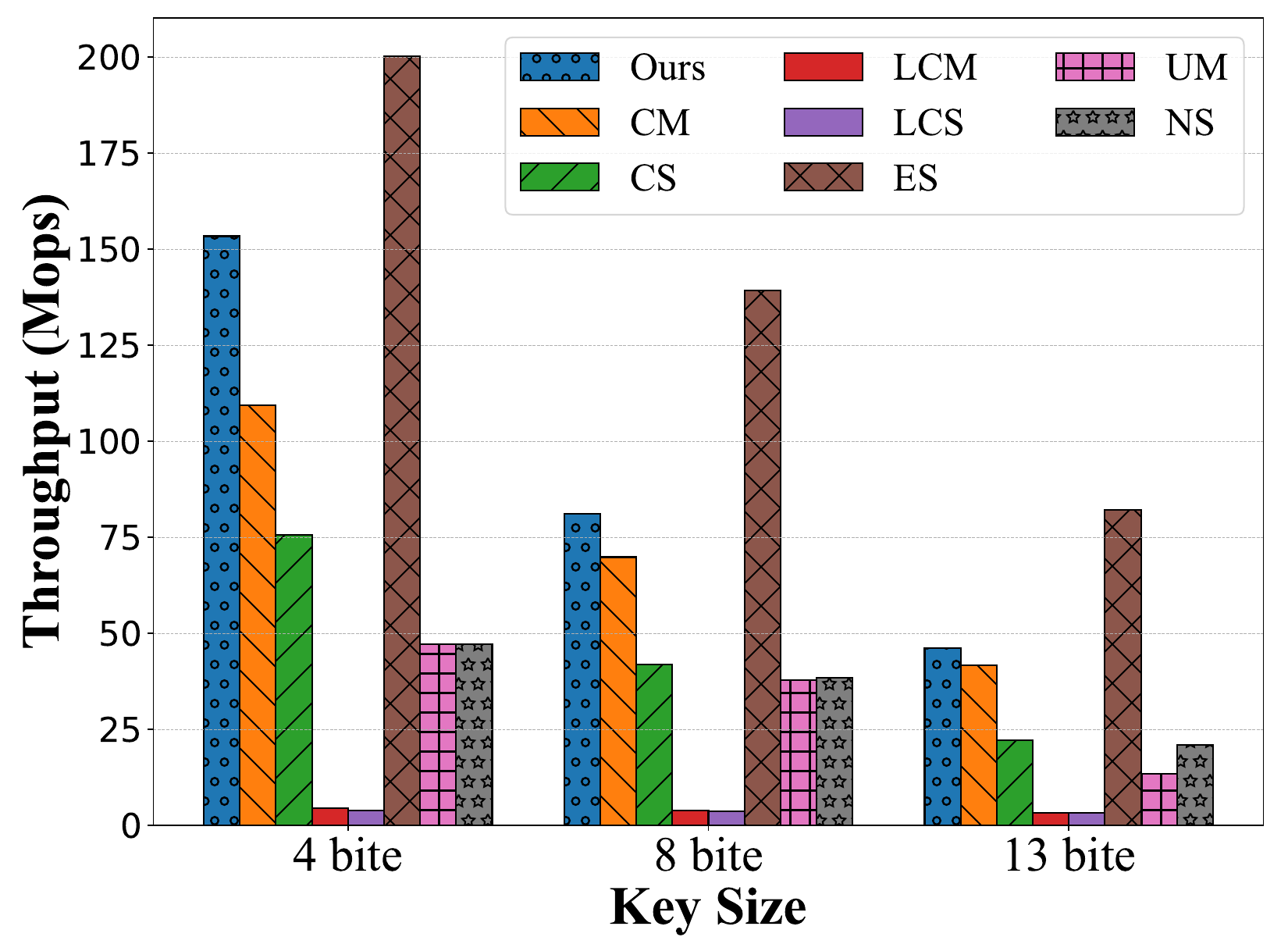}
}
% \\
% \vspace{-0.1cm}
\subfigure[Per-key Query Throughput]{
\includegraphics[width=0.465\linewidth]{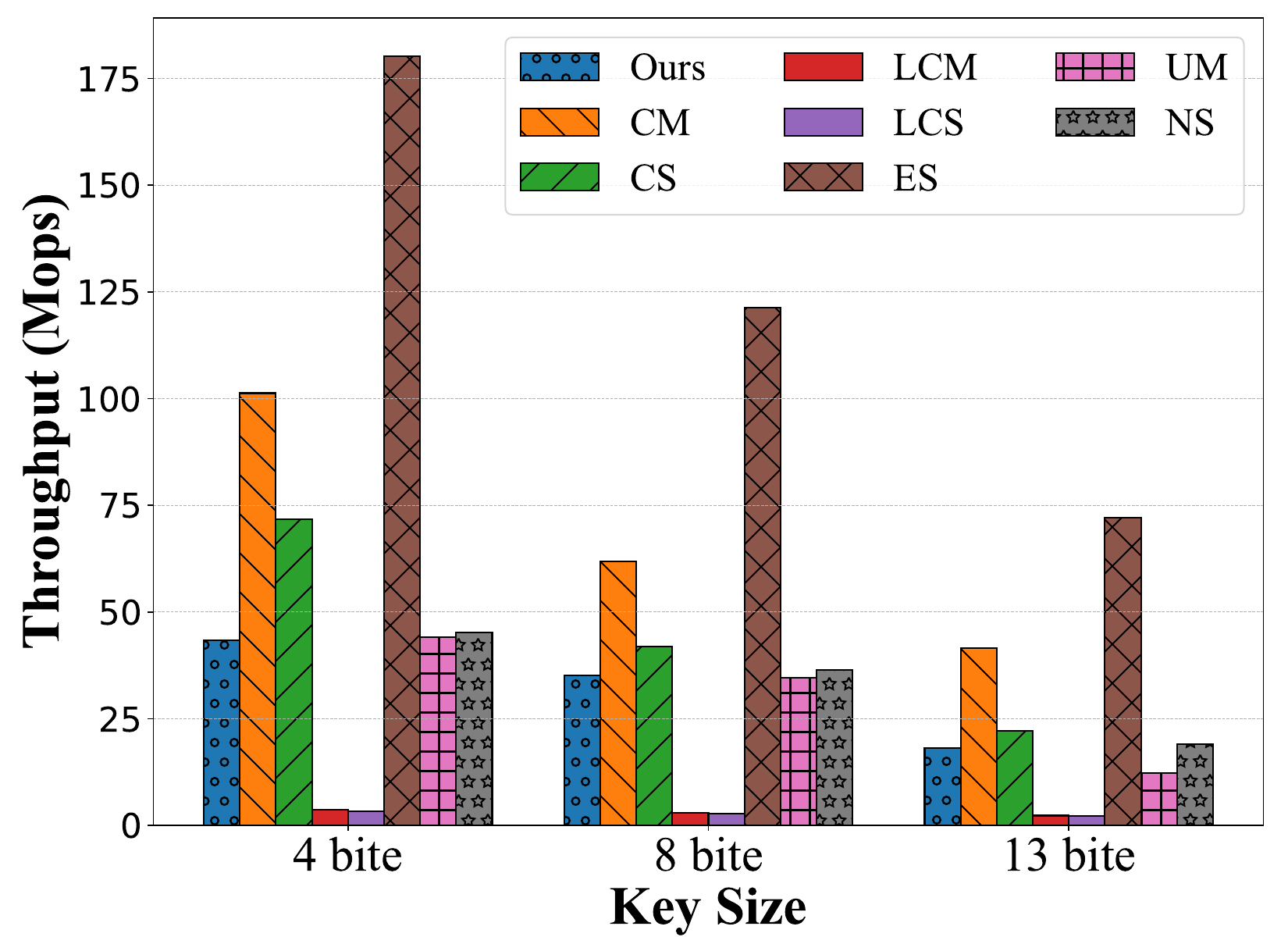}
}
\caption{Results on processing speed comparison.}
\label{throughput}
\vspace{-0.3cm}
\end{figure}

\textbf{Processing Speed.} We perform insertions of all items in a stream, record the total time used. We tested the C++ runtime of CM-sketch and proportionally scaled up all Python methods accordingly. We calculate million of operations per second (Mops) as throughput to measure the processing speed of various sketching algorithms. By using Kosarak with 64KB of memory, we evaluate the throughput of UCL-sketch and other solutions under different key sizes. Fig.~\ref{throughput} (a) provides the comparison results of insertion processing speed, in which we run a simple two-layer RNN model once before each update operation to simulate the actual practice of learning-augmented algorithms. From the figure, it becomes evident that previous learning-based sketch, i.e. LCM and LCS, fails to meet the requirements of processing high-speed data streams, since the propagation time overwhelms the insertion time, encumbering the efficiency of sketching. Meanwhile, ES achieves the highest throughput through its lightweight data structure. Fig.~\ref{throughput} also shows that UCL-sketch which performed as the second fastest can achieve almost 50$\times$ speed of learning-augmented sketch due to the model-free update and heavy filter in the data plane. 
The query processing throughput is shown in Fig.~\ref{throughput} (b). The query items are uniformly sampled from the incoming stream, which means in the skewed data stream, high-frequency items are queried more frequently than low-frequency ones. So, although the prediction time of neural networks exceeds that of hashing operations, our hash table is capable of answering most of the queries. As expected, Fig.~\ref{throughput} (b) shows that UCL-sketch still maintains performance comparable to that of the Nitrosketch and slightly inferior to C-sketch. Given the consistently slow processing speed of existing learning-augmented sketches, our results attest the practicality of UCL-sketch.

\begin{figure}
    \subfigure[Original loss curve\label{fig:loss}]{
        \centering
        \includegraphics[width=0.465\linewidth]{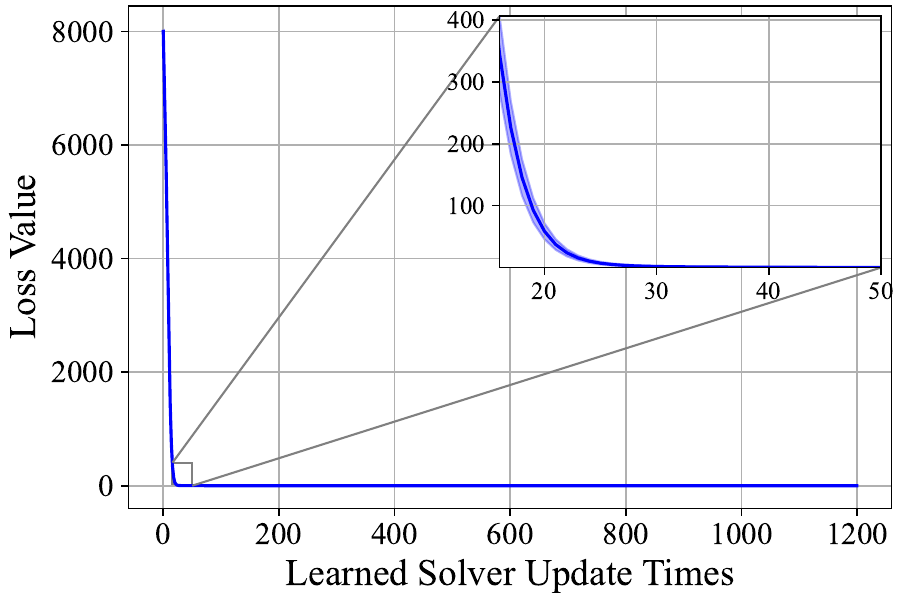}
    }
    \hfill
    \subfigure[Log-transformed loss curve\label{fig:log_loss}]{
        \centering
        \includegraphics[width=0.465\linewidth]{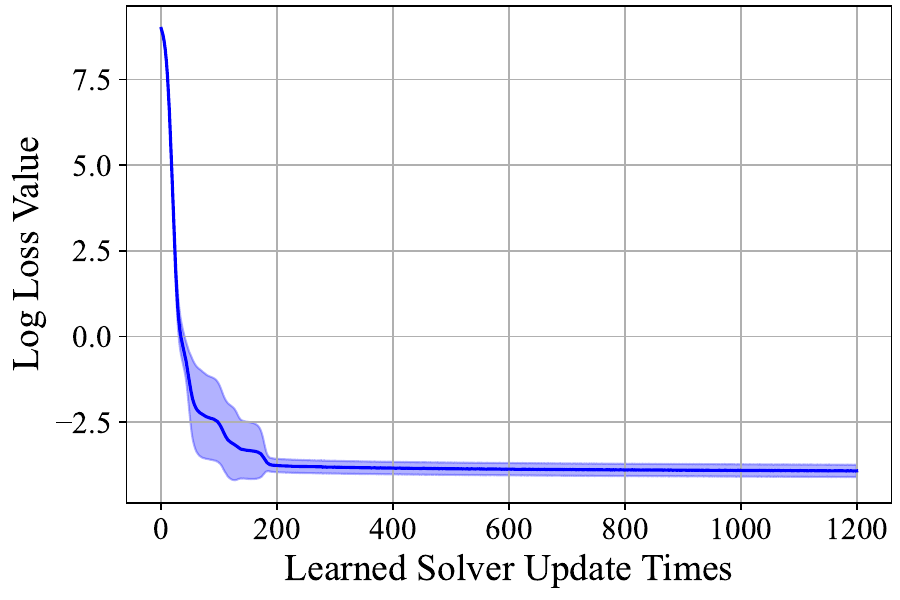}
    }
    \caption{Training loss curve of the 128KB UCL-sketch on the CAIDA dataset.}
    \label{fig:losses}
    \vspace{-0.3cm}
\end{figure}

\textbf{Convergence Rate.} Next, we evaluate the convergence rate of UCL-sketch, a crucial factor since, after configuring or resetting the measurement framework, it is vital to quickly obtain an effective solver ready for query processing. As shown in Fig.~\ref{fig:losses}, the training loss on the CAIDA dataset drops to an acceptable level within just 30 iterations and fully converges after approximately 190 iterations. Each training step—including both forward and backward passes—takes only milliseconds or even microseconds, enabling rapid convergence from scratch. Thus, (re-)training UCL-sketch minimally affects the system’s routine operation.

% \vspace{-0.3cm}

\subsection{Handling Changing Data Streams}

We evaluate the robustness and generalization of UCL-sketch under temporal and spatial variations in data streams. Massively unforeseen items are not specifically addressed, as such scenarios are of lesser practical concern. Throughout these experiments, we randomly sample 15\% of the key set to simulate variations in data stream behavior. Additionally, the experimental setup uses the CAIDA dataset with 64 KB.

\vspace{-0.2cm}

\begin{table}[htbp]
  \centering
  \caption{(ARE) Performance degradation with changing conditions}
  \vspace{-0.1cm}
    \resizebox{0.48\textwidth}{!}{\begin{tabular}{c|c|c|c|c|c|c|c|c}
    \toprule
    \multicolumn{1}{c|}{\textbf{\# of Interval}} & \multicolumn{4}{c|}{\textbf{Temporal Fluctuation Factor}} & \multicolumn{4}{c}{\textbf{Total Freq. Prop. of Hot Keys}} \\
    \cmidrule{2-9} \multicolumn{1}{c|}{\textbf{after Change}} & \textbf{2$\times$} & \textbf{10$\times$} & \textbf{20$\times$} & \textbf{50$\times$} & \textbf{90\%} & \textbf{75\%} & \textbf{50\%} & \textbf{25\%} \\
    \midrule
    SW 0 & 0.04\%    & 0.52\%    & 1.58\%    & 2.55\%    & 0.06\%    & 0.40\%    & 1.26\%    & 2.14\% \\
    \midrule
    SW 1 & 0.02\%    & 0.34\%    & 1.08\%    & 1.52\%    & 0.03\%    & 0.16\%    & 0.71\%    & 1.36\% \\
    \midrule
    SW 2 & 0.01\%    & 0.16\%    & 0.57\%    & 0.75\%    & 0.01\%    & 0.06\%    & 0.27\%    & 0.52\% \\
    \midrule
    SW 4 & 0.00\%    & 0.04\%    & 0.08\%    & 0.21\%    & 0.00\%    & 0.02\%    & 0.04\%    & 0.16\% \\
    \bottomrule
    \end{tabular}}
  \label{tab:burst}
  \vspace{-0.15cm}
\end{table}

\textbf{Temporal Fluctuation.}  We introduce various temporal fluctuations to the streaming data. For each selected key, we calculate its occurrence in the sliding window, and then multiply it by 2, 10, 20, and 50 respectively. As shown in Table~\ref{tab:burst}, UCL-sketch exhibits no significant degradation in ARE under these fluctuating conditions. Even with a 50$\times$ scaling factor, the increase in ARE remains below 3\%. Furthermore, the learned solver rapidly adapts to these fluctuations.

\textbf{Spatial Distribution.} 
We redistribute these packets across IP addresses to simulate variations in their spatial distribution. Specifically, we reassign the top 10\% of hot keys to the selected ones, so that they subsequently represent 90\%, 75\%, 50\%, and 25\% of the their (original) volume, respectively. Table~\ref{tab:burst} shows that UCL-sketch still consistently satisfies the near-optimal ARE across all spatial distributions. 

% \vspace{-0.3cm}

\subsection{Sensitivity Analysis}

We measure the influence of some key parameter settings, i.e. size of the bucket and hidden dimension, on accuracy, distribution, and resource usage. We use the Kosarak dataset in these experiments.

\begin{figure}
    \subfigure[w.r.t bucket length\label{bucket_usage}]{
        \centering
        \includegraphics[width=0.475\linewidth]{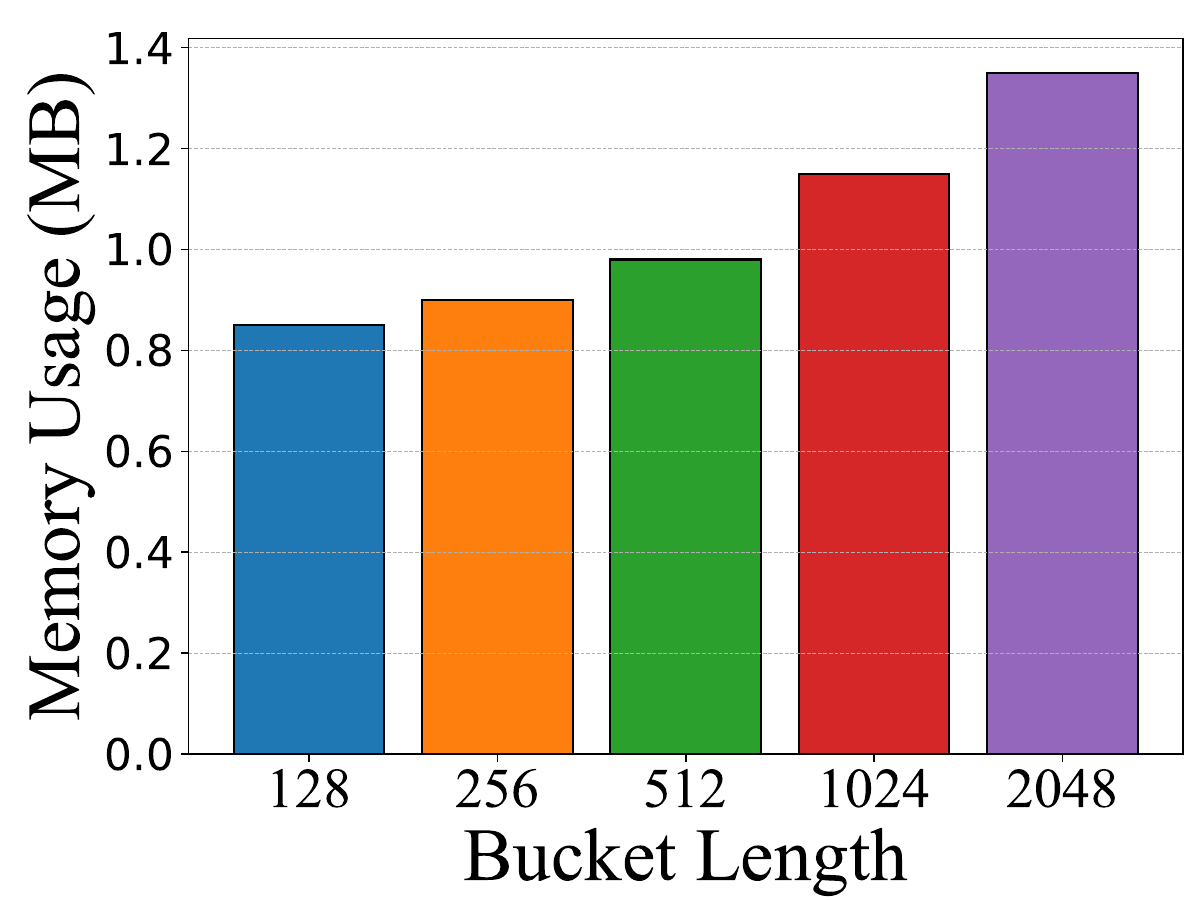}
    }
    \hfill
    \subfigure[w.r.t hidden dimension\label{hd_usage}]{
        \centering
        \includegraphics[width=0.455\linewidth]{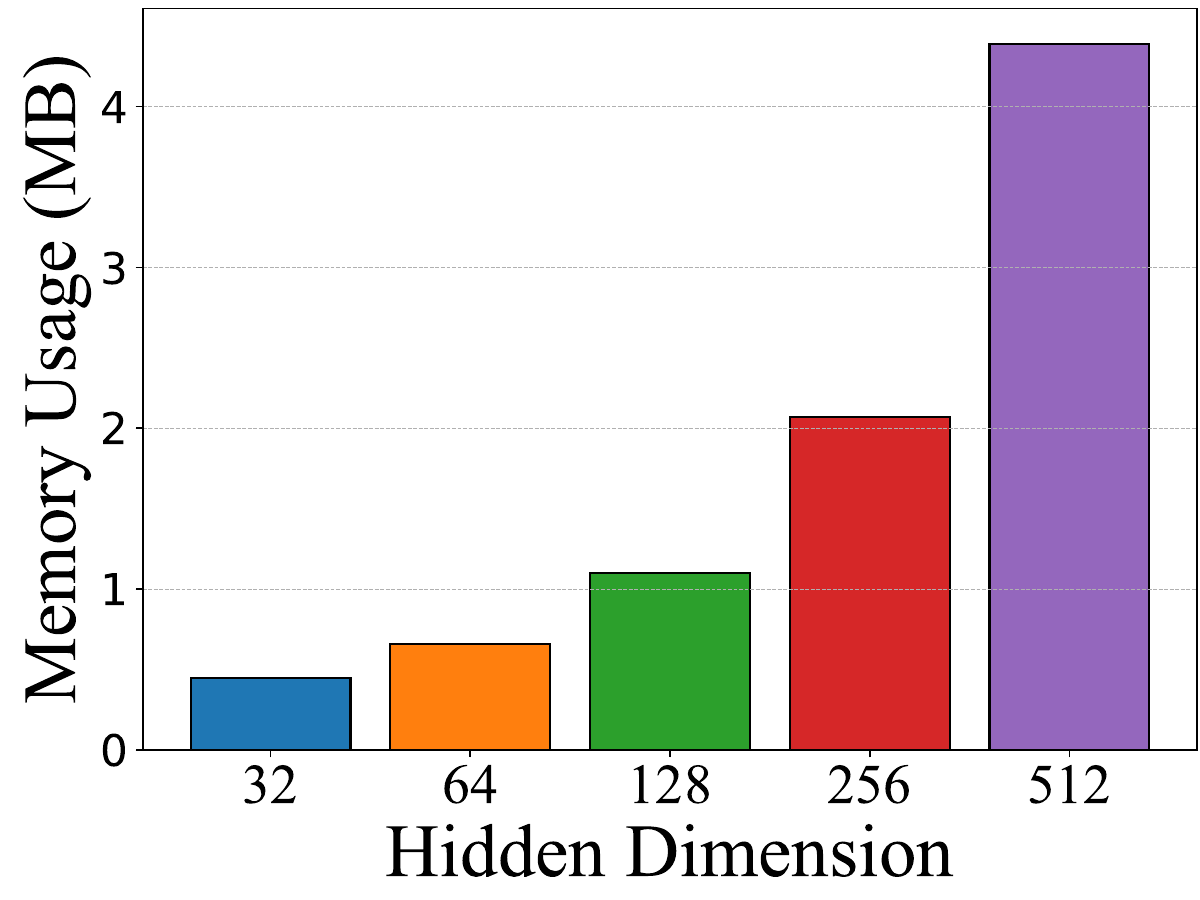}
    }
    \caption{Total trainable parameters of the learned solver in UCL-sketch.}
\end{figure}

\begin{figure}[htbp]
\centering
\subfigure[Average Relative Error]{
\includegraphics[width=0.46\linewidth]{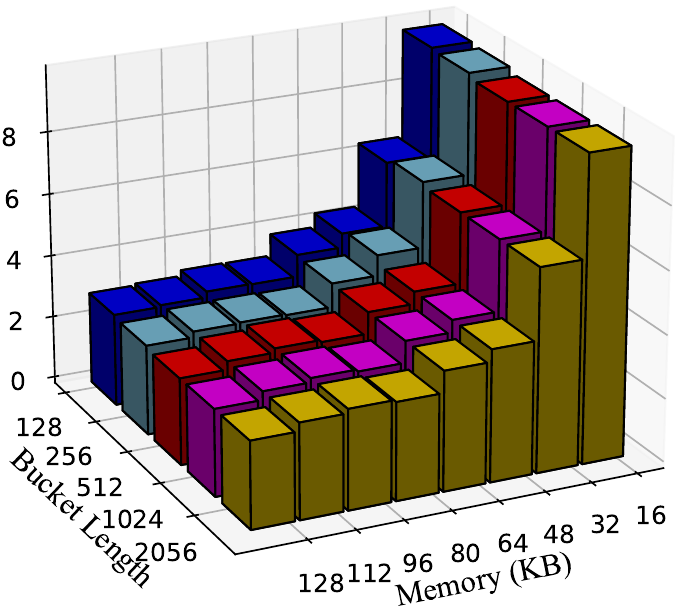}
}
\subfigure[Entorpy Absolute Error]{
\includegraphics[width=0.46\linewidth]{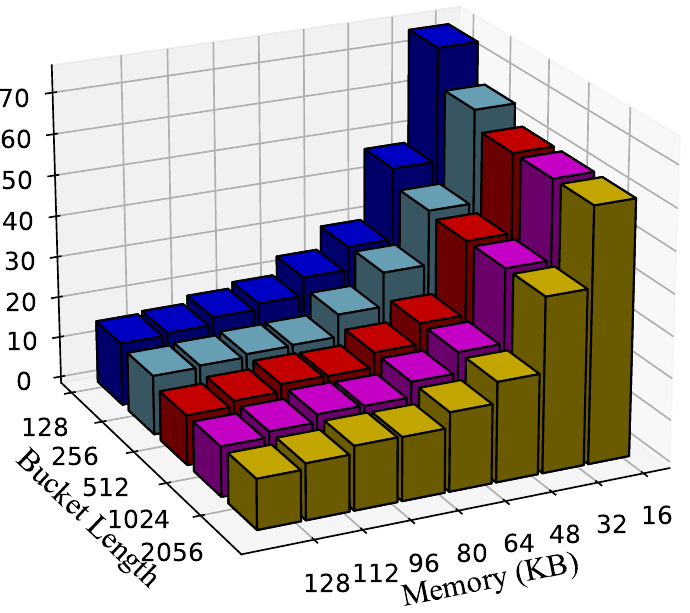}
}
\caption{Effects of the shared bucket length.}
\label{bucket_ae}
% \vspace{-0.45cm}
\end{figure}

\textbf{The impact of bucket length.} In this experiment, we vary bucket length in [128, 256, 512, 1024, 2056]. The results in Fig.~\ref{bucket_ae} show that a small size of bucket can achieve similar AREs, but it leads to notable degradation in the entropy of estimated frequencies. Therefore, we choose 512 or 1024 as its value in other experiments to make a balance between overall performance and memory usage (see Fig.~\ref{bucket_usage}). In fact, users can set bucket length according to the key space they are interested in, and a length of 1024 will be enough in general.

\begin{figure}[htbp]
\centering
\subfigure[Average Relative Error]{
\includegraphics[width=0.46\linewidth]{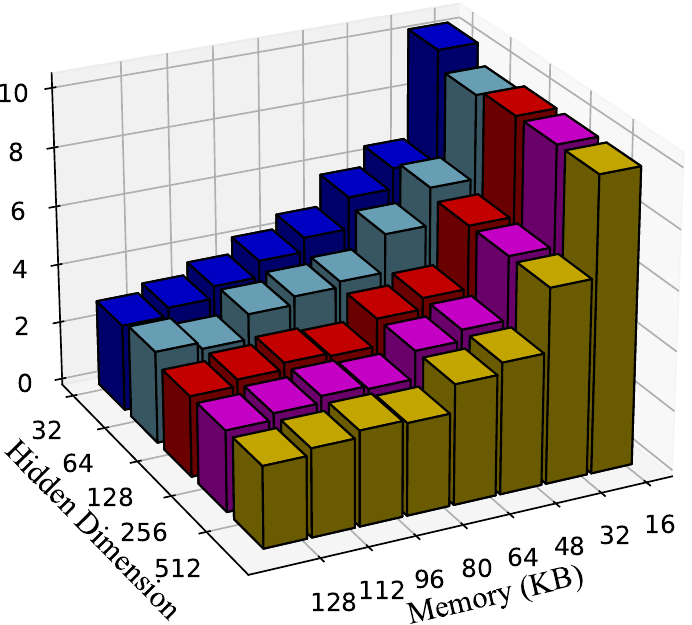}
}
\subfigure[Entorpy Absolute Error]{
\includegraphics[width=0.46\linewidth]{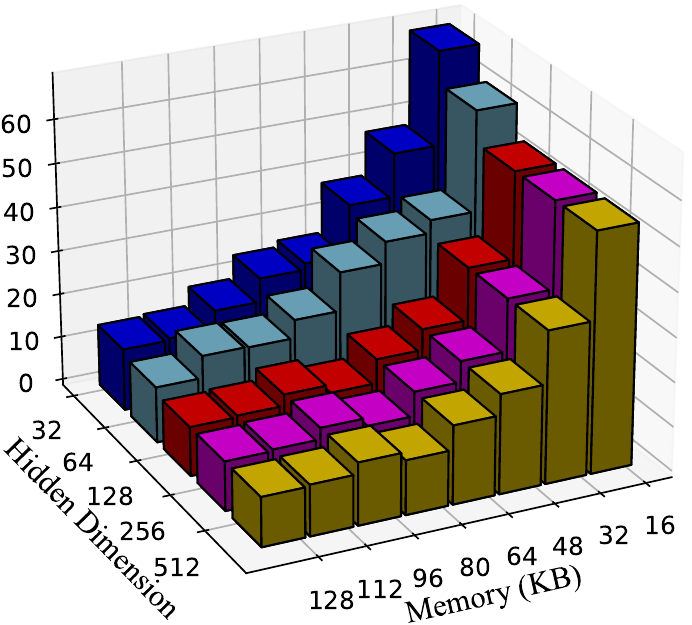}
}
\caption{Effects of the hidden dimension in the learned solver.}
\label{hd_ae}
% \vspace{-0.45cm}
\end{figure}

\textbf{The impact of hidden dimension.} We retrain models with hidden dimensions from 32 to 512 to obtain 5$\times$2 groups of experimental results in Fig.~\ref{hd_ae}. As shown in Fig.~\ref{hd_ae}, the performance of frequency estimation improves with increasing the hidden dimension, indicating that the higher-dimensional representation is available, the better the model can be trained and the better results can be achieved by the extracted feature. However, after the size reaches a certain value e.g. 128, the increase of dimension does not boost the recovery performance obviously. Also note that the memory overhead of the model exhibits exponential growth in high-dimensional (over 64) hidden spaces, which can be observed in Fig.~\ref{hd_usage}. We thus set it to 128 in all experiments.

% \vspace{-0.2cm}

% \vspace{-0.15cm}

\subsection{Abaltional Study}
\begin{figure}
    \centerline{\includegraphics[width=1.\linewidth]{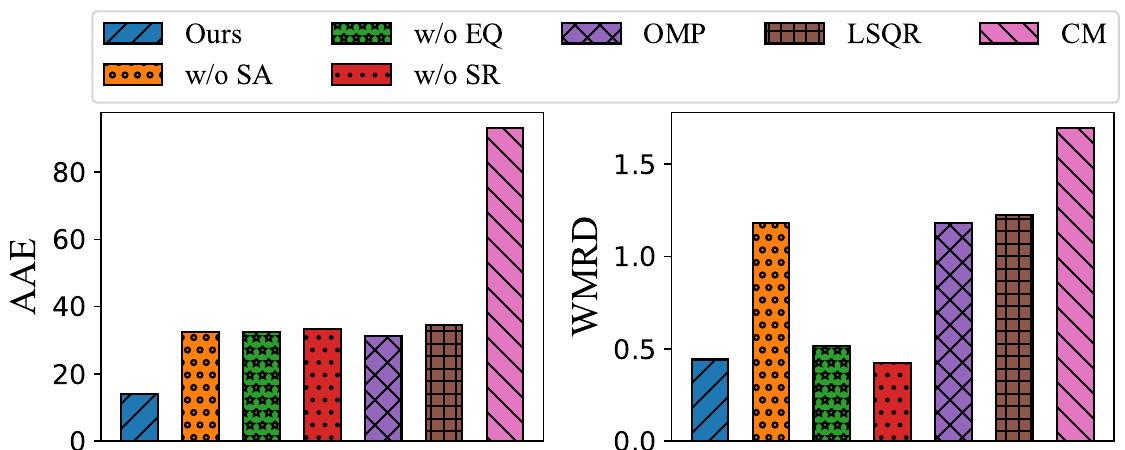}}
    \caption{Quantitative results of ablational studies.}
    \label{ablation_1}
    % \vspace{-0.5cm}
\end{figure}

In Fig.~\ref{ablation_1} and Table~\ref{entropy_ab}, we compare the original UCL-sketch with its modified version to analyze the impact of each component. Our used dataset is Kosarak and memory is 64KB.

% \vspace{-0.3cm}

\begin{table}[htbp]
  \centering
  \caption{Entropy absolute error with viariates of UCL-sketch}
  % \vspace{-0.2cm}
    \resizebox{0.45\textwidth}{!}{\begin{tabular}{c|c|c|c|c|c|c|c}
    \toprule
    Viariates & \cellcolor[rgb]{ .851,  .851,  .851}{UCL-sketch} & w/o SA & w/o EQ & w/o SR & OMP  & LSQR  & CM \\
    \midrule
    Absolute Error & \cellcolor[rgb]{ .851,  .851,  .851}\textbf{16.85} & 408.93 & 376.53 & 358.09 & 346.78 & 301.32 & 204.29 \\
    \bottomrule
    \end{tabular}}
  \label{entropy_ab}
  % \vspace{-0.25cm}
\end{table}

\begin{figure}
    \centerline{\includegraphics[width=1.\linewidth]{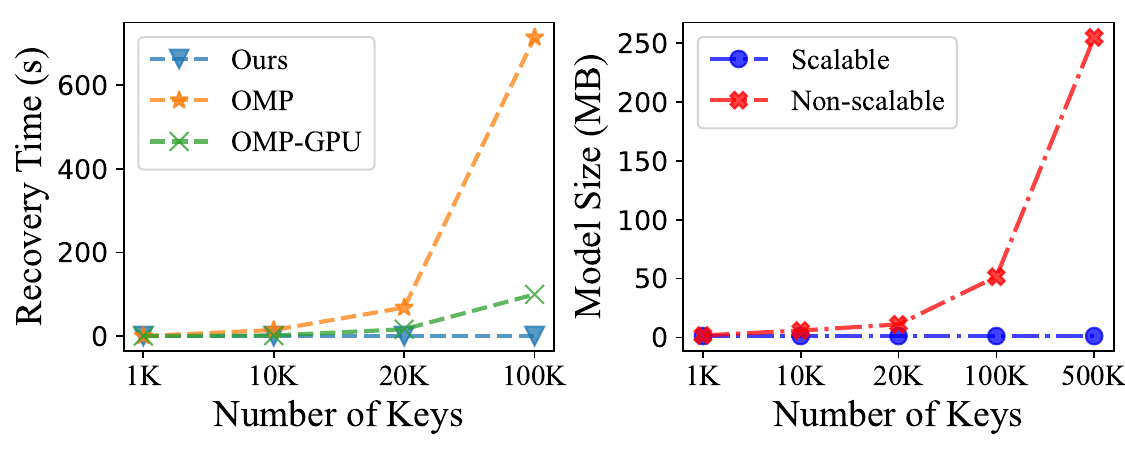}}
    \caption{\textbf{Left:} Recovery time. \textbf{Right:} Size of parameters.}
    \label{ablation_2}
    % \vspace{-0.5cm}
\end{figure}

\textbf{Learning version v.s. Non-learning version.} Three traditional versions of the decoding method are used for comparison: OMP used in SeqSketch, LSQR~\cite{paige1982lsqr}, and Count-Min (CM). We also implemented a Pytorch-version OMP (OMP-GPU) to further enhance the algorithm's parallelization. As shown in Fig.~\ref{ablation_1}, we can find that the accuracy under our learning-based algorithm is consistently better than that under non-learning methods. CM performs the worst since it does not utilize information from the linear system. Meanwhile, OMP is more precise than LSQR owing to its sparse greedy solution. We then measure its inference time for different numbers of keys. The left of Fig.~\ref{ablation_2} gives a recovery time of OMP or/and OMP-GPU that is over 60 seconds in some cases, which is much slower than ours which keeps the time around 0.5 seconds. At the same time, see the Appendix for additional experiments comparing UCL-sketch to similar equation-based sketching methods. Importantly, UCL-sketch also outperforms all other considered sketches on our real-world datasets. Therefore, proposed learning technologies do boost the design of high-performance sketches. 

\textbf{Basic training v.s. Optimized training.} Also see Fig.~\ref{ablation_1} for the ablation results on training options, whose details are as follows. \textit{Without EQ}: We retrain the solver by removing the third term in Eqn.~\ref{objective}. Noticeable performance degradation is observed in both AAE and WMRD when compared to ours. This has indicated the effectiveness of equivalent learning for handling the null-space ambiguity without ground truth. \textit{Without SR}: We remove the second loss term in Eqn.~\ref{objective} and retrain the model. The accuracy is close to the situation without EQ. But it's interesting to see that WMRD after discarding the sparse regularization is slightly lower than the original version. The reason may be the sparsity assumption limits the model's ability to learn the heavy-tailed distribution weakly. However, Table~\ref{entropy_ab} shows that UCL-sketch offers a much better estimation of frequency entropy than the other two variates for all memory sizes.

% \vspace{-0.3cm}

\textbf{Scalable network v.s. Non-scalable network.} To demonstrate the impact of \textit{scalable architecture (SA)}, we train a network with unshared buckets. Surprisingly, the unshared version does not achieve the best performance in Fig.~\ref{ablation_1}. This is because the parameter sharing acts as a form of regularization, preventing overfitting so that the solver is more likely to generalize well to future counters, especially for very large-scale streams. A summary of the required memory is reported on the right of Fig.~\ref{ablation_2}. Obviously, in the scalable network, with the increase in the number of keys, the parameters significantly decrease. By setting 500K unique items, the non-scalable network takes up a memory of over 250MB, while ours only requires a consumption of 1MB. This indicates that our compression scheme in parameters is very promising and practical, with an even better effect on estimation performance.

% \vspace{-0.2cm}

\section{Conclusion}
\label{sec:7}

In this paper, we present the first GT-free learning-based frequency estimation algorithm, called UCL-sketch which provides a novel perspective for approximate measurement by leveraging the binding between sketch sensing and machine learning. UCL-sketch extends the existing equation-based streaming algorithms with a ML technique complementing solver. It designs a training strategy to allow users to on-line train the deep solver without any ground truth. Using a scalable model architecture, UCL-sketch automatically and efficiently configures parameters for adapting to continuously expanding data streams. Through extensive evaluation, the efficiency and accuracy demonstrate the power of our methodology.

We believe that our investigation of learning-based sketches from the CS perspective has but scratched the surface and outline below current limitations of our approach, as well as intriguing directions for future research. \circled{1} Our method’s intuition and proof partly rely on Zipf’s law. While experiments show strong generalization, more realistic (learnable) priors may improve accuracy and training speed. \circled{2} We analyzed the Bloom Filter’s false positive rate for new keys, and it implies that ensuring a very low number of missed keys requires an almost linear increase in space. Designing a more compact data structure is left for future work. Finally, we hope that this work will spark more research in the area of learning combinatorial sketching techniques.

% \vspace{-0.3cm}

\bibliographystyle{IEEEtran}
\bibliography{main}

\newpage

\appendix

\subsection{Proofs} \label{proof}

In this section, we prove the theoretical results in the paper by partially following~\cite{candes2008restricted} and~\cite{dalt2022bayesian}.

\begin{figure}[htbp]
\centering
\subfigure[Count-Min Sketch Sensing]{
\includegraphics[width=0.85\linewidth]{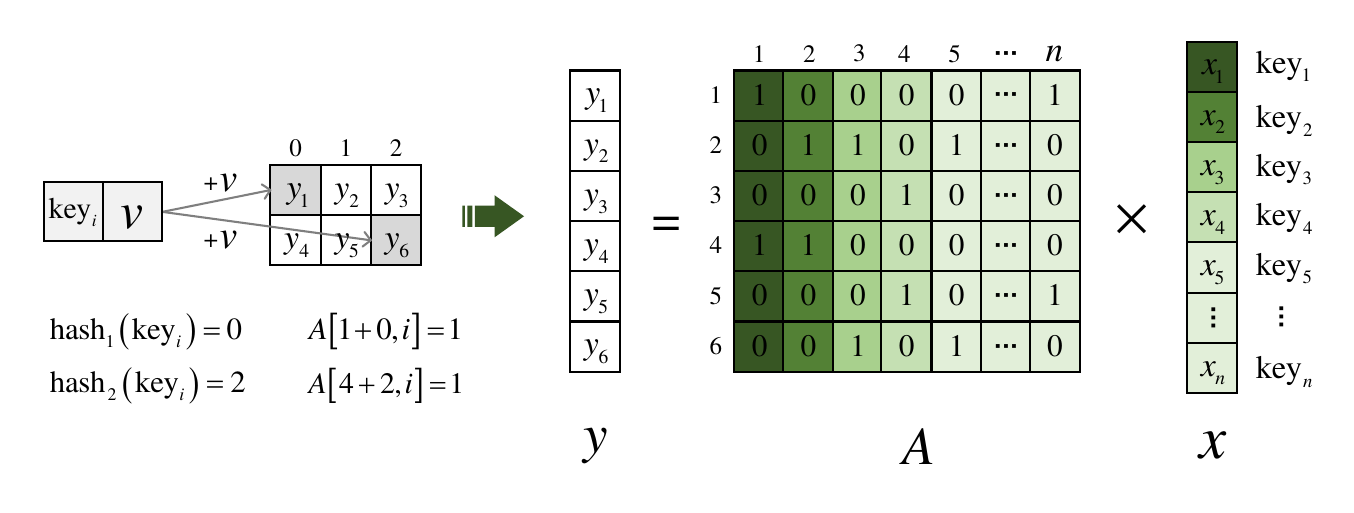}
}
\subfigure[Count Sketch Sensing]{
\includegraphics[width=0.85\linewidth]{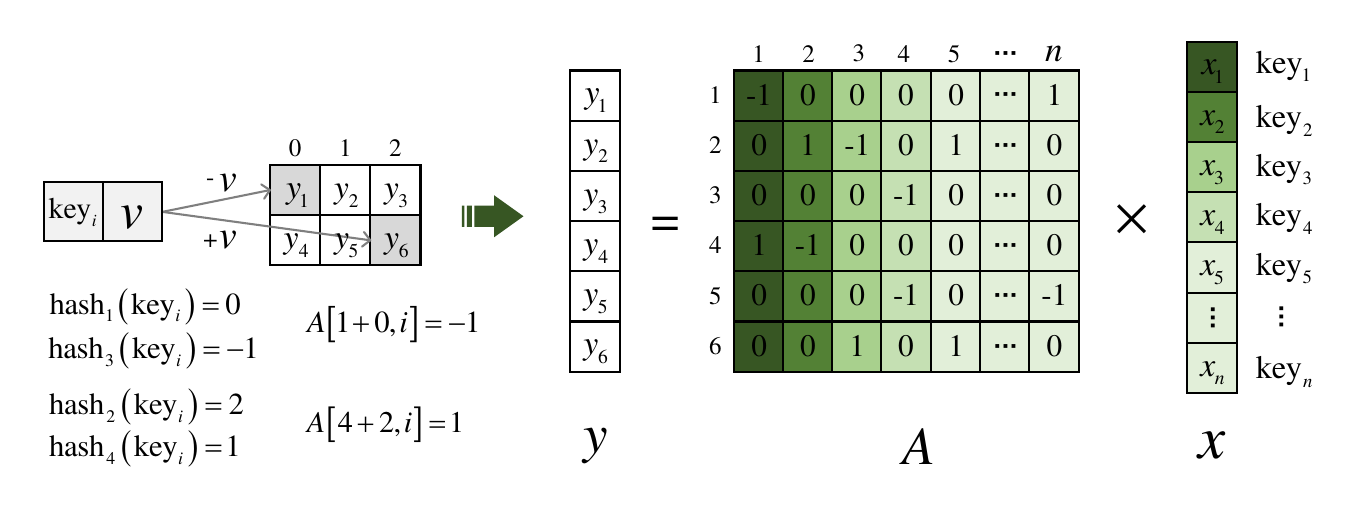}
}
\caption{The sensing process of and C-sketch and CM-sketch based on the linear hash operation.}
\label{sketch_sensing}
\end{figure}

\textbf{Theorem 1.} \textit{The {goal of} frequency estimation based on a {linear sketch} is equivalent to solving linear equations from the given keys, hash functions, and counters. Let $x\in C^N$ denote the vector of the streaming key-value sequence and $y\in C^M$ denote sketch counters, the insertion process corresponds to $y=\boldsymbol{A}x$, while the {result of} recovery phase corresponds to}
\begin{equation}
    x=\boldsymbol{A}^\dag y+(I-\boldsymbol{A}^\dag \boldsymbol{A})x,
    % \label{general_solution}
\end{equation} 
\textit{where $\boldsymbol{A}\in C^{M\times N}$ is an indicator matrix of mapping the vector $x$ to a buckets array $y$, and $\boldsymbol{A}^\dag \in C^{N\times M}$ satisfies $\boldsymbol{A}\boldsymbol{A}^\dag \boldsymbol{A}\equiv \boldsymbol{A}$.}

\textit{Proof.} Suppose that the hashing operation randomly maps incoming items to a bucket array uniformly at random. For an incoming key-value pair, the sketch selects one counter indexed by hashing the key with a hash function at each array. Let $\boldsymbol{A}\left[i,j\right]= 1$ or $-1$ if the $j$-th key is mapped to the $i$-th bucket, and set other entries in this row vector to $0$s. Then the insertion process for all key-value pairs can be equivalently represented as an algebraic equation $y=\boldsymbol{A}x$. For example, we present the mapping matrices for CM-sketch and C-sketch in Fig.~\ref{sketch_sensing}. Thus the approximated counters of a sketch can be calculated as
a decoding phase: we can obtain the general solution of per-key $x$, i.e., $x=\boldsymbol{A}^\dag y+(I-\boldsymbol{A}^\dag \boldsymbol{A})t, \forall t \in C^N$, where the first part is in the range-space of $\boldsymbol{A}$ while the latter is in the null-space. One can justify this by left multiplying each side of the equation by $\boldsymbol{A}$. Then 
\begin{equation}
    x = \boldsymbol{A}^\dag \boldsymbol{A}x+(I-\boldsymbol{A}^\dag \boldsymbol{A})t \Leftrightarrow x - t = \boldsymbol{A}^\dag \boldsymbol{A}\left(x - t\right)
\end{equation}
Since $\boldsymbol{A}^\dag \boldsymbol{A} \ne I$, we thus have $t = x$, which concludes the proof. $\hfill \Box$

\textbf{Lemma 1}. \textit{The space complexity of UCL-sketch is ${\mathcal O}\left( m_b + \frac{e}{{{\varepsilon _c}}}\ln \frac{1}{{{\delta _c}}} + bs \right)$, and the time complexity of update operation is ${\mathcal O}\left( {\log \frac{1}{{{\varepsilon _b}{\delta _b}}} + \ln \frac{1}{{{\delta _c}}}} \right)$ in the data plane.}

\textit{Proof.} Since the bucket arrays of CM-sketch contain $w \times d$ counters, the total size of the heavy filter is fixed as $b \times s$, and the Bloom Filter occupies $m_b$ bits, the total space complexity in the data plane is ${\mathcal O}\left( m_b + wd + bs \right)={\mathcal O}\left( m_b + \frac{e}{{{\varepsilon _c}}}\ln \frac{1}{{{\delta _c}}} + bs \right)$. For each item in the stream, it first requires 1 hash operation to locate its slot in the heavy filter. Thus, the time complexity of filtering is only ${\mathcal O}(1)$. Then the sketch hashes the key $d$ times, and the Bloom Filter hashes that $k_b$ times. Therefore, the time complexity of insertion is ${\mathcal O}\left( k_b + d + 1\right)={\mathcal O}\left( {\log \frac{1}{{{\varepsilon _b}{\delta _b}}} + \ln \frac{1}{{{\delta _c}}}} \right)$. $\hfill \Box$

\textbf{Lemma 2}. \textit{The keys coverage of the Bloom Filter in UCL-sketch obeys}
\begin{equation}
    \Pr \left(Y \ge K_y \right) \le \frac{K}{K_y}{\left( {1 - {e^{- \frac{{k_b}{K}}{m_b}}}} \right)},
    % \label{coverage}
\end{equation}
\textit{where variable $Y$ denotes the number of keys that are not covered but viewed as covered.}

\textit{Proof.} Suppose that independent hash functions uniformly map keys to random bits, the probability that a certain bit will still be $0$ after one insertion is $1-\frac{1}{m_b}$. Consequently, the probability that any bit of the Bloom Filter is $1$ after $K_i$ distinct items have been seen is given by ${1 - \left( {1 - \frac{1}{m_b}} \right)^{k_bK_i}}$. We can use the identity ${\left( {1 - \frac{1}{{{m_b}}}} \right)^{{m_b}}} = \frac{1}{e}$ for large $m_b \to \infty$, then we have approximation ${\left( {1 - \frac{1}{{{m_b}}}} \right)^{{k_b}K_i}} \approx {\left( {\frac{1}{e}} \right)^{{{{k_b}K_i} \mathord{\left/
 {\vphantom {{{k_b}K_i} {{m_b}}}} \right.
 \kern-\nulldelimiterspace} {{m_b}}}}}$. Therefore, the false positive probability of an unobserved key is $1 - {{e}^{-{{{k_b}K_i} \mathord{\left/
 {\vphantom {{{k_b}K_i} {{m_b}}}} \right.
 \kern-\nulldelimiterspace} {{m_b}}}}}$. Also note that $E\left( Y \right) = \sum\limits_{i = 1}^K {\left( {1 - {e^{{{ - {k_b}{i}} \mathord{\left/
 {\vphantom {{ - {k_b}{i}} {{m_b}}}} \right.
 \kern-\nulldelimiterspace} {{m_b}}}}}} \right)} \le K{\left( {1 - {e^{{{ - {k_b}{K}} \mathord{\left/
 {\vphantom {{ - {k_b}{K}} {{m_b}}}} \right.
 \kern-\nulldelimiterspace} {{m_b}}}}}} \right)}$. Now by Markov's inequality, the bound can be derived as: $\Pr \left(Y \ge K_y \right) \le \frac{E\left( Y \right)}{K_y} \le \frac{K}{K_y}{\left( {1 - {e^{{{ - {k_b}{K}} \mathord{\left/
 {\vphantom {{ - {k_b}{K}} {{m_b}}}} \right.
 \kern-\nulldelimiterspace} {{m_b}}}}}} \right)}$, which concludes the proof. $\hfill \Box$

 \textbf{Lemma 3}. \label{lemma3} \textit{Given disjoint subsets $T_a,T_b \subseteq \left\{ 1,2,3,\dots \right\}$ with $|T_a|, |T_b| = s$, and $x$ an arbitrary vector can be supported on them, we have} $\left\langle {A{x_{{T_a}}},A{x_{{T_b}}}} \right\rangle  \le {\sigma _{2s}}{\left\| {{x_{{T_a}}}} \right\|_2}{\left\| {{x_{{T_b}}}} \right\|_2}$.

 \textit{Proof.} The core idea of the proof is Parallelogram Identity. First, without loss of generality, we may assume that $x$ and $x'$ are \textit{unit} vectors with disjoint support sets as described above. Then, by Definition~\ref{RIC}, we have
 \[
     2(1 - \sigma_{s+s'}) \leq {\left\|\boldsymbol{A} x \pm \boldsymbol{A} x'\right\|}_{2}^2 \leq 2(1 + \sigma_{s+s'})
 \]
 The parallelogram identity states that:
 \[
     \left|\langle \boldsymbol{A} x, \boldsymbol{A} x' \rangle\right| = \frac{1}{4} {\|\boldsymbol{A} x + \boldsymbol{A} x'\|}_{2}^2 - {\left\|\boldsymbol{A} x - \boldsymbol{A} x'\right\|}_{2}^2 \leq \sigma_{s+s'}
 \]

 The vectors $x$ and $x'$ can be replaced by $x_{T_a}$ and $x_{T_b}$, respectively, yielding
 \[
     \left\langle \boldsymbol{A}{x_{T_a}}, \boldsymbol{A}{x_{T_b}} \right\rangle \le \left| \left\langle \boldsymbol{A}{x_{T_a}}, \boldsymbol{A}{x_{T_b}} \right\rangle \right| \le \sigma_{2s} {\left\|x_{T_a}\right\|}_2 {\left\|x_{T_b}\right\|}_2    
 \]
 This completes the proof of Lemma~\hyperref[lemma3]{3}. $\hfill \Box$

% \textit{Proof.} The proof of Lemma~\hyperref[lemma3]{3} can be concluded from~\cite{candes2008restricted}. According to~\cite{candes2008restricted}, $\left\langle {A{x},A{x'}} \right\rangle  \le {\sigma _{s+s'}}{\left\| {{x}} \right\|_2}{\left\| {{x'}} \right\|_2}$ holds for all $x, x'$ supported on the disjoint subsets. We replace $x, x'$ with $x_{{T_a}}, x_{{T_b}}$. Then, we have $\left\langle {A{x_{{T_a}}},A{x_{{T_b}}}} \right\rangle  \le {\sigma _{2s}}{\left\| {{x_{{T_a}}}} \right\|_2}{\left\| {{x_{{T_b}}}} \right\|_2}$. $\hfill \Box$

\textbf{Theorem 2}. \textit{Let $f=\left(f(1), f(2), \dots, f(n)\right)$ be the real volume vector of a stream that is stored in the sketch, where $f(i)$ denotes the volume of $i$-th distinct item. Consider $T_0$ as the locations of the $s$ largest volume of $f$. Assume that the reported volume vector $f^*$ is the optimal solution that minimizes the first two objective in Eqn.~3. Then the worst-case frequency estimation error is bounded by}
\begin{equation}
    {\left\| {{f^*} - f} \right\|_1} \le \frac{{ {{2 + \left( {2\sqrt 2  + 2} \right){\sigma _{2s}}}}}}{{1 - \left( {\sqrt 2  + 1} \right){\sigma _{2s}}}}{\left\| {{f_{T_0^c}}} \right\|_1}
    % \label{CS_Bound}
\end{equation}

\textit{Proof.} Given $f=f^*+l$, we start by dividing the residual vector $l$: let $T_1$ denote the locations of the largest $s$ value in $l_{T_0^c}$, $T_2$ denote the locations of the next largest $s$ value in $l_{T_0^c}$, and so on. By this definition, we can obtain ${\left\| {{l_{{T_j}}}} \right\|_2} = \sqrt {\sum\limits_{i} {l_{{T_j}}^2\left( i \right)} }  \le \sqrt {s{{\mathop {\max }\limits_{i} }^2}\left( {\left| {{l_{{T_j}}}\left( i \right)} \right|} \right)}  \le \frac{1}{{\sqrt s }}\sum\limits_{i} {\left| {{l_{{T_{j - 1}}}}\left( i \right)} \right|}  = \frac{1}{{\sqrt s }}\left\| {{l_{{T_{j - 1}}}}} \right\|_1$. This holds for $j \ge 2$, which derives the following useful inequality:
\begin{equation}
    \sum\limits_{j \ge 2} {{{\left\| {{l_{{T_j}}}} \right\|}_2}} \le \frac{1}{{\sqrt s }} \sum\limits_{j \ge 1} {{{\left\| {{l_{{T_j}}}} \right\|}_1}} = \frac{1}{{\sqrt s }} {{{\left\| {{l_{T_0^c}}} \right\|}_1}}
    \label{eq1}
\end{equation}

Since $f^*$ minimize $\left\| {{f^*}} \right\|_1$ subject to $Af^*=Af$, which intuitively means $\left\| {{f^*}} \right\|_1$ would not bigger than $\left\| f \right\|_1$, we have $\left\| f \right\|_1 \ge \left\| {{f^*}} \right\|_1$ and $\left\| Af^*-Af \right\|_2=\left\| Al \right\|_2 \approx 0$. It gives
\begin{equation}
    \begin{aligned}
    &{\left\| {{f_{{T_0}}}} \right\|_1} + {\left\| {{f_{T_0^c}}} \right\|_1}  = {\left\| f \right\|_1}\\
    &\ge {\left\| {{f^*}} \right\|_1} = {\left\| {f - l} \right\|_1} = {\left\| {{{\left( {f - l} \right)}_{{T_0}}}} \right\|_1} + {\left\| {{{\left( {f - l} \right)}_{T_0^c}}} \right\|_1} \\
    &\ge {\left\| {{f_{{T_0}}}} \right\|_1} - {\left\| {{f_{T_0^c}}} \right\|_1} + {\left\| {{l_{T_0^c}}} \right\|_1} - {\left\| {{l_{{T_0}}}} \right\|_1}\\
    &\Leftrightarrow {\left\| {{l_{T_0^c}}} \right\|_1} \le 2{\left\| {{f_{T_0^c}}} \right\|_1} + {\left\| {{l_{{T_0}}}} \right\|_1}.
    \end{aligned}
    \label{eq2}
\end{equation}

Also note that ${\left\| {{l_{{T_0}}}} \right\|_1} \le \sqrt s {\left\| {{l_{{T_0}}}} \right\|_2} \le \sqrt s {\left\| {{l_{{T_{0 \cup 1}}}}} \right\|_2}$, which is is derived by 
\begin{equation}
    \begin{aligned}
        \frac{1}{{\sqrt s }}{\left\| {{l_{{T_0}}}} \right\|_1} & = \sqrt {{{\left( {\sum\limits_{i} {\frac{1}{{\sqrt s }}|{l_{{T_0}}}\left( i \right)|} } \right)}^2}} \\
        &\le \sqrt {{\left({\sum\limits_{j = 1}^s {{\frac{1}{{s }}}} }\right)}\sum\limits_{i} {l_{{T_0}}^2\left( i \right)} }  = {\left\| {{l_{{T_0}}}} \right\|_2}
    \end{aligned}
\end{equation}
using Cauchy–Schwarz inequality. 
And following from Definition~\hyperref[def]{1}, one can get $\left( {1 - {\sigma _{2s}}} \right)\left\| {{l_{{T_{0 \cup 1}}}}} \right\|_2^2 \le \left\| {A{l_{{T_{0 \cup 1}}}}} \right\|_2^2$. Therefore, we instead bound $\left\| {A{l_{{T_{0 \cup 1}}}}} \right\|_2^2$ as follows:
\begin{equation}
    \begin{aligned}
        & \left\| {A{l_{{T_{0 \cup 1}}}}} \right\|_2^2 = \left\langle {A{l_{{T_{0 \cup 1}}}},A{l_{{T_{0 \cup 1}}}}} \right\rangle  = \left\langle {A{l_{{T_{0 \cup 1}}}},A\left( {l - {l_{T_{0 \cup 1}^c}}} \right)} \right\rangle \\
        & \le \left| {\left\langle {A{l_{{T_{0 \cup 1}}}},Al} \right\rangle } \right| + \left| {\left\langle {A{l_{{T_{0 \cup 1}}}},A{l_{T_{0 \cup 1}^c}}} \right\rangle } \right| \\
        & \le {\left\| {A{l_{{T_{0 \cup 1}}}}} \right\|_2}{\left\| {Al} \right\|_2} + \sum\limits_{j \ge 2} {\left| {\left\langle {A{l_{{T_{0 \cup 1}}}},A{l_{{T_j}}}} \right\rangle } \right|} \\
        & \approx 0 + \sum\limits_{j \ge 2} {\left| {\left\langle {A{l_{{T_0}}} + A{l_{{T_1}}},A{l_{{T_j}}}} \right\rangle } \right|} \\
        & \le \sum\limits_{j \ge 2} {\left| {\left\langle {A{l_{{T_0}}},A{l_{{T_j}}}} \right\rangle } \right|} + \sum\limits_{j \ge 2} {\left| {\left\langle {A{l_{{T_1}}},A{l_{{T_j}}}} \right\rangle } \right|}.
    \end{aligned}
\end{equation}
Here using the Lemma~\hyperref[lemma3]{3}, we have 
\begin{equation}
    \sum\limits_{j \ge 2} {\left| {\left\langle {A{l_{{T_0}}},A{l_{{T_j}}}} \right\rangle } \right|} \le {\sigma _{2s}} \left\| {{l_{{T_0}}}} \right\|_2 \sum\limits_{j \ge 2} {\left\| {{l_{{T_j}}}} \right\|_2}
\end{equation}
The proof for the upper bound of $\sum\limits_{j \ge 2} {\left| {\left\langle {A{l_{{T_1}}},A{l_{{T_j}}}} \right\rangle } \right|}$ follows from the similar procedure, and thus applying Ineqn.~\ref{eq1}, 
\begin{equation}
    \begin{aligned}
        & \left\| {A{l_{{T_{0 \cup 1}}}}} \right\|_2^2 \le {\sigma _{2s}} \left(\left\| {{l_{{T_0}}}} \right\|_2 + \left\| {{l_{{T_1}}}} \right\|_2 \right) \sum\limits_{j \ge 2} {\left\| {{l_{{T_j}}}} \right\|_2} \\
        & \le \sqrt 2 {\sigma _{2s}} \left\| {{l_{{T_{0 \cup 1}}}}} \right\|_2\sum\limits_{j \ge 2} {\left\| {{l_{{T_j}}}} \right\|_2} \le \sqrt {\frac{2}{{s }}} {\sigma _{2s}} \left\| {{l_{{T_{0 \cup 1}}}}} \right\|_2 {{{\left\| {{l_{T_0^c}}} \right\|}_1}}
    \end{aligned}
\end{equation}
where the first part of the second line is derived by 
\begin{equation}
    \begin{aligned}
        \left\| {{l_{{T_0}}}} \right\|_2 + \left\| {{l_{{T_1}}}} \right\|_2 & = \sqrt { {\left\| {{l_{{T_0}}}} \right\|_2^2 + \left\| {{l_{{T_1}}}} \right\|_2^2 + 2{{\left\| {{l_{{T_0}}}} \right\|}_2}{{\left\| {{l_{{T_1}}}} \right\|}_2}} } \\
        &\le \sqrt {2\left( {\left\| {{l_{{T_0}}}} \right\|_2^2 + \left\| {{l_{{T_1}}}} \right\|_2^2} \right)} = \sqrt 2 \left\| {{l_{{T_{0 \cup 1}}}}} \right\|_2
    \end{aligned}
\end{equation}

Then recall that 
\begin{equation}
    \begin{aligned}
        & {\left\| {{l_{{T_{0 \cup 1}}}}} \right\|_2^2} \le \frac{1}{{1 - {\sigma _{2s}}}} \left\| {A{l_{{T_{0 \cup 1}}}}} \right\|_2^2 \le \frac{\sqrt{{2 \mathord{\left/
 {\vphantom {2 s}} \right.
 \kern-\nulldelimiterspace} s}}{\sigma _{2s}}}{{1 - {\sigma _{2s}}}}  \left\| {{l_{{T_{0 \cup 1}}}}} \right\|_2 {{{\left\| {{l_{T_0^c}}} \right\|}_1}} \\
        & \Leftrightarrow {\left\| {{l_{{T_{0 \cup 1}}}}} \right\|_2} \le \frac{\sqrt{{2 \mathord{\left/
 {\vphantom {2 s}} \right.
 \kern-\nulldelimiterspace} s}}{\sigma _{2s}}}{{1 - {\sigma _{2s}}}} {{{\left\| {{l_{T_0^c}}} \right\|}_1}}
    \end{aligned}
\end{equation}
which gives ${\left\| {{l_{{T_0}}}} \right\|_1} \le \sqrt s {\left\| {{l_{{T_{0 \cup 1}}}}} \right\|_2} \le \frac{\sqrt 2{\sigma _{2s}}}{{1 - {\sigma _{2s}}}} {{{\left\| {{l_{T_0^c}}} \right\|}_1}}$. Now we combine it with Ineqn.~\ref{eq2} to obtain the certain bound
\begin{equation}
    \begin{aligned}
    &{\left\| {{l_{{T_0}}}} \right\|_1} \le \frac{\sqrt 2{\sigma _{2s}}}{{1 - {\sigma _{2s}}}} {{{\left\| {{l_{T_0^c}}} \right\|}_1}} \le \frac{\sqrt 2{\sigma _{2s}}}{{1 - {\sigma _{2s}}}} \left( 2{\left\| {{f_{T_0^c}}} \right\|_1} + {\left\| {{l_{{T_0}}}} \right\|_1} \right) \\
    &\Leftrightarrow {\left\| {{l_{{T_0}}}} \right\|_1} \le \frac{{2\sqrt 2 {\sigma _{2s}}}}{{1 - \left( {\sqrt 2  + 1} \right){\sigma _{2s}}}} {\left\| {{f_{T_0^c}}} \right\|_1}.
    \end{aligned}
\end{equation}
Finally, the error bound of reported volumes is given by
\begin{equation}
    \begin{aligned}
        & {\left\| {{f^*} - f} \right\|_1} = {\left\| {{l}} \right\|_1} = {\left\| {{l_{T_0^c}}} \right\|_1} + {\left\| {{l_{T_0}}} \right\|_1} \\
        & \le 2{\left\| {{f_{T_0^c}}} \right\|_1} + 2{\left\| {{l_{{T_0}}}} \right\|_1} \le \frac{{ {{2 + \left( {2\sqrt 2  + 2} \right){\sigma _{2s}}}}}}{{1 - \left( {\sqrt 2  + 1} \right){\sigma _{2s}}}}{\left\| {{f_{T_0^c}}} \right\|_1},
    \end{aligned}
\end{equation}
which concludes the proof. $\hfill \Box$

\textbf{Theorem 3}. \textit{A necessary condition for recovering the true volume from compressed counters is that the following linear system has a unique solution:}
\begin{equation}
    \boldsymbol{B}x = \left( {\begin{array}{*{20}{c}}
\boldsymbol{A}\\
{\boldsymbol{A}{T_{{p_1}}}}\\
 \vdots \\
{\boldsymbol{A}{T_{{p_{\left| \mathcal{P} \right|}}}}}
\end{array}} \right)x = \left( {\begin{array}{*{20}{c}}
y\\
{{y^{{1}}}}\\
 \vdots \\
{{y^{{{\left| \mathcal{P} \right|}}}}}
\end{array}} \right),
\end{equation}
where $y^{\left(\cdot\right)}$ is the measurement corresponding to the transformation $p_{\left(\cdot\right)}$. Rigorously, rank$\left(\boldsymbol{B}\right) = n$.

\textit{Proof.} Recall that the general form of the true frequency is $x=\boldsymbol{A}^\dag y+(I-\boldsymbol{A}^\dag \boldsymbol{A})x$ as Theorem~\hyperref[theorem1]{1} shows, thus for $p_1, p_2, \dots, p_{|\mathcal{P}|} \in \mathcal{P}$, 
\begin{equation}
    \begin{aligned}
        &x=\boldsymbol{A}^\dag y+(I-\boldsymbol{A}^\dag \boldsymbol{A})x\\
        &x = \left((\boldsymbol{A}T_{p_1})^\dag y^1+(I-(\boldsymbol{A}T_{p_1})^\dag (\boldsymbol{A}T_{p_1})\right) x\\
        &\ \vdots \ \ \ \ \ \ \ \ \ \ \ \ \vdots \ \ \ \ \ \ \ \ \ \ \ \ \ \ \ \ \ \ \ \ \ \ \ \ \ \ \ \ \ \ \ \ \ \ \vdots \\
        &x = \left((\boldsymbol{A}T_{p_{|\mathcal{P}|}})^\dag y^{|\mathcal{P}|}+(I-(\boldsymbol{A}T_{p_{|\mathcal{P}|}})^\dag (\boldsymbol{A}T_{p_{| \mathcal{P}|}})\right) x.
    \end{aligned}
\end{equation}
Stacking all these equations together into

\begin{equation}
    0 = \left( {\begin{array}{*{20}{c}}
\boldsymbol{A}^\dag y\\
(\boldsymbol{A}T_{p_1})^\dag y^1\\
 \vdots \\
(\boldsymbol{A}T_{p_{|\mathcal{P}|}})^\dag y^{|\mathcal{P}|}
\end{array}} \right) - \left( {\begin{array}{*{20}{c}}
\boldsymbol{A}^\dag \boldsymbol{A}\\
(\boldsymbol{A}T_{p_1})^\dag \boldsymbol{A}T_{p_1}\\
 \vdots \\
(\boldsymbol{A}T_{p_{|\mathcal{P}|}})^\dag \boldsymbol{A}T_{p_{|\mathcal{P}|}}
\end{array}} \right) x
\label{remove}
\end{equation}

By left multiplying each side of Eqn.~\ref{remove} by $\left( \boldsymbol{A}, \boldsymbol{A}{T_{{p_1}}},\cdots, \boldsymbol{A}{T_{{p_{\left| \mathcal{P} \right|}}}} \right)$, we have that

\begin{equation}
    \boldsymbol{B}x = \left( {\begin{array}{*{20}{c}}
\boldsymbol{A}\\
{\boldsymbol{A}{T_{{p_1}}}}\\
 \vdots \\
{\boldsymbol{A}{T_{{p_{\left| \mathcal{P} \right|}}}}}
\end{array}} \right)x = \left( {\begin{array}{*{20}{c}}
y\\
{{y^{{1}}}}\\
 \vdots \\
{{y^{{{\left| \mathcal{P} \right|}}}}}
\end{array}} \right),
\end{equation}

and therefore $\boldsymbol{B} \in \mathbb{R}^{|\mathcal{P}|m \times n}$ needs to be of full rank $n$ so that the true frequency $x$ can be accurately recovered from the null space, which concludes the proof. $\hfill \Box$

\textbf{Theorem 4}. \textit{Assume that the prior of each $k_i$-associated frequency in per-key GT vector $x$ satisfies $\Pr \left( x(i) = a \right) = \mathcal{N}\left( a; \mu_i, \sigma_i^2 \right)$, then for $\forall k_i$ and ${\left\| x \right\|}_1 \rightarrow \infty$, through asymptotic approximation, its posterior estimate based on sketch counters $y$ has the following general form:}
\begin{equation}
    \hat x(i) = \frac{{{\mu _i}\left\| x \right\|_2^2 + \sigma _i^2\left( {w\sum\nolimits_{j = 1}^d {{y_{j \times {H_j}\left( {{k_i}} \right)}} - d{{\left\| x \right\|}_1}} } \right)}}{{\left\| x \right\|_2^2 + \sigma _i^2d(w - 1)}}
\end{equation}
\textit{where $H_j$ is the indep. hash function: $\boldsymbol{\Omega} \rightarrow \{1,\dots,w\}$.}

\textit{Proof.} We first introduce indicator variables $\mathcal{I}_{j,k_i,k_f}$ which are 1 if $(k_i \ne k_f) \wedge (H_j(k_i)=H_j(k_f))$ and 0 otherwise. Then we have
\begin{equation}
    {y_{j \times {H_j}\left( {{k_i}} \right)}} = x(i) + \sum\limits_{{k_f} \in \Omega} \mathcal{I}_{j,k_i,k_f} x(f)
\end{equation}

Apparently, $\sum\limits_{{k_f} \in \Omega} \mathcal{I}_{j,k_i,k_f} x(f)$ is a Gaussian random variable due to linear properties of the Gaussian distribution, and assuming $x(f) \in \Omega$ are independent and identically distributed, we can get the following approximation
\begin{equation}
    {\left\{\begin{aligned}
        &E \left(\mathcal{I}_{j,k_i,k_f} x(f) \right)= \frac{x(f)}{w}, \\
        &Var \left(\mathcal{I}_{j,k_i,k_f} x(f) \right)= \frac{w-1}{w^2}x^2(f)
    \end{aligned}\right.}
\end{equation}
So
\begin{equation}
    \sum\limits_{{k_f} \in \Omega} \mathcal{I}_{j,k_i,k_f} x(f) \sim \mathcal{N} \left( \frac{\|x\|_1 - x(i)}{w}, \frac{w-1}{w^2}\left(\|x\|^2_2-x^2(i)\right) \right)
\end{equation}
Now conditioned on the event $x(i)=a$,
\begin{equation}
    y_{j \times {H_j}\left( {{k_i}} \right)} \sim \mathcal{N} \left(a + \frac{\|x\|_1 - a}{w}, \frac{w-1}{w^2}\left(\|x\|^2_2-a^2\right) \right), 
\end{equation}
which also means 
\begin{equation}
    \begin{aligned}
        &\Pr \left(y_{j \times {H_j}\left( {{k_i}} \right)}=v_{j,k_i} | x(i)=a \right) \\
        &=\mathcal{N} \left(y_{j \times {H_j}\left( {{k_i}} \right)}; a + \frac{\|x\|_1 - a}{w}, \frac{w-1}{w^2}\left(\|x\|^2_2-a^2\right) \right)
    \end{aligned}
    \label{dist}
\end{equation}

Recalling the problem definition section in the main text, the sketching goal $\mathop {\max }\limits_{{x}} \log p\left( {{x}|{y}} \right), \  {\rm {s.t.}} \ {y}={\boldsymbol{A}}{x}$ can also be reformulated in the form of MAP (Maximum A Posteriori):
\begin{equation}
    \begin{aligned}
        \mathop {\max } \limits_a & \sum\limits_{j = 1}^d {\log \left( {\Pr \left(y_{j \times {H_j}\left( {{k_i}} \right)}=v_{j,k_i} | x(i)=a \right)} \right)} \\
        &+ \log \left( \Pr \left( x_i=a \right) \right)
    \end{aligned}
\end{equation}
By combining the prior and Eqn.~\ref{dist}, we can write
\begin{equation}
    \begin{aligned}
        \hat x(i)=\mathop {\max }\limits_a & \sum\limits_{j = 1}^d \left( - \frac{1}{2}\log \left( {\left\| x \right\|_2^2 - {a^2}} \right) \right) - \frac{{{{\left( {a - {\mu _i}} \right)}^2}}}{{2\sigma _i^2}}\\
        &+ \sum\limits_{j = 1}^d \left( {  - \frac{{{{\left( {w(y_{j \times {H_j}} - a) - ({{\left\| x \right\|}_1} - a)} \right)}^2}}}{{2(w - 1)(\left\| x \right\|_2^2 - {a^2})}}} \right)
    \end{aligned}
\end{equation}
Notice that $a^2 \ll \|x\|^2_2$ especially when $\|x\|_1 \rightarrow \infty$, $\left\| x \right\|_2^2 - {a^2}$ is thus dominated by $\left\| x \right\|_2^2$. Then by setting the term to the constant and ignoring it, we have the following surrogate approximation result:
\begin{equation}
    \begin{aligned}
        \hat x(i) \approx & \mathop {\max }\limits_a \frac{{{w^2}}}{{2(w - 1)\left( {\left\| x \right\|_2^2 - {a^2}} \right)}}\sum\limits_{j = 1}^d {\left( {y - a - \frac{{{{\left\| x \right\|}_1} - a}}{w}} \right)} \\
        +&  \frac{{{w^2}}}{{2(w - 1)\left( {\left\| x \right\|_2^2 - {a^2}} \right)}} \cdot \frac{{(w - 1){{(a - {\mu _i})}^2}\left( {\left\| x \right\|_2^2 - {a^2}} \right)}}{{{w^2}\sigma _i^2}}
    \end{aligned}
\end{equation}

It is also reasonable to drop the relatively constant coefficient ${{w^2}}/{2(w - 1)\left( {\left\| x \right\|_2^2 - {a^2}} \right)}$, the formula can be thus further simplified to:
\begin{equation}
    \begin{aligned}
        \hat x(i) \approx \mathop {\max }\limits_a & \frac{{(w - 1){{(a - {\mu _i})}^2}\left( {\left\| x \right\|_2^2 - {a^2}} \right)}}{{{w^2}\sigma _i^2}} \\
        &+ \sum\limits_{j = 1}^d {\left( {y - a - \frac{{{{\left\| x \right\|}_1} - a}}{w}} \right)}
    \end{aligned}
\end{equation}
Finally, we take its derivative with respect to $a$ and set it to 0, then the maximum is attained when
\begin{equation}
    \hat x(i) \approx \frac{{{\mu _i}\left\| x \right\|_2^2 + \sigma _i^2\left( {w\sum\nolimits_{j = 1}^d {{y_{j \times {H_j}\left( {{k_i}} \right)}} - d{{\left\| x \right\|}_1}} } \right)}}{{\left\| x \right\|_2^2 + \sigma _i^2d(w - 1)}}
\end{equation}
$\hfill \Box$

\begin{figure}
    \centerline{\includegraphics[width=1.\linewidth]{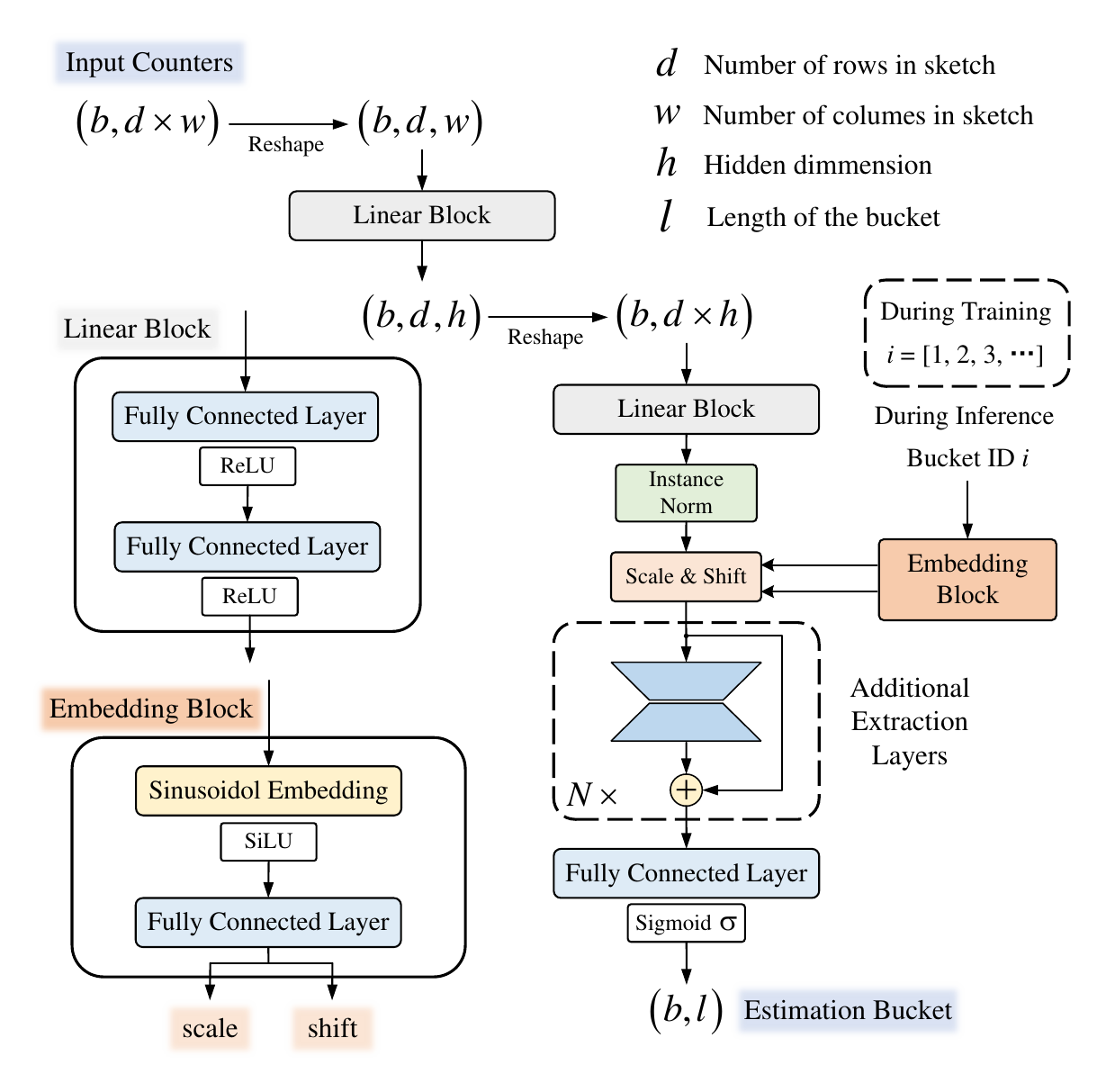}}
    \caption{Neural network architecture of the learned solver used in UCL-sketch.}
    \label{model}
\end{figure}

\subsection{Model Details} The details of our neural network (NN) architecture are illustrated in Fig.~\ref{model}. To recover an estimation bucket of length $l$, the size of the NN input measurements is $d \times w$. The measurement vector is first transformed to the shape $(d, w)$. After projection by a linear block, each containing two fully connected layers (FCs) using ReLU Nonlinear, the size of the tensor returns to $d \times h$. Then, it will be fed into the second linear block. We adopt this design with the following motivations: (a) Considering the need to reduce the computational cost, the total number of parameters is several times less than that of its counterpart without transformations. (b) Motivated by the studies in sketching algorithms, hash (row) independence should have a certain regularization effect on the estimation process. Next, we integrate an embedding block to learn the bucket offset with information in the sketch. For the bucket ID, we employ Sinusoidal embedding to represent each \textit{i} as a \textit{h}-dimensional vector, and then apply one fully connected layer after activating by SiLU Nonlinear. \textit{i} is injected into the network with ${\rm{scale}}_i x + {\rm{shift}}_i$, where $x$ is the shallow representation of our sampled counters. The fused feature is then extracted by a series of layers, e.g. FCs. Finally, features are projected back to the ground-truth domain with the Sigmoid function, as the normalized frequency is guaranteed to be $\le 1$.

\subsection{Implementation Details}\label{details}

Our experiments run in a machine with one AMD 6-Core CPU (3.70 GHz), 32GB DRAM, and a single 12GB NVIDIA GeForce RTX 3060 GPU. For the learning-driven part, we used the PyTorch implementation. Besides, all these experiments are repeated multiple times using different fixed random seeds, and then their average results are reported in this paper.

\textbf{Parameters of Baselines.}
The key details have already been presented at the beginning of the experimental section in the main text. Here, we list the parameters of each baseline method on the Retail dataset in detail in Table.~\ref{tab:baseline_params}, and the other datasets follow similarly. We omit SeqSketch here, as its data structure setup is essentially identical to ours.

\begin{table}[htbp]
  \centering
  \caption{Key parameters of baselines (\# of hash func. is fixed to 4)}
    \resizebox{0.5\textwidth}{!}{\begin{tabular}{c|c|c|c|c|c|c|c|c}
    \toprule
    \textbf{Memory Usage} & \textbf{16KB} & \textbf{32KB} & \textbf{48KB} & \textbf{64KB} & \textbf{80KB} & \textbf{96KB} & \textbf{112KB} & \textbf{128KB} \\
    \midrule
    \multicolumn{9}{c}{\textit{\textbf{Count-Min Sketch (CM)}}} \\
    \midrule
    \textbf{Width of sketch} & 1024  & 2048  & 3072  & 4096  & 5120  & 6144  & 7168  & 8192 \\
    \midrule
    \multicolumn{9}{c}{\textit{\textbf{Count Sketch (CS)}}} \\
    \midrule
    \textbf{Width of sketch} & 1024  & 2048  & 3072  & 4096  & 5120  & 6144  & 7168  & 8192 \\
    \midrule
    \multicolumn{9}{c}{\textit{\textbf{Learned Count-Min Sketch (LCM)}}} \\
    \midrule
    \textbf{\# of heavy buckets} & 512   & 1536  & 2048  & 3072  & 3584  & 4608  & 5632  & 6144 \\
    \midrule
    \textbf{Width of sketch} & 512   & 1024  & 2048  & 2048  & 3096  & 3096  & 4096  & 4096 \\
    \midrule
    \multicolumn{9}{c}{\textit{\textbf{Learned Count Sketch (LCS)}}} \\
    \midrule
    \textbf{\# of heavy buckets} & 512   & 1536  & 2048  & 3072  & 3584  & 4608  & 5632  & 6144 \\
    \midrule
    \textbf{Width of sketch} & 512   & 1024  & 2048  & 2048  & 3096  & 3096  & 4096  & 4096 \\
    \midrule
    \multicolumn{9}{c}{\textit{\textbf{Elastic Sketch (ES)}}} \\
    \midrule
    \textbf{\# of buckets} & 50    & 50    & 50    & 75    & 70    & 90    & 92    & 100 \\
    \midrule
    \textbf{\# of entries} & 20    & 38    & 30    & 50    & 50    & 64    & 65    & 75 \\
    \midrule
    \textbf{Width of sketch} & 512   & 1024  & 2048  & 2048  & 3096  & 3096  & 4096  & 4096 \\
    \midrule
    \multicolumn{9}{c}{\textit{\textbf{Nitrosketch (NS)}}} \\
    \midrule
    \textbf{Width of sketch} & 1024  & 2048  & 3072  & 4096  & 5120  & 6144  & 7168  & 8192 \\
    \midrule
    \multicolumn{9}{c}{\textit{\textbf{UnivMon (UM)}}} \\
    \midrule
    \textbf{Width of sketch} & 1024  & 1248  & 1536  & 2972  & 3560  & 4096  & 4580  & 5120 \\
    \midrule
    \multicolumn{9}{c}{\textit{\textbf{PR-sketch}}} \\
    \midrule
    \textbf{Width of sketch} & 900   & 1800  & 2700  & 3600  & 4500  & 5400  & 6300  & 7200 \\
    \midrule
    \multicolumn{9}{c}{\textit{\textbf{FlowRadar}}} \\
    \midrule
    \textbf{Width of sketch} & 810  & 1620  & 2430  & 3240  & 4060  & 4870  & 5680  & 6490  \\
    \bottomrule
    \end{tabular}}
  \label{tab:baseline_params}
\end{table}

\textbf{Parameters of UCL-sketch.}
The parameters of the local sketch and hash table for each memory setting on Retail dataset are listed in Table~\ref{sketch_parameters}. As for the Bloom Filter, we determine the maximum number of bits by setting the coverage proportion over 99$\%$ according to Lemma~\hyperref[lemma2]{2} and fixing $k_b$ as 8. Regarding other datasets, when the space is small, the proportion of the Bloom Filter is larger (over 50$\%$), but there is a memory limit (for example, lower than 10KB for Kosarak and Retail dataset). As the space becomes more abundant, we gradually increase the allocation to the hash table. We provide detailed parameters for the learned solver of this work in Table~\ref{model_parameters}. We adopt the same key hyperparameter of neural network throughout the experiments. In particular, $\lambda$ is the hyperparameter that reweights the sparse term in Eqn.~3. For optimization, we use Adam optimizer~\cite{2014Adam} with default $(\beta_1, \beta_2)$ for all the experiments.

\vspace{-2mm}

\begin{table}[htbp]
  \centering
  \caption{Parameter configurations of the local sketch and hash table}
    \resizebox{0.5\textwidth}{!}{\begin{tabular}{c|cccccccc}
    \toprule
    \textbf{Memory Usage} & \textbf{16KB}  & \textbf{32KB}  & \textbf{48KB}  & \textbf{64KB}  & \textbf{80KB}  & \textbf{96KB}  & \textbf{112KB} & \textbf{128KB} \\
    \midrule
    \textbf{\# of slots in HF} & 500   & 1500  & 2000  & 3000  & 3500  & 4500  & 5500  & 6000 \\
    \midrule
    \textbf{Depth of sketch} & 4     & 4     & 4     & 4     & 6     & 6     & 6     & 8 \\
    \midrule
    \textbf{Width of sketch} & 512   & 512   & 1024  & 1024  & 1024  & 1024  & 1024  & 1024 \\
    \bottomrule
    \end{tabular}}
  \label{sketch_parameters}
\end{table}

\vspace{-2mm}

\begin{table}[htbp]
  \centering
  \caption{hyperparameters setting and overhead for learning parts}
    \begin{tabular}{ccc}
    \toprule
    Hyperparameter & Setting value & Refer range \\
    \midrule
    Bucket length & 512  & [128, 256, 512, 1024, 2056] \\
    Hidden dimension & 128   & [32, 64, 128, 256, 512] \\
    Trade-off $\lambda$ & 0.1   & 0.05 $\sim$ 1 \\
    Training epoch & 300   & 100 $\sim$ 500 \\
    Patience & 30    & 10 $\sim$ 50 \\
    Learning rate & 0.001 & 0.0001 $\sim$ 0.01 \\
    Batch size & 32    & [8, 16, 32, 64, 128] \\
    Sliding window length & 128    & [32, 64, 128, 256, 512] \\
    Sampling interval & 1000 & [500, 1000, 2000] \\
    \midrule
    Training time (per epoch) & \multicolumn{2}{c}{0.2s} \\
    Inference time & \multicolumn{2}{c}{0.3s $\sim$ 0.5s} \\
    Trainable parameters & \multicolumn{2}{c}{0.75MB $\sim$ 1.25MB} \\
    \bottomrule
    \end{tabular}
  \label{model_parameters}
\end{table}

\textbf{Data Normalization.}
Since there is no predefined end to a stream, meaning that reliable statistics (mean, variance, and maximum) do not exist, the data normalization in our implementation is operated on each individual sample in a batch separately. Now recalling the sketch update procedure, when a streaming item arrives, its volume is added to one counter in each row, where the counter is determined by $h_j$, $1 \le j \le d$. Therefore, counters in our sketch have the following guarantee: any inserted frequency should $\le 
{\rm{scale}} := \mathop {\min }\limits_{1 \le j \le d} {\mathop {\max }\limits_{1 \le i \le w} {{\rm{sketch\_count}}}\left[ {j,i} \right]}$. All we need is to find the minimum of the maximum counts from all the rows, which can be done in linear time. Then we calculate the instance-normalized measurement $y' = \frac{y}{{\rm{scale}}}$, and the inverse transformation for final estimations can be written as $x = {\rm{scale}} \times x_{\theta}$, in which $x_{\theta}$ is the output of the learned solver.

\begin{figure}[htbp]
    \centering
    \centerline{\includegraphics[width=1.\linewidth]{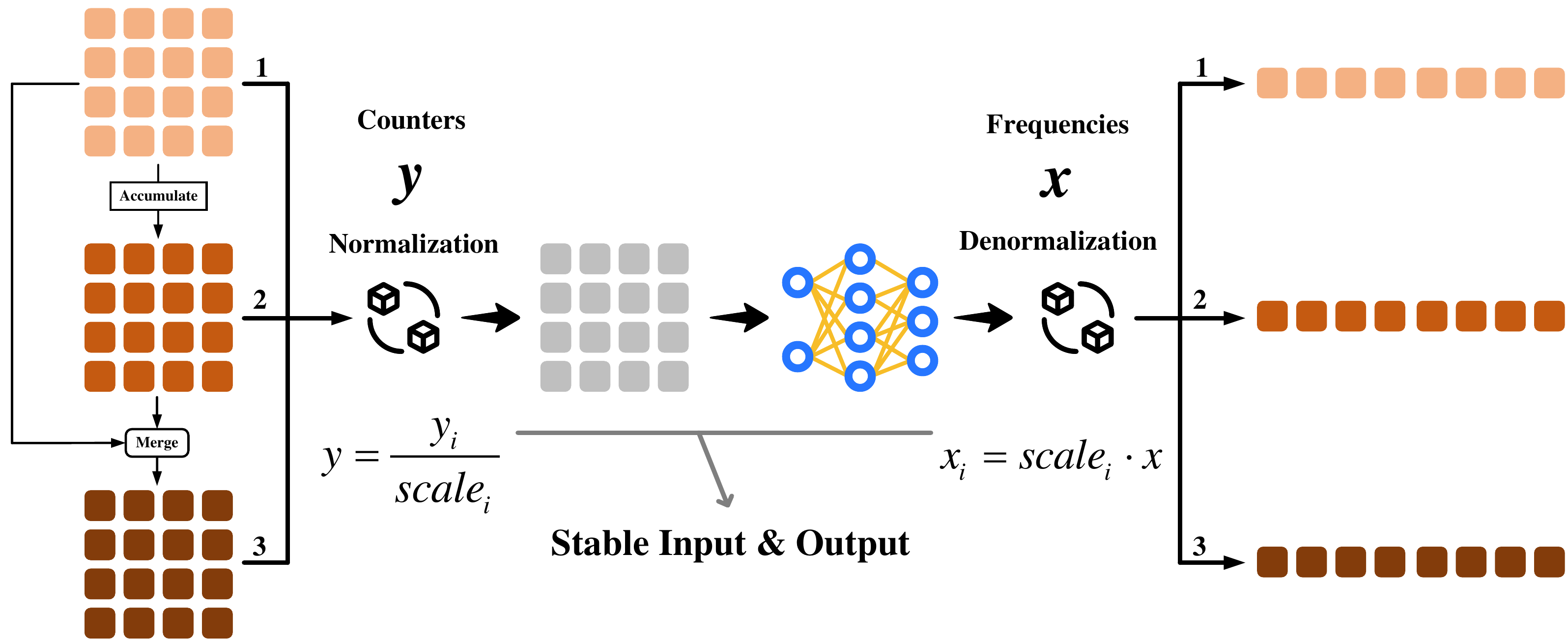}}
    \caption{Our normalization mechanism enables the solver to maintain stable reconstruction under natural changes.}
    % \vspace{-0.3cm}
    \label{fig:normalization}
\end{figure}

\subsection{Linear Sketch Case Study}
In this section, we present several plug-and-play case studies of our proposed framework, demonstrating that it is not restricted to any specific base sketch design.  
Recall the definition of a linear sketch, we consider the following five representative instances:  
\circled{1} \textbf{Count-Min Sketch}~\cite{cormode2005improved}: each update adds a positive increment ($+\Delta$) to fixed counters.  
\circled{2} \textbf{Count Sketch}~\cite{charikar2002finding}: updates can be either positive or negative ($\pm\Delta$), and counter entries may take negative values.  
\circled{3} \textbf{Augmented Sketch}~\cite{roy2016augmented}: the data stream first enters the \emph{heavy part}, and evicted flows are then inserted into a Count-Min sketch.  
\circled{4} \textbf{FlowRadar}~\cite{li2016flowradar}: similar to Count-Min but equipped with a Bloom Filter; the sketch width is fixed to one.  
\circled{5} \textbf{Elastic Sketch}~\cite{yang2018elastic}: similar to Augmented Sketch, but the heavy part is organized as a multi-layer structure.

For both training and inference, we only replace the sketch matrix $\boldsymbol{A}$ (of the light part), without modifying the solver architecture or learning pipeline. Next, the modifications we focus on are solely structural optimizations of the data plane, as heavy part and hash style can be different. Taking Count Sketch as an example, we only need to add an additional sign hash (for constructing $\boldsymbol{A}$ and recording) on top of the original UCL-sketch. Table~\ref{tab:implementation} summarizes the implementation effort, measured in terms of additional lines of code (LoC), when adapting UCL-sketch to different linear sketches. From the table, adapting UCL-sketch to various linear sketches incurs negligible implementation effort: all variants require fewer than 20 additional lines of code (due to the complex heavy part of Elastic Sketch), and none require changes to the training (or/and testing) process.  
These results confirm the plug-and-play nature of our framework—once trained, the learned solver can be seamlessly reused across different sketch instances with only lightweight modifications to the recording component.

\begin{table}[t]
\centering
\caption{Implementation complexity across different linear sketches.}
\begin{tabular}{lccc}
\toprule
\textbf{Linear Sketch} & \textbf{Add. LoC} & \textbf{Recording Mod.} & \textbf{Training Mod.} \\
\midrule
Count-Min Sketch & 0 & No & No \\
Count Sketch & 2$\sim$5 & Minor (hash) & Minor (hash) \\
Augmented Sketch & 5$\sim$10 & Minor (HF) & No \\
FlowRadar & 2$\sim$5 & Minor (arrays) & No \\
Elastic Sketch & 10$\sim$20 & Minor (HF) & No \\
\bottomrule
\end{tabular}
\label{tab:implementation}
\end{table}

\subsection{Robustness to Sketch Changes} In this part, we discuss the impact of sketch changes on our strategy by considering three different scenarios. Here, one measurement period of a sketch is referred to as a cycle.

\subsubsection{Natural change between (or during) cycles} First, natural changes in the magnitude of frequencies (e.g.,  those caused by accumulating data over time or merging two sketches with identical configurations) do not affect the final estimation of UCL-sketch after de-normalization. The robustness of the learned solver to instance-level normalization stems from the fact that the values within the counters are always scaled proportionally. From Fig.~\ref{fig:normalization}, this invariance ensures that the relative relationships among frequency estimates are preserved, rendering the normalization transparent to the sketch.

\subsubsection{Sudden change during current cycle} 
\begin{figure}[htbp]
    \centering
    \centerline{\includegraphics[width=1.\linewidth]{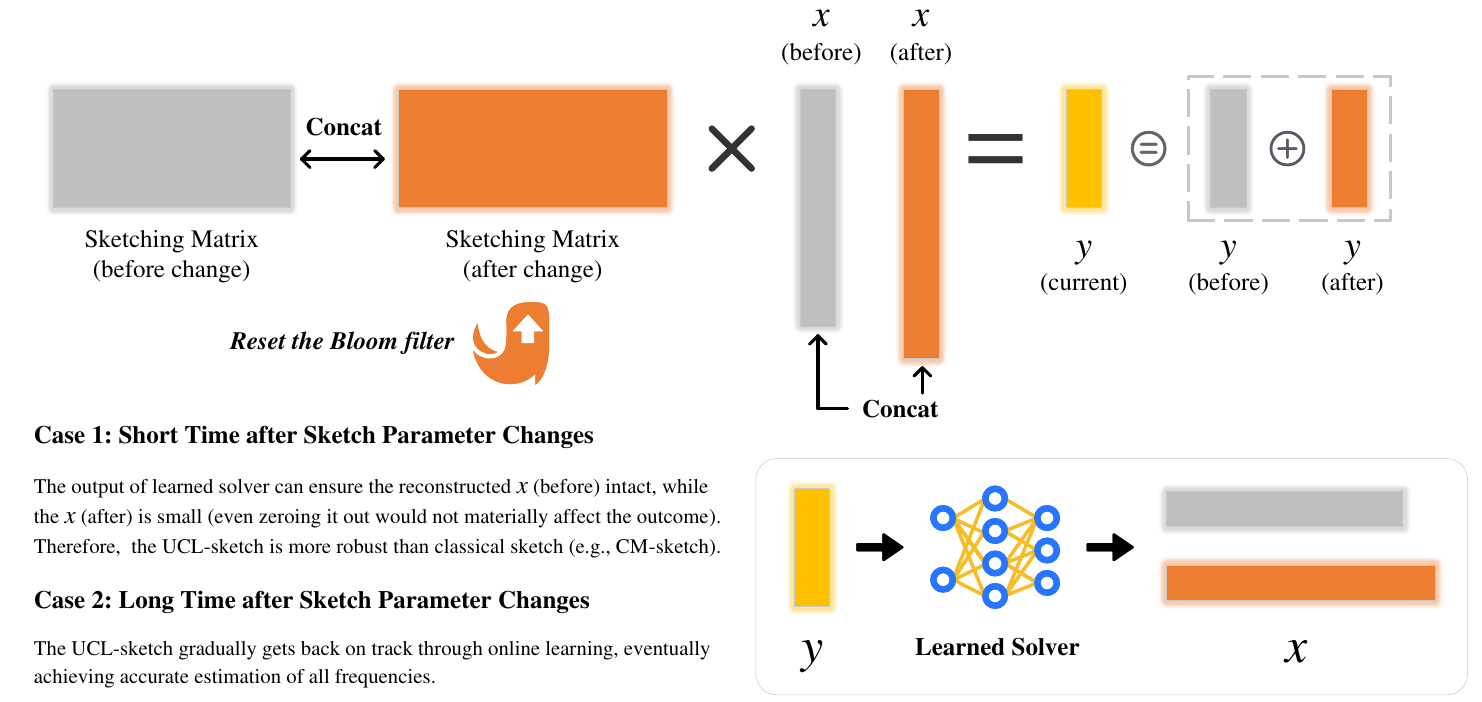}}
    \caption{The overall framework under changes to the sketch during execution.}
    % \vspace{-0.3cm}
    \label{fig:sketch_changes}
\end{figure}

Second, consider the scenario where critical components—such as hash functions—are modified mid-operation. In classical sketches (e.g., Count-Min), this immediately invalidates all prior aggregations, as the key-to-counter mapping changes fundamentally. The entire sketch state thus becomes inconsistent and effectively unusable without a full reset. In contrast, our framework accommodates such transitions gracefully. As illustrated in Fig.~\ref{fig:sketch_changes}, we can retain the original sketch parameters (tagging pre- and post-change data accordingly) and append new soft constraints for the updated structure. This is equivalent to expanding the sensing matrix with additional columns—an operation naturally supported by our online learning mechanism. Two temporal regimes emerge:
(i) Shortly after the change – UCL-sketch continues to yield valid frequency estimates for items inserted before the change. New items—particularly those filtered via the hash table—have negligible impact on per-key estimation accuracy~\cite{aamand2024improved}.
(ii) Long after the change – The solver fully adapts to the new hash functions, delivering accurate estimates for all items.
The transition is rapid, typically completing within 5 seconds (see the related experiment in the main text). By preserving history rather than discarding it, our approach offers markedly greater robustness than conventional methods.

\subsubsection{Reset-induced change at cycle onset} Third, changes to sketch parameters are typically planned—for example, during system reconfiguration or shifts in traffic regimes. If a full reset is necessary, retraining is indeed required; however, the cost is minimal thanks to our model’s lightweight architecture (see related experiment in the main text). During this brief retraining period (typically 1–5 seconds), the system can safely revert to classical estimation methods, such as standard Count-Min decoding or linear reconstruction. This fallback is particularly effective immediately after a reset, when the number of active flows ($|F|_1$) is small, making the system less underdetermined. In such cases, classical methods can remain reasonably accurate, with estimation error bounded by $\varepsilon_c |F|_1$. Whenever they consistently outperform UCL-sketch, the system can seamlessly default to them.

\subsection{Comparison with Equation-based Sketches} 

We conduct additional analysis comparing UCL-sketch with three other sketching algorithms equipped with a Bloom Filter: namely, FlowRadar~\cite{li2016flowradar} SeqSketch~\cite{huang2021toward}, and PR-sketch~\cite{sheng2021pr}. 

In addition to the Bloom filter, FlowRadar maintains a counting table where each cell consists of three fields: FlowXOR, FlowCount, and PacketCount. When updating (and hashing) a flow count, the PacketCount field is incremented. If a new flow is inserted, its key is additionally XORed with FlowXOR, and FlowCount is incremented by 1. During decoding, the table is traversed to find cells where FlowCount equals 1; for these cells, the decoding key is XORed with FlowXOR, the PacketCount value is retrieved and then reset to 0. Note that not all flows can be successfully decoded. For keys that cannot be decoded but are recorded by the Bloom Filter, we use Count-Min to estimate their corresponding frequencies. 

Regarding SeqSketch and PR-sketch, they are based on exactly the same principle (i.e., compressive sensing) as ours. The main differences in data structures are that SeqSketch uses a fractional sensing matrix, while PR-sketch does not employ a hash table. Moreover, unlike our approach equipped with a learning-based solver for decoding pre-key frequencies, the \textsc{LeastSqareEqationSolver} used in SeqSketch and PR-sketch is OMP~\cite{pati1993orthogonal} and LSMR~\cite{fong2011lsmr}, respectively. We implemented these two iterative optimization algorithms using the SciPy library~\cite{2020SciPy-NMeth}. For a fair comparison, we also implemented their parallel versions in PyTorch, called SeqSketch (G) and PR-sketch (G).

\begin{figure}[htbp]
\centering
\subfigure[CAIDA Dataset]{
\includegraphics[width=1.\linewidth]{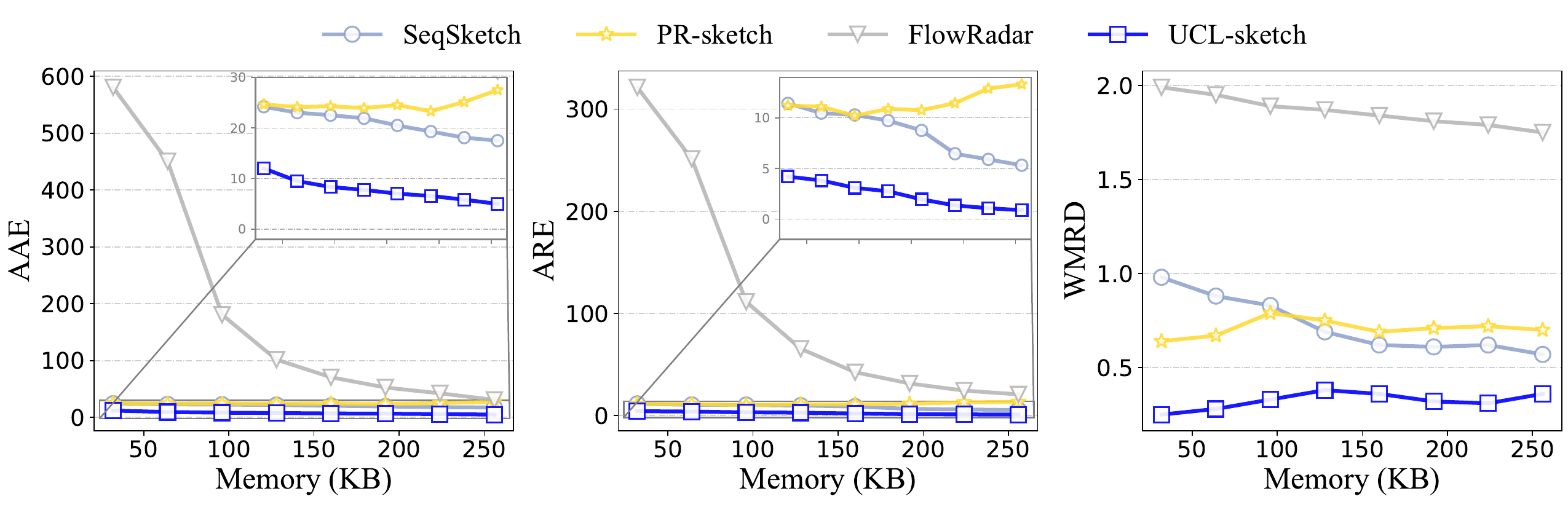}
}
\\
% \vspace{-0.2cm}
\subfigure[Kosarak Dataset]{
\includegraphics[width=1.\linewidth]{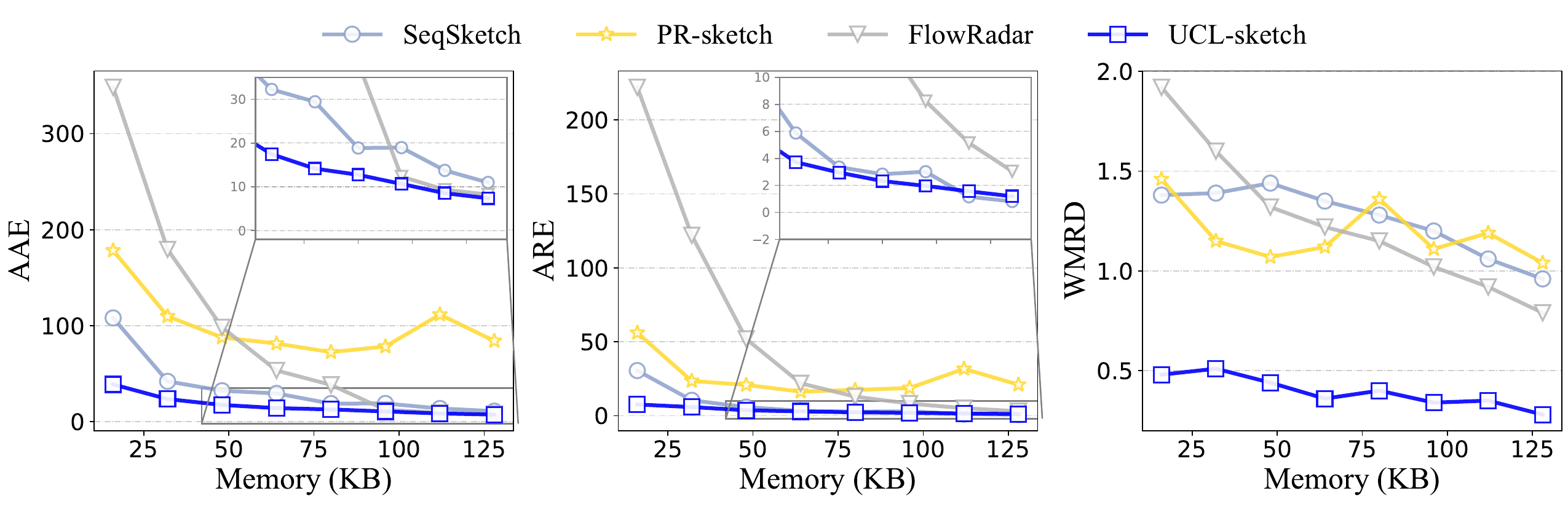}
}
\\
% \vspace{-0.2cm}
\subfigure[Retail Dataset]{
\includegraphics[width=1.\linewidth]{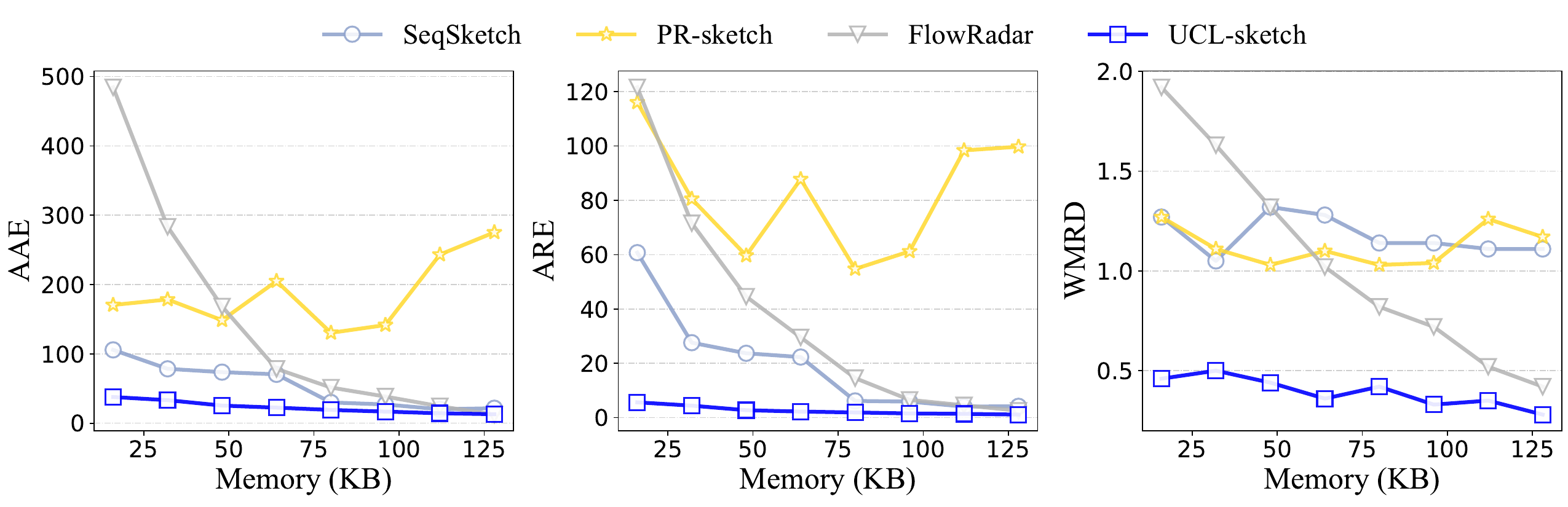}
}
\caption{Performance comparison of equation-based sketches.}
\label{fig:equation_sketch_performance}
% \vspace{-0.45cm}
\end{figure}

\begin{table}[htbp]
  \centering
  \caption{Entropy absolute error of equation-based sketches}
    \resizebox{0.5\textwidth}{!}{\begin{tabular}{c|c|c|c|c|c|c|c|c}
    \toprule
    \multicolumn{9}{c}{\textit{\textbf{CAIDA Dataset}}} \\
    \midrule
    \textbf{Memory Usage} & \textbf{32KB} & \textbf{64KB} & \textbf{96KB} & \textbf{128KB} & \textbf{160KB} & \textbf{192KB} & \textbf{224KB} & \textbf{256KB} \\
    \midrule
    \textbf{FlowRadar} & 2430.44 & 729.56 & 560.10 & 291.34 & 226.56 & 184.73 & 93.94 & \multicolumn{1}{c}{69.60} \\
    \midrule
    \textbf{PR-sketch} & 30.73 & 46.53 & 54.08 & 53.26 & 69.28 & 84.98 & 99.4  & 126.53 \\
    \midrule
    \textbf{SeqSketch} & 33.49 & 40.49 & 27.29 & 34.81 & 26.49 & 21.57 & 14.95 & 12.06 \\
    \midrule
    \rowcolor[rgb]{ .851,  .882,  .949} \textbf{UCL-sketch} & 9.68  & 7.54  & 4.51  & 3.39  & 3.12  & 2.82  & 2.9   & 1.42 \\
    \midrule
    \multicolumn{9}{c}{\textit{\textbf{Kosarak Dataset}}} \\
    \midrule
    \textbf{Memory Usage} & \textbf{16KB} & \textbf{32KB} & \textbf{48KB} & \textbf{64KB} & \textbf{80KB} & \textbf{96KB} & \textbf{112KB} & \textbf{128KB} \\
    \midrule
    \textbf{FlowRadar} & 2908.92   & 1326.97   & 741.77   & 569.50   & 397.03   & 207.72   & 141.95    & 75.07  \\
    \midrule
    \textbf{PR-sketch} &  1290.96  & 1378.43   & 1057.47   &  1822.91  &  1029.92  & 1236.26   & 1567.18    & 2500.03  \\
    \midrule
    \textbf{SeqSketch} &  515.29  &  380.91  & 385.68   &  347.38  &  39.31  &  26.47   &  28.99   & 56.76  \\
    \midrule
    \rowcolor[rgb]{ .851,  .882,  .949} \textbf{UCL-sketch} & 58.24  & 35.10 & 21.63 & 16.85 & 14.29 & 13.19 & 10.79 & 10.18 \\
    \midrule
    \multicolumn{9}{c}{\textit{\textbf{Retail Dataset}}} \\
    \midrule
    \textbf{Memory Usage} & \textbf{16KB} & \textbf{32KB} & \textbf{48KB} & \textbf{64KB} & \textbf{80KB} & \textbf{96KB} & \textbf{112KB} & \textbf{128KB} \\
    \midrule
    \textbf{FlowRadar} & 4638.81   & 2339.08   & 1058.84   & 631.51   & 380.87   & 245.74   & 159.66    & 109.21  \\
    \midrule
    \textbf{PR-sketch} & 1117.30   & 644.23   & 585.01   & 706.02   & 1793.55   & 758.57   & 1331.14 & 808.83  \\
    \midrule
    \textbf{SeqSketch} & 315.25    & 168.68   & 98.35   & 84.04   & 92.37   & 128.70   & 102.68    & 111.06  \\
    \midrule
    \rowcolor[rgb]{ .851,  .882,  .949} \textbf{UCL-sketch} & 69.16   & 41.02   & 27.14   & 22.20   & 18.87   &  15.07  &  12.56   & 10.56  \\
    \bottomrule
    \end{tabular}}
  \label{tab:eae_es}
\end{table}

The first two columns in Fig.~\ref{fig:equation_sketch_performance} present the performance of each method in terms of accuracy. There is a clear ranking among UCL-sketch, SeqSketch and PR-sketch on each real-world dataset, where the learning-based algorithms consistently yield improvement over traditional optimization-based algorithms. The measurement system is always undetermined, and the fact that traditional approaches rely solely on sparsity constraints is the key reason. Also note that the performance of PR-sketch is highly unstable in comparison with SeqSketch, and even degrades as available memory increases. 
In addition to OMP being a more advanced—and consequently more computationally expensive—algorithm, we believe that the sparsity pattern of $\boldsymbol{A}$ (where each column only contains exactly 4 non-zero entries, independent of the number of counters or rows) may be a key factor resulting in this phenomenon. Regarding FlowRadar, its accuracy is similar to that of traditional sketches and is even worse than the CM-sketch presented in the main text. This is because it is primarily designed for collaborative measurement across multiple sketches in a network. This performance difference arises as it allocates extra space for components like Bloom Filters and can decode only a very limited number of flows. This phenomenon is most pronounced in the CAIDA dataset, highlighting the superiority of compressive sensing.

From Table~\ref{tab:eae_es} and the third column of Fig.~\ref{fig:equation_sketch_performance}, it can be observed that UCL-sketch exhibits a clear, step-change advantage over other methods in the distribution of estimated frequencies. As previously discussed, FlowRadar consistently performs poorly in global distribution estimation, as it is designed to accurately decode only a small subset of (large) flows. Overall, PR-sketch ranks second to last in terms of WMRD and entropy absolute error, and its performance is still unstable across evaluations.  Although SeqSketch is comparable to PR-sketch in terms of WMRD, it performs notably well in entropy estimation. However, an obvious gap remains between SeqSketch and UCL-sketch.

\begin{figure}[htbp]
    \centering
    \centerline{\includegraphics[width=1.\linewidth]{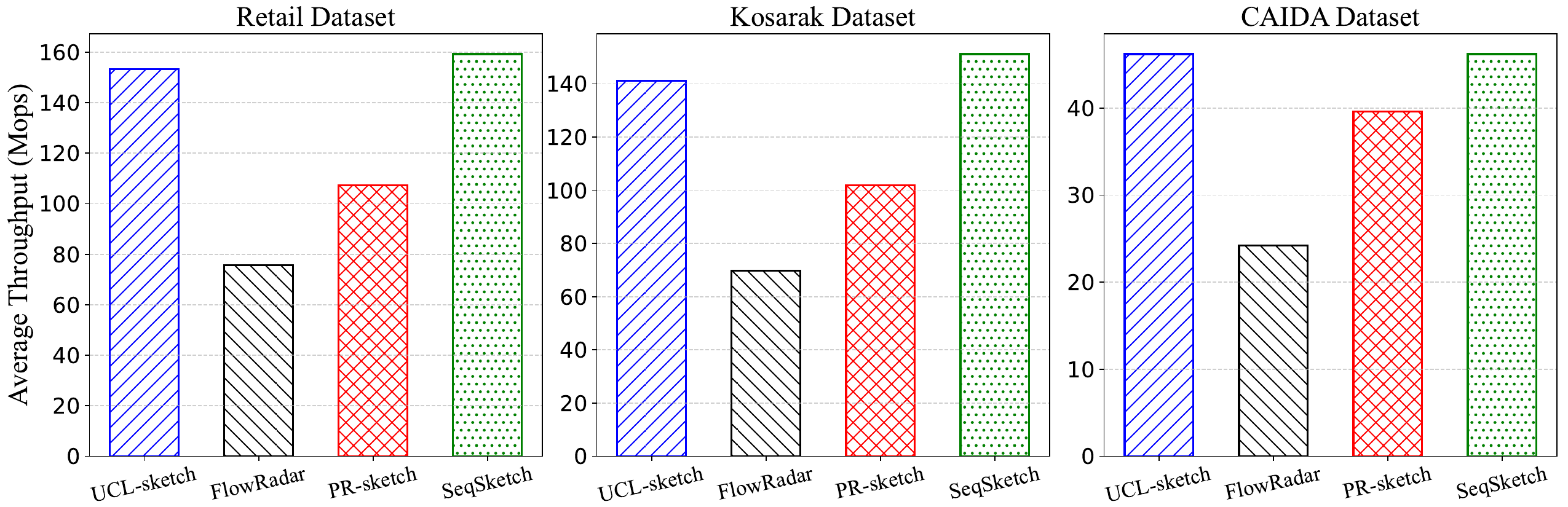}}
    \caption{Average results on insertion throughput comparison.}
    % \vspace{-0.3cm}
    \label{fig:es_throughput}
\end{figure}

\begin{figure}[htbp]
    \centering
    \centerline{\includegraphics[width=1.\linewidth]{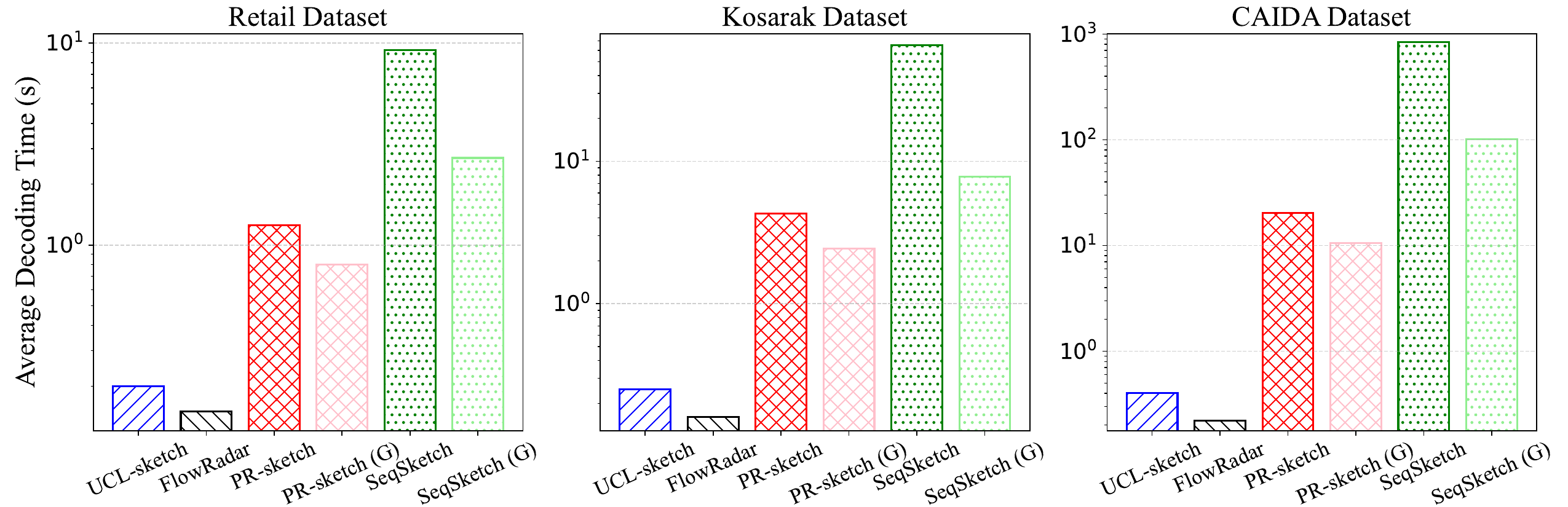}}
    \caption{Average results on decoding speed comparison.}
    % \vspace{-0.3cm}
    \label{fig:es_decoding}
\end{figure}

Next, we present the insertion throughput of these sketching algorithms in Fig.~\ref{fig:es_throughput}. The high processing speed of UCL-sketch and SeqSketch comes from the fact that most insert operations occur in the hash table. PR-sketch follows, while FlowRadar requires additional XOR operations, resulting in the lowest throughput. Moreover, as shown in Fig.~\ref{fig:es_decoding}, we measure their average decoding time. FlowRadar has the fastest decoding speed, as it only performs a few loop checks on the underlying counters (and quickly exits the loop due to decoding failure), which is fundamentally different from the compressed sensing approach used by other methods. Nevertheless, the inference time of UCL-sketch also takes less than 0.5 seconds. Doubling the key space only slightly increases the inference time. In contrast, the decoding speed of SeqSketch and PR-sketch is significantly slower compared to UCL-sketch. This disparity can be attributed to the inherent computational complexity associated with these traditional compressed sensing techniques. Both OMP and LSMR require iterative processes to reconstruct the original data from the sketch, leading to a substantial increase in processing time. Even when parallel acceleration techniques are applied to enhance the performance of SeqSketch and PR-sketch, their decoding speeds still lag behind UCL-sketch. UCL-sketch, with its ML-driven solver, thus emerges as a more efficient alternative, particularly in environments where high-speed data processing is essential.

\end{document}